\pdfoutput=1
\documentclass{article}
\usepackage{iclr2026_conference,times}

\usepackage{amsmath,amsfonts,bm}

\def\eqref#1{equation~\ref{#1}}

\def\plaineqref#1{\ref{#1}}

\def\1{\bm{1}}

\DeclareMathAlphabet{\mathsfit}{\encodingdefault}{\sfdefault}{m}{sl}
\SetMathAlphabet{\mathsfit}{bold}{\encodingdefault}{\sfdefault}{bx}{n}

\usepackage{hyperref}
\usepackage{url}
\usepackage{graphicx}
\usepackage{wrapfig}
\usepackage{xcolor}
\usepackage{array}
\usepackage{booktabs}
\usepackage{longtable}

\newcounter{figurepanel}[figure]

\hypersetup{
  pdftitle={Scaling Laws for LLM Memory Sparsity},
  pdfauthor={Maciej Stefaniak, Romain Lepert, Mart Van Baalen, Rob Hesselink, Paul N. Whatmough, Babak Ehteshami Bejnordi}
}

\title{Scaling Laws for LLM Memory Sparsity}

\author{Maciej Stefaniak$^{1,2}$\thanks{Work done during an internship at Qualcomm AI Research.} \quad Romain Lepert$^{3}$ \quad Mart Van Baalen$^{3}$ \\
\textbf{Rob Hesselink$^{3}$ \quad Paul N. Whatmough$^{3}$ \quad Babak Ehteshami Bejnordi$^{4}$} \\
$^{1}$IDEAS Research Institute \quad $^{2}$University of Warsaw \quad $^{3}$Qualcomm AI Research\addtocounter{footnote}{1}\thanks{Qualcomm AI Research is an initiative of Qualcomm Technologies, Inc.\newline Correspondence to: \href{mailto:m.stefaniak8@uw.edu.pl}{\texttt{m.stefaniak8@uw.edu.pl}}} \quad $^{4}$NVIDIA
}

\iclrfinalcopy
\begin{document}

\maketitle
\fancyhead{}
\renewcommand{\headrulewidth}{0pt}
\addtocontents{toc}{\protect\setcounter{tocdepth}{-1}}

\begin{abstract}
Sparse memory architectures offer a route to expanding language-model capacity
while accessing only a small fraction of parameters per token. Their finer
access granularity promises greater sparsity than conventional mixture-of-experts
(MoE), but how memory capacity should scale with active computation and training
data remains insufficiently understood.
We study how architecture and sparsity should vary with parameter and
training-compute budgets. We introduce scaling laws for Product Key Memory (PKM)
that capture how active parameters, memory capacity, and training tokens jointly
determine loss, using 421 measured training endpoints from 44 configurations
spanning dense Transformers, PKM, token-indexed memory (STEM), and MoE, with total
parameter capacities reaching 21.9B and training budgets reaching 128B
tokens.
The fitted laws reveal distinct regions of architectural advantage: PKM achieves
the lowest predicted loss when capacity is abundant relative to compute, MoE is
preferred in an intermediate regime, and dense Transformers become preferable
when compute is abundant relative to capacity. Within PKM, increasing capacity
at fixed compute favors greater sparsity, whereas increasing compute at fixed
capacity favors more active parameters. Along a hardware-calibrated scaling
path, PKM maintains lower predicted loss at higher optimal sparsity than MoE,
permitting more training tokens and fewer operational FLOPs per token.
A complementary roofline analysis estimates a $7.6\times$ PKM
token-throughput advantage over MoE in the modeled mobile inference setting.
Together, these findings characterize when sparse memory offers advantages in
predicted loss and execution efficiency.
\end{abstract}

\section{Introduction}

Scaling laws have become central to the development of large language models
(LLMs), turning smaller experiments into predictions that guide costly training
runs~\citep{kaplan,chinchilla}. By relating performance to capacity, data, and
computation, they help identify promising model structures and training settings
before committing resources to large-scale experiments~\citep{granularity,architecure_sl}.
Among these design decisions, sparsity is a central architectural choice. Mixture-of-Experts
(MoE) expands capacity while activating only selected experts per token,
improving the trade-off between model quality and computation~\citep{clark,granularity}.
It has become a standard architectural choice across leading open-weight and
proprietary LLM families~\citep{kimiteam2026k25,glm5team2026glm5,deepmind2026gemini31pro,xu2026deepseek}.
Its success motivates exploring how far this separation of capacity from
computation can extend.

An emerging direction is sparse parametric memory: large trainable tables from
which each token accesses only a few entries. Addressing can use token identities
or learned queries from hidden states~\citep{stem,gemma3n_ple,pkm}.
Such fine-grained retrieval offers a route to extreme sparsity and cheaper
capacity expansion, with recent results showing competitive quality and lower
memory-access costs than conventional MoE~\citep{memory_scale,ultramem,cltramemv2}.
This requires understanding how additional memory interacts
with active computation and training data, and when its benefits survive
hardware constraints.

\begin{center}
\refstepcounter{figure}\label{fig:architecture-memory-compute-optimal-loss}
\begin{minipage}[t]{0.62\textwidth}
\vspace{0pt}
\centering
\includegraphics[width=0.8664\linewidth,pagebox=cropbox,clip]{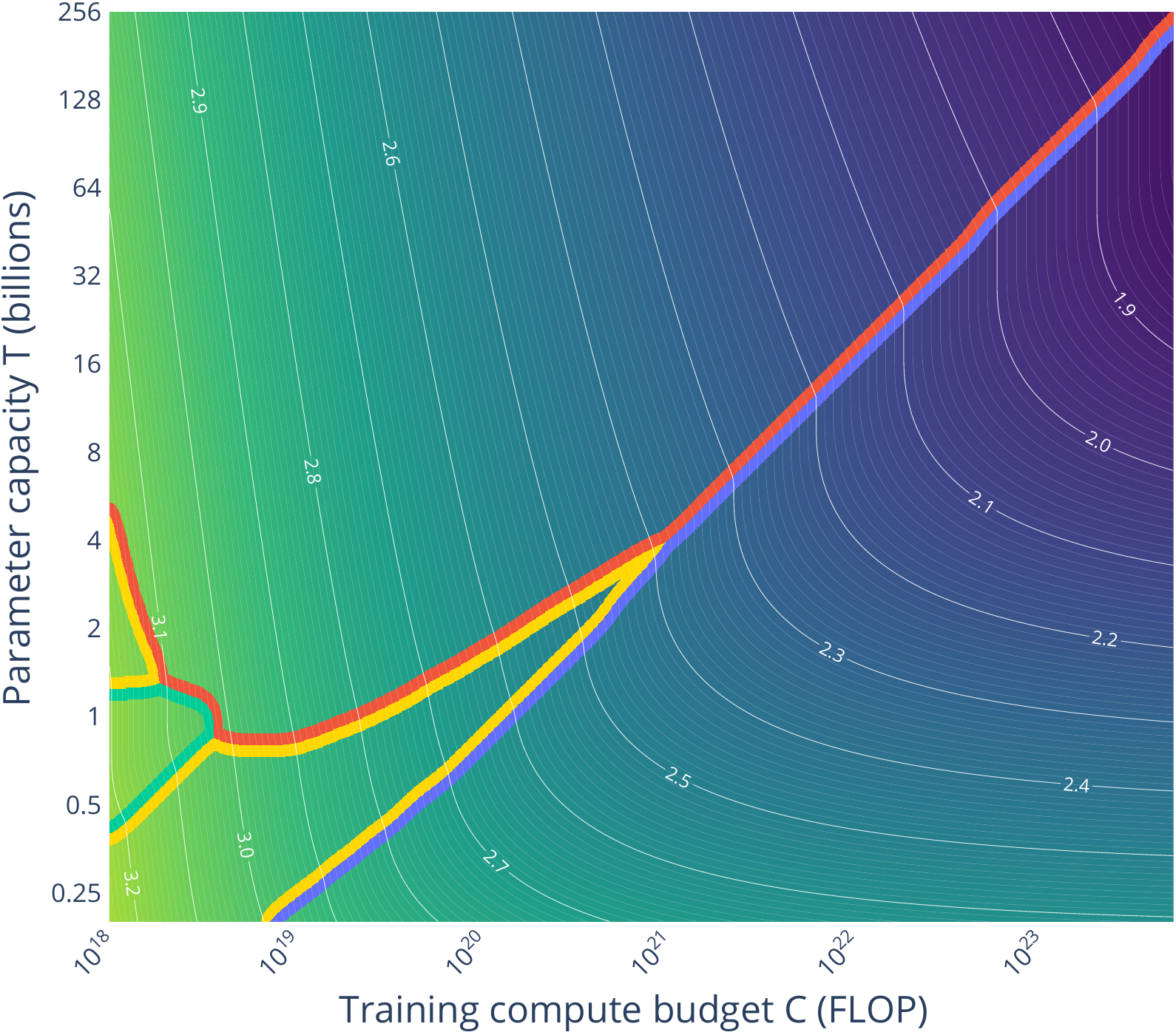}
\end{minipage}\hfill
\begin{minipage}[t]{0.355\textwidth}
\vspace{3.75bp}
\includegraphics[width=\linewidth,pagebox=cropbox,clip]{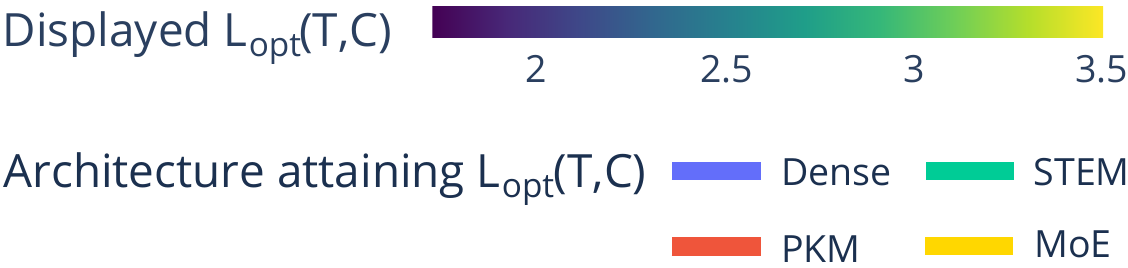}
\par\vspace{12pt}
{\small\raggedright\textbf{Figure \thefigure.} \emph{Architecture attaining lowest loss at parameter and FLOPs constrains.} At each training compute budget $C$ and parameter-capacity
budget $T$, heatmap shading shows the lowest predicted loss across architectures. Thin contours mark equal displayed $\widehat
L_{\mathrm{opt}}$ levels every $0.01$. Colored solid boundaries delineate regions
where each architecture attains the lowest predicted loss. Stability and failure regions of this comparison: Appendix~\ref{sec:stability}. \par}
\end{minipage}
\end{center}

We therefore study the scaling laws of sparse memory, focusing on Product Key
Memory (PKM), whose learned retrieval supports varying bank size independently
of vocabulary size and retrieval count. Using 421 measured training endpoints
from 44 PKM, MoE, STEM, and dense Transformer configurations, we connect loss
prediction to resource allocation and hardware efficiency. Our contributions are:

\begin{itemize}
\item \textbf{Scaling laws for sparse memory.} We formulate $L(N,D,M)$ to
capture how memory capacity interacts with active parameters and training data,
together with a compact two-variable law for dense and STEM models.
The latter achieves lower held-out error than Chinchilla in all 32 reported
comparisons and Farseer~\citep{steplaw2} in 27. We further find that
architecture-specific fits improve prediction: sharing coefficients between
PKM and MoE worsens held-out MoE prediction in every matched test.

\item \textbf{Resource-dependent architecture and sparsity selection.} We map
architectures minimizing predicted loss across parameter and
compute budgets (Figure~\ref{fig:architecture-memory-compute-optimal-loss}): PKM
is favored when parameter capacity is abundant relative to compute, MoE in an intermediate
regime, and dense Transformers when compute is abundant relative to parameter capacity.
Optimal PKM sparsity increases with the total parameter budget at fixed compute
and decreases with compute at a fixed parameter budget. Along a hardware-calibrated scaling path, PKM
maintains lower predicted loss at higher optimal sparsity than MoE, permitting
more training tokens and fewer operational FLOPs per token.

\item \textbf{Hardware efficiency of sparse memory.} We use a roofline model
to show how retrieval granularity affects mobile inference when model capacity
exceeds DRAM. PKM's fine-grained retrieval reduces flash traffic, yielding a
modeled $7.6\times$ token-throughput advantage over MoE at a processing window
of 512 tokens. In this setting, PKM sustains throughput as memory capacity
grows, whereas MoE slows as more experts are added.
\end{itemize}

\section{Related work}

\noindent\textbf{Memory models.}
Parametric memory can use token-based addresses or learned content-based
retrieval. Gemma 3n's Per-Layer Embeddings (PLE)~\citep{gemma3n_ple} and
STEM~\citep{stem} use token-indexed embeddings; STEM replaces the FFN
up-projection while retaining contextual gating. Engram~\citep{engram} extends
deterministic lookup to $n$-grams. Product Key Memory (PKM)~\citep{pkm} learns
queries from hidden states, allowing memory selection to vary with context.
Memory layers at scale~\citep{memory_scale} and
UltraMem/UltraMemV2~\citep{ultramem,cltramemv2} develop this retrieval-based
line through improvements to capacity, retrieval, and value processing.
PEER~\citep{moe_million} retrieves tiny experts, while hierarchical
memories~\citep{hierarchical} separate long-tail knowledge from the active model.
We focus on baseline PKM because its bank size can vary independently of
vocabulary size and the number of retrieved values, providing a controlled
setting for studying capacity scaling with learned contextual addressing.
STEM, dense Transformers, and MoE serve as comparative references.

\noindent\textbf{Scaling laws.}
Power-law descriptions of language-model loss~\citep{kaplan} underpin
compute-optimal allocation of parameters and training tokens~\citep{chinchilla}.
Farseer refines their joint loss surface by allowing token scaling to depend
on model size~\citep{steplaw2}; related work models
architectural choices for inference efficiency~\citep{architecure_sl}, downstream
benchmark performance~\citep{eval_sl}, and training hyperparameters such as
learning rate and batch size~\citep{steplaw}.
Building on Chinchilla and Farseer, we introduce interactions among active
parameters, memory capacity, and training tokens, and compare our related
two-variable law against both baselines using held-out prediction on dense
and STEM models.

\noindent\textbf{MoE scaling laws.}
Routed-model laws separate parameter capacity from computation~\citep{clark};
later studies examine expert granularity and fine-grained
architectures~\citep{granularity,moe_more},
optimal sparsity~\citep{pvsf}, and joint dense--MoE scaling under memory and
compute constraints~\citep{joint}. We extend this resource-allocation analysis
to PKM, compare its optimal sparsity and predicted loss with MoE at matched
parameter and compute budgets, and test whether a shared coefficient vector
describes both architectures. Our analysis uses architecture-specific fits
and complements loss-based comparisons with roofline models of
communication and memory-access costs.

\raggedbottom
\section{Experimental Setup}
\label{sec:experimental-setup}

Our experiments quantify how Product Key Memory (PKM)~\citep{pkm} scales. We
ask how increasing its sparsely accessed memory capacity changes predictive
loss at a fixed active model size and training-token budget, and how this
trade-off informs allocation under parameter and compute constraints. Scaling
Transformers with Embedding Modules (STEM)~\citep{stem} provides a second
memory-model reference. Mixture-of-Experts (MoE)~\citep{shazeer2017moe}
provides comparative context from an architecture with sparsely activated
expert parameters, while dense Transformers provide the baseline.

\subsection{Model Families and Scaling Variables}
\label{sec:model-families-and-variables}

We consider decoder-only Transformers with different feed-forward modules.
The dense baseline applies its feed-forward parameters to every token.
PKM uses product-key retrieval to access a subset of a learned value
bank, whereas MoE routes tokens to a subset of
feed-forward experts. STEM replaces the feed-forward
up-projection with a token-indexed embedding lookup while retaining dense
gating and down-projection. We vary PKM memory-bank
size and MoE expansion across backbone sizes. For STEM, we vary backbone
size under a fixed memory configuration rather than independently sweeping
memory capacity.

Let $L$ denote the measured final cross-entropy using a common evaluation data subset, $N$ the active non-embedding
parameter count, and $D$ the training token budget. For PKM
and MoE, we additionally use the capacity coordinate $M$, representing
value-bank parameters for PKM and inactive expert parameters for MoE.
We define the law-accounted total capacity and sparsity as
$T=N+M$.

These coordinates separate active computation from additional stored capacity.
Here, non-embedding accounting excludes vocabulary embeddings, but retains
the internal memory modules being studied. In particular, the PKM coordinate
$T$ is an accounting proxy, not an exact measure of physical checkpoint or
training-memory storage.

For the dense baseline, $M=0$ and $T=N$. STEM is also fitted as $L(N,D)$,
because its memory capacity is not varied independently of backbone size;
consequently, its effect cannot be identified as a separate scaling-law input.
This does not imply that STEM has no inactive parameters. PKM and MoE are
instead fitted as $L(N,D,M)$.  

\subsection{Training Setup and Experimental Coverage}
\label{sec:training-setup-and-coverage}
The training corpus is DCLM-Edu, and the models use the GPT-NeoX-20B
tokenizer~\citep{black2022gptneox}. We distinguish a model configuration,
which fixes the architecture and its capacity settings (N, M), from an endpoint,
which records its measured loss at a particular token budget (B).  All endpoints
belonging to one configuration form a model lineage. 

Across PKM, MoE, STEM, and the dense Transformer baseline, we use a shared
backbone and common training configuration, with architecture-specific
memory and feed-forward modules. These settings and the architecture-specific
hyperparameters are detailed in Appendix~\ref{sec:model-training-details}.
The experimental design targets approximately uniform coverage on logarithmic
scales for model sizes, training-token budgets, and sparsity ratios where
these quantities are varied.

The fitting dataset comprises 421 measured endpoints from 44 model
configurations (192 PKM, 124 MoE, 57 dense, and 48 STEM endpoints).
Active non-embedding parameters reach 2.0B, and
training budgets reach 128B tokens. The largest PKM models reach 21.9B
total parameters, with memory up to 139 times the active parameter count. MoE models reach 6.4B
total parameters.

\section{Scaling Laws for Memory Sparsity}
\label{sec:memory-scaling-laws}

Scaling laws are empirical functions that describe how model performance changes as computational resources are scaled.  Dense-model laws commonly express loss in terms of parameter count $N$ and training tokens $D$, because nearly all model parameters are active for every token.  Sparse memory architectures break this correspondence: two models with the same active parameter count and training budget can have very different stored capacities. We therefore extend the usual $L(N,D)$ formulation with the independently varied inactive parameters coordinate $M$ defined. We fit and validate the resulting $L(N,D,M)$ laws for PKM and MoE, then test whether a common scaling relationship can describe both sparse mechanisms.

\subsection{Functional Form}
\label{sec:scaling-functional-form}

The Chinchilla formulation~\citep{chinchilla} decomposes
loss into an irreducible term and separate model- and data-dependent terms,
\begin{equation}
\widehat L_{\mathrm C}(N,D)
=E+A N^{-\alpha}+B D^{-\beta}.
\label{eq:memory-law-chinchilla}
\end{equation}
Here, the amplitude and exponent of the data-dependent term are independent
of model size. Farseer~\citep{steplaw2} relaxes this assumption by
allowing both to depend on $N$, expressing loss as
$F_U(N)+B_{\mathrm F}(N)D^{-A_{\mathrm F}(N)}$.

Our formulation starts from the Chinchilla decomposition and progressively
introduces the additional-capacity coordinate $M$ according to practices established in literature (~\citep{joint}, ~\citep{steplaw2}, ~\citep{granularity},  ~\citep{pvsf},  ~\citep{clark}). During this process, we
also introduce an interaction between $N$ and $D$, following the same general
motivation as Farseer ~\cite{steplaw2} while extending it to architectures with independently
varied active and total capacity. The candidate formulas and fitting
comparisons are described in Appendix~\ref{sec:fitting-scaling-laws}.
For PKM and MoE, these extensions yield our central result: a memory-sparsity
scaling-law family in which inactive capacity enters both the finite-model and
data-dependent contributions,
\begin{equation}
\boxed{
\begin{aligned}
\widehat L(N,D,M)
={}&\underbrace{A N^{-\alpha}M^{-\kappa}}_{
U(N,M):\ \text{finite-model term}}\\
&+\underbrace{B N^{-\lambda}M^{-\mu}
D^{-\left(\beta+\zeta\log M+\xi\log N\right)}}_{
V(N,D,M):\ \text{data-dependent term}}.
\end{aligned}
}
\label{eq:memory-interaction-law}
\end{equation}
We denote the token exponent by
$q_D(N,M)=\beta+\zeta\log M+\xi\log N$. Unlike a dense $L(N,D)$ law, this
form lets $M$ influence the finite-model contribution, the amplitude of the
data-dependent contribution, and its token-scaling exponent.

For the dense Transformer and STEM, memory is not independently varied in
the fitting data. We therefore use the related six-coefficient family
\begin{equation}
\boxed{
\widehat L(N,D)=A N^{-\alpha}
+B N^{-\lambda}D^{-(\beta+\xi\log N)}.
}
\label{eq:two-variable-interaction-law}
\end{equation}
This family retains the model-dependent amplitude and token exponent but omits
all independently fitted $M$ terms; This is not obtained by substituting $M=0$ into
Equation~\ref{eq:memory-interaction-law}. The learnable parameter vectors
for the two formula families are
\[
\boldsymbol{\theta}_{N,D,M}
=(A,B,\alpha,\beta,\kappa,\lambda,\mu,\zeta,\xi),
\qquad
\boldsymbol{\theta}_{N,D}
=(A,B,\alpha,\beta,\lambda,\xi).
\]
The first vector is fitted separately for PKM and MoE, and the second is fitted
separately for the dense Transformer and STEM. These architecture-specific
fits do not share coefficients; a joint PKM--MoE law fitted to the pooled data is examined separately
in Section~\ref{sec:shared-scaling-laws}.

\begin{center}
\begin{minipage}{\textwidth}
\centering
\refstepcounter{figure}\label{fig:pkm-scaling-law-diagnostics}
\begin{minipage}[t]{0.495\textwidth}
\centering
\refstepcounter{figurepanel}\label{fig:pkm-observed-predicted}
{\small\textbf{(\alph{figurepanel})} Observed versus predicted loss\par}
\smallskip
\includegraphics[width=0.802\linewidth,trim=75.3bp 10.1bp 95.6bp 22.5bp,clip]{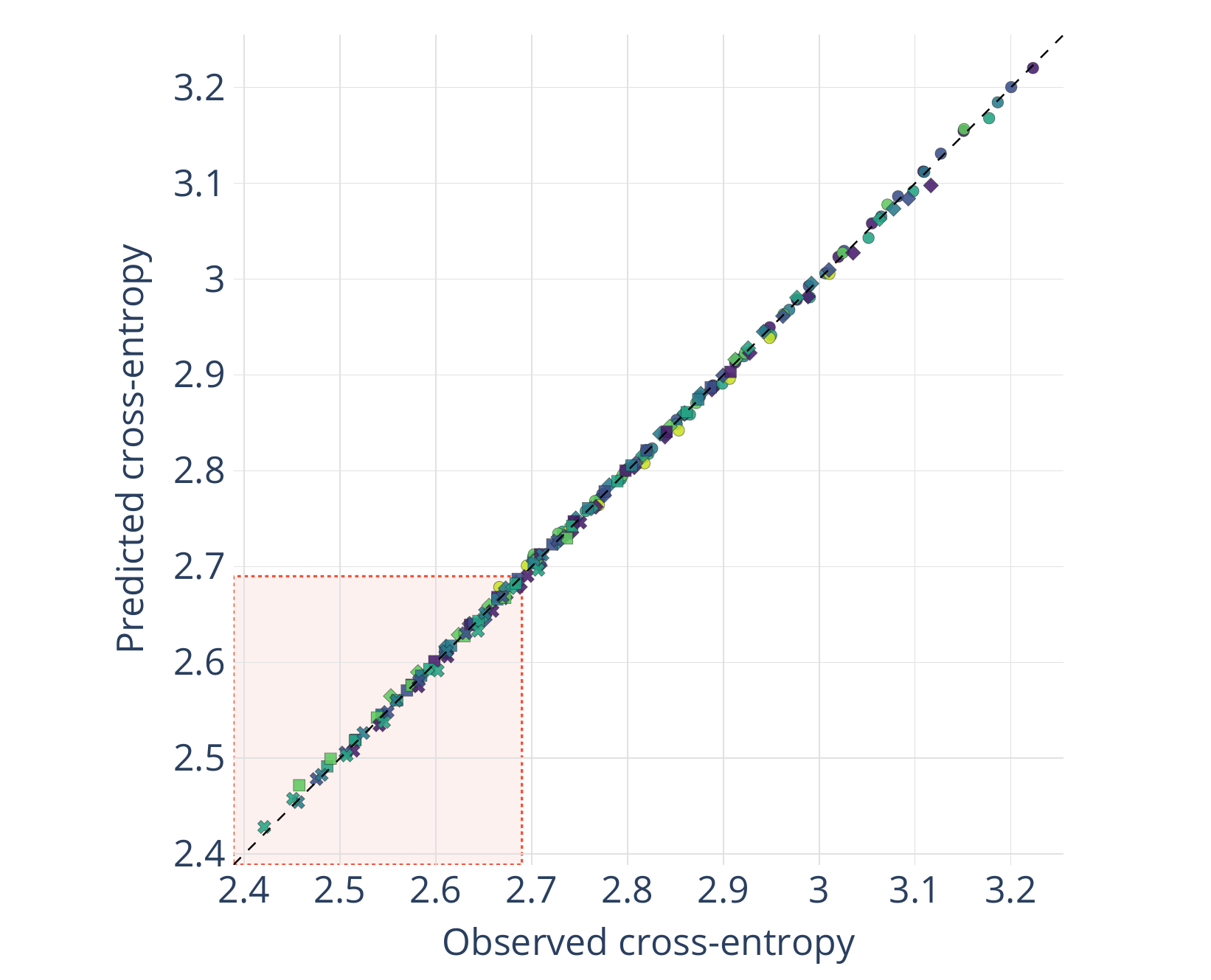}
\end{minipage}\hfill
\begin{minipage}[t]{0.495\textwidth}
\centering
\refstepcounter{figurepanel}\label{fig:pkm-token-curves}
{\small\textbf{(\alph{figurepanel})} Loss versus training tokens\par}
\smallskip
\includegraphics[width=0.98\linewidth,trim=13.4bp 10.1bp 14.3bp 17.3bp,clip]{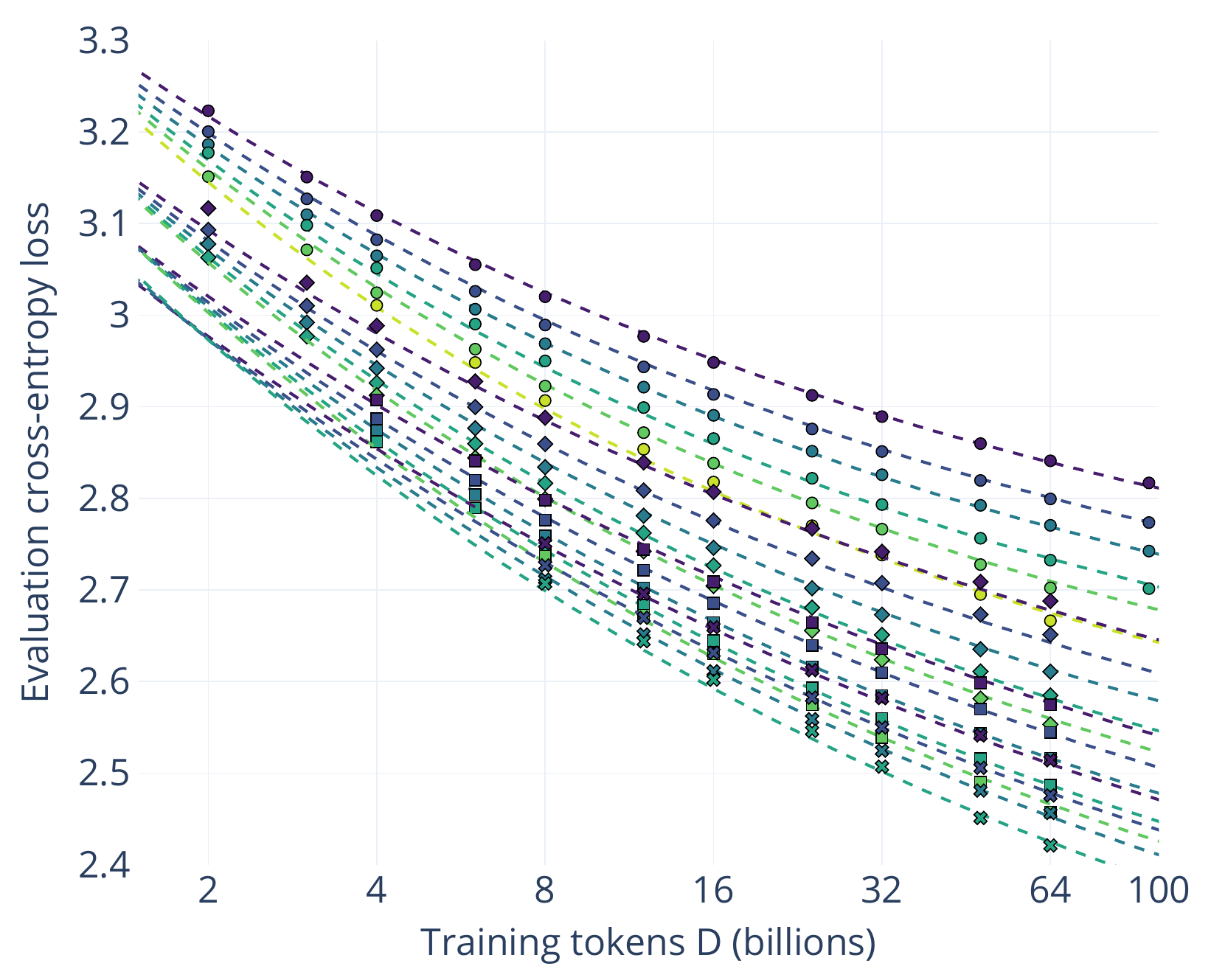}
\end{minipage}
\par\vspace{7pt}
\includegraphics[width=0.98\textwidth,trim=24.7bp 18.8bp 136.9bp 25.9bp,clip]{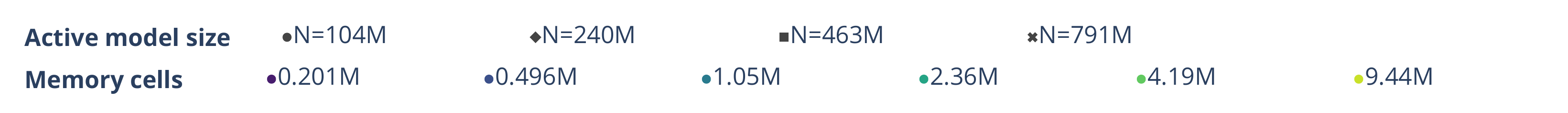}
\par\vspace{4pt}

{\small\raggedright\textbf{Figure \thefigure.} \emph{PKM scaling-law diagnostics.}
(a) Observed versus predicted cross-entropy under LLLO$_{30}$. The training subset has RMSE $4.805\mathrm{e}{-3}$; the held-out subset
has RMSE $5.860\mathrm{e}{-3}$. The shaded square marks
the validation envelope and the diagonal denotes exact prediction.
(b) Measured endpoint losses (markers) and fitted curves (dashed lines). Marker shapes indicate active model size;
colors indicate the number of cells in memory bank, not parameters.\par}
\end{minipage}
\end{center}

\subsection{Predictive Accuracy and Extrapolation}
\label{sec:scaling-predictive-accuracy}

We compare 19 three-variable candidates on PKM, MoE, and the joint PKM--MoE data,
and three two-variable candidates on the dense Transformer and STEM.
The candidates progressively introduce memory dependence, model--data
interactions, and alternative amplitude parameterizations.
Appendix~\ref{sec:fitting-scaling-laws} gives the complete formulas and their evaluation results. Comparisons use the refitting and validity protocol in
that appendix, rather than in-sample error alone.

Refitting every candidate on the complete data and each declared holdout
produces 3,115 reported architecture--function--split fits. The evaluation
suite tests prediction toward the lowest observed losses, withheld model
configurations, size and memory-capacity families, future token budgets,
and combinations of unseen capacities and training budgets, including the
largest jointly supported size--capacity--token corner.

In the leave-lowest-losses-out test, denoted LLLO$_p$, we withhold the
lowest-loss $p\%$ of endpoints, refit the law on the remaining higher-loss
endpoints, and predict the withheld losses. We use
$p\in\{10,20,30,40,50\}$. This tests prediction toward the best observed
losses; it does not necessarily withhold entire model configurations.
Figure~\ref{fig:pkm-observed-predicted} shows LLLO$_{30}$ predictions of the
PKM law (Equation~\ref{eq:memory-interaction-law}), and
Figure~\ref{fig:pkm-token-curves} its fitted loss curves against the
training-token budget.
The grouped and directional holdouts complement this test by withholding
complete configurations or specified ranges of the scaling coordinates. The
full-data and LLLO RMSE evaluations of the reported formulas are summarized in
Table~\ref{tab:selected-law-full-lllo-rmse}. Evaluation used in other scaling
law papers: LLLO$_{20\%}$ RMSE = $19\mathrm{e}{-3}$ for MoE and
LLLO$_{20\%}$ RMSE = $15\mathrm{e}{-3}$ for Dense (~\citep{granularity}),
LLLO$_{10\%}$ RMSE = $3.9\mathrm{e}{-3}$ for joint MoE and dense Transformer
(~\citep{joint}), full fit mean RMSE = $11\mathrm{e}{-3}$ for dense Transformer
downstream tasks evaluation (~\citep{eval_sl}), error ranging from
$2.6\mathrm{e}{-3}$ to $7.2\mathrm{e}{-3}$ for dense Transformer beyond both
unseen capacities and training budget (~\citep{steplaw2}).

\begin{center}
\begin{minipage}{\textwidth}
\centering
\refstepcounter{table}\label{tab:selected-law-full-lllo-rmse}
\textbf{Table \thetable.} \emph{Full-data and LLLO RMSE of the reported scaling laws
($\mathrm{e}{-3}$); fitted coefficients are reported in Appendix
Table~\ref{tab:selected-law-coefficients}.}

\resizebox{0.75\textwidth}{!}{%
\begin{tabular}{lcccccc}
\hline
Architecture & Full-data & LLLO$_{10}$ & LLLO$_{20}$ & LLLO$_{30}$ & LLLO$_{40}$ & LLLO$_{50}$ \\
\hline
PKM & 4.58 & 4.32 & 5.01 & 5.86 & 7.39 & 9.27 \\
MoE & 5.80 & 5.36 & 6.08 & 8.49 & 9.11 & 9.79 \\
Joint PKM--MoE & 8.74 & 8.28 & 8.16 & 9.76 & 11.9 & 12.6 \\
Dense Transformer & 4.65 & 6.49 & 6.24 & 6.08 & 9.30 & 10.6 \\
STEM & 1.76 & 1.27 & 2.82 & 2.88 & 4.45 & 5.18 \\
\hline
\end{tabular}%
}
\end{minipage}
\end{center}

For the dense Transformer and STEM, our six-coefficient law
(Equation~\ref{eq:two-variable-interaction-law}) has lower held-out error than
Chinchilla in all 32 reported architecture--metric--family comparisons,
and lower error than Farseer in 27 of 32; the complete comparison is given in
\hyperref[sec:two-variable-law-selection]{Appendix~\ref*{sec:fitting-scaling-laws},
\emph{Selection of the two-variable law}}. These counts combine two
architectures, MAE and RMSE, and eight held-out settings per architecture;
they are not 32 independent experiments. Farseer achieves lower full-data
error on both datasets, but uses nine coefficients rather than six.
The results favor our two-variable law as a compact predictive description of these data,
without implying it universally outperforms Farseer.

The selected law couples memory capacity and training data: at fixed active
size, $M$ changes both the finite-model component and the amplitude and exponent
of the data-dependent component. Over the observed PKM domain, the fitted loss
is locally less sensitive to $M$ than to $N$ or $D$. We therefore use the
complete predicted loss surface for the allocation analysis in
Section~\ref{sec:optimal-allocation};
Appendix~\ref{sec:capacity-token-analysis} derives the effective exponents and
reports the detailed sensitivity analysis.

\subsection{Joint PKM--MoE versus Architecture-Specific Scaling Laws}
\label{sec:shared-scaling-laws}

To test whether a common scaling surface describes PKM and MoE, we fit
Equation~\ref{eq:memory-interaction-law} with a single coefficient vector on
the pooled data (316 endpoints across 35 model configurations) and ask
whether pooling improves MoE prediction, whose separately fitted law has
higher LLLO$_{30}$ error than PKM. Table~\ref{tab:selected-law-full-lllo-rmse}
shows higher RMSE for the joint law than for the MoE-specific law on the
full-data fit and all five LLLO evaluations, although their LLLO test sets
are selected independently. We therefore perform a matched evaluation
(Appendix~\ref{sec:joint-matched-moe-lllo}): for each
LLLO$_{10}$--LLLO$_{50}$ split, we refit the joint law on all 192 PKM
endpoints plus the retained MoE endpoints and evaluate both laws on the same
held-out MoE endpoints. All five refits converge and pass the
endpoint-identity and train--test separation checks, and the joint law's
RMSE is $1.35$--$2.05$ times higher in every split. Pooling thus degrades MoE
prediction under matched evaluation, providing direct evidence of different
PKM and MoE scaling behavior; we therefore use
architecture-specific laws both to compare the scaling behavior of PKM and
MoE and for the analysis in Section~\ref{sec:optimal-allocation}.

\section{Compute and sparsity optimality}
\label{sec:optimal-allocation}

The fitted scaling laws predict loss as a function of active parameters,
training tokens, and inactive parameters. For each architecture and under
fixed total-parameter and compute budgets, we identify the allocation between
active and inactive parameters that minimizes predicted loss.

\subsection{Sparsity at Fixed Parameter and Compute Budgets}
\label{sec:fixed-budget-allocation}

At a fixed total capacity $T=N+M$, we express the sparsity between active
parameters and memory through the active fraction $r=N/T$ and memory fraction
$S=M/T=1-r$.
We refer to $S$ as sparsity under our parameter-accounting convention. For
PKM, $M$ counts the full value bank, so $S$ is an accounting-based memory
fraction rather than an exact fraction of physically inactive parameters.
We approximate training compute by
$C\approx6ND$, following active-parameter FLOP accounting used in scaling
studies~\citep{kaplan}, choosing $r$ at fixed
$(T,C)$ determines all three inputs to the sparse-model scaling law:
\begin{align}
N&=rT,\qquad M=(1-r)T,\qquad D=\frac{C}{6rT}.
\label{eq:main-allocation-coordinates}
\end{align}
Substitution into Equation~\ref{eq:memory-interaction-law} gives the
one-dimensional objective
\begin{align}
\widehat L_{\mathrm{alloc}}(r;T,C)
&:=\widehat L\!\left(rT,\frac{C}{6rT},(1-r)T\right).
\label{eq:main-allocation-objective}
\end{align}
For the feasible active-fraction set $\mathcal R(T,C)$, we define
\par\noindent
\begin{minipage}[t]{0.49\linewidth}
\vspace{-8pt}
\begin{align}
r_{\mathrm{opt}}(T,C)&\in
\mathop{\mathrm{argmin}}_{r\in\mathcal R(T,C)}
\widehat L_{\mathrm{alloc}}(r;T,C),
\label{eq:main-optimal-allocation}
\end{align}
\end{minipage}\hfill
\begin{minipage}[t]{0.49\linewidth}
\vspace{-8pt}
\begin{align}
\widehat L_{\mathrm{opt}}(T,C)&=
\widehat L_{\mathrm{alloc}}\!\left(r_{\mathrm{opt}}(T,C);T,C\right).
\label{eq:main-optimal-loss-sparsity}
\end{align}
\end{minipage}
\par
The optimal inactive fraction, active parameters, inactive
parameters, and training tokens are
\par\noindent
\begin{minipage}[t]{0.49\linewidth}
\vspace{-8pt}
\begin{align}
S_{\mathrm{opt}}(T,C)
&=1-r_{\mathrm{opt}}(T,C),
\label{eq:main-optimal-sparsity}
\end{align}
\end{minipage}\hfill
\begin{minipage}[t]{0.49\linewidth}
\vspace{-8pt}
\begin{align}
N_{\mathrm{opt}}(T,C)
&=r_{\mathrm{opt}}(T,C)T,
\label{eq:main-optimal-active}
\end{align}
\end{minipage}

\noindent
\begin{minipage}[t]{0.49\linewidth}
\vspace{-8pt}
\begin{align}
M_{\mathrm{opt}}(T,C)
&=\left[1-r_{\mathrm{opt}}(T,C)\right]T,
\label{eq:main-optimal-inactive}
\end{align}
\end{minipage}\hfill
\begin{minipage}[t]{0.49\linewidth}
\vspace{-8pt}
\begin{align}
D_{\mathrm{opt}}(T,C)
&=\frac{C}{6r_{\mathrm{opt}}(T,C)T}.
\label{eq:main-optimal-tokens}
\end{align}
\end{minipage}
\par
When an architecture
superscript is shown below, it denotes the optimum obtained from that
architecture's independently fitted scaling law.
We perform this minimization numerically and separately for PKM and MoE,
using their independently fitted coefficients. The expanded objective
and numerical procedure are given in
Appendix~\ref{sec:lowest-loss-allocation}.

\begin{center}
\begin{minipage}{\textwidth}
\centering
\refstepcounter{figure}\label{fig:sparse-optimal-allocation}
\begin{minipage}[t]{0.495\textwidth}
\centering
\refstepcounter{figurepanel}\label{fig:pkm-optimal-sparsity}
\includegraphics[width=0.98\linewidth,trim=13.7bp 51.9bp 13.1bp 31.0bp,clip]{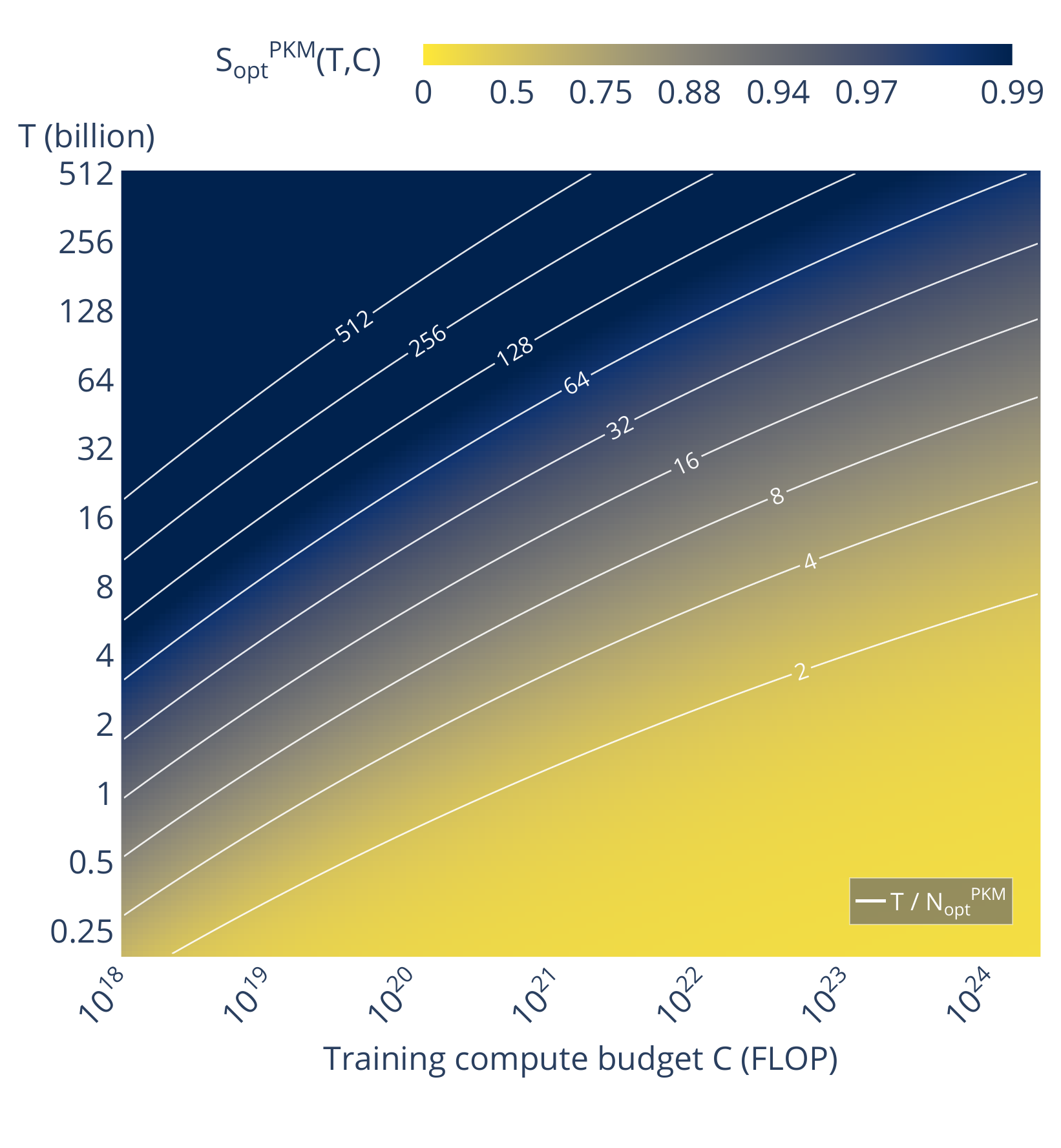}
\par\smallskip
{\small\textbf{(\alph{figurepanel})} PKM optimal sparsity S\par}
\end{minipage}\hfill
\begin{minipage}[t]{0.495\textwidth}
\centering
\refstepcounter{figurepanel}\label{fig:pkm-moe-active-parameter-ratio}
\includegraphics[width=0.98\linewidth,trim=13.7bp 51.9bp 14.9bp 31.0bp,clip]{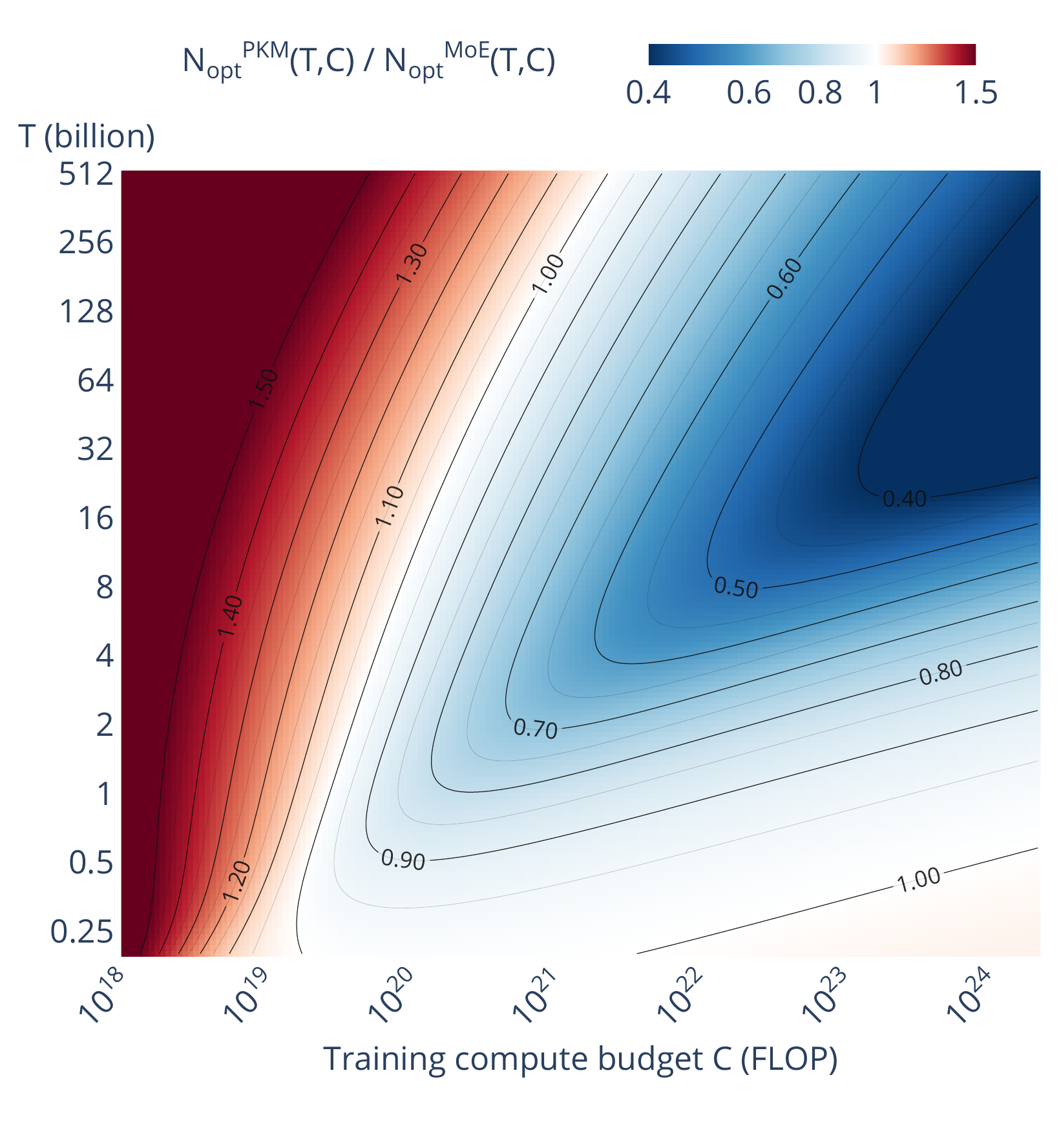}
\par\smallskip
{\small\textbf{(\alph{figurepanel})} PKM / MoE optimal active parameters ratio\par}
\end{minipage}
\par\medskip
{\small\raggedright\textbf{Figure \thefigure.} \emph{Optimal sparsity in PKM and MoE.}
Both panels show training compute $C$ (FLOP) horizontally and parameter
capacity $T$ (billions) vertically. Both sparse architectures use the full
available capacity in every displayed cell, so
$N_{\mathrm{opt}}+M_{\mathrm{opt}}=T$.
(a) Color shows the loss-minimizing PKM sparsity
$S_{\mathrm{opt}}^{\mathrm{PKM}}(T,C)$; White contours show
$T/N_{\mathrm{opt}}^{\mathrm{PKM}}$.
The corresponding MoE optimal-sparsity map is shown in
Appendix~\ref{sec:additional-results}, Figure~\ref{fig:moe-optimal-sparsity}.
(b) Each architecture is optimized independently at matched budgets.
Blue indicates fewer active parameters for PKM, red indicates fewer for
MoE, and white denotes parity.\par}
\end{minipage}
\end{center}

\subsection{Optimal Sparsity in PKM and MoE}
\label{sec:optimal-sparsity}

Figure~\ref{fig:pkm-optimal-sparsity} shows the predicted optimal PKM sparsity
$S_{\mathrm{opt}}^{\mathrm{PKM}}(T,C)$ over parameter and compute budgets;
the contours report the corresponding expansion ratio
$T/N_{\mathrm{opt}}^{\mathrm{PKM}}$.
At a fixed parameter budget, increasing compute favors a larger active
fraction, and therefore lower sparsity and a smaller expansion ratio.
Conversely, increasing parameter capacity at fixed compute favors allocating
a larger fraction of parameters to inactive capacity, producing higher
sparsity and a larger expansion ratio. The optimal sparsity therefore depends
jointly on parameter and compute budgets, rather than on total model size
alone.

Figure~\ref{fig:pkm-moe-active-parameter-ratio} compares the independently
optimized sparsities through
$N_{\mathrm{opt}}^{\mathrm{PKM}}(T,C)/
N_{\mathrm{opt}}^{\mathrm{MoE}}(T,C)$ at matched parameter and compute
budgets. Values below one indicate that PKM uses fewer active parameters than
MoE, and therefore has higher sparsity and a larger expansion ratio; values
above one indicate the converse. Because per-token inference compute scales
with the active parameter count, a higher-sparsity allocation also lowers the
cost of serving the model after training. PKM favors fewer active parameters when
compute is high relative to parameter capacity, whereas MoE favors fewer in
the lower-compute regime. This ratio characterizes differences in the
preferred sparsities, but does not compare the minimum losses achieved by
the two architectures. At the highest compute-to-parameter ratios, the ratio
approaches one because both architectures favor predominantly active
parameters and approach the dense limit, $N_{\mathrm{opt}}\approx T$.

\subsection{Architecture Comparison across Resource Budgets}
\label{sec:architecture-budget-comparison}

Figure~\ref{fig:architecture-memory-compute-optimal-loss} compares the
architecture-specific lowest predicted losses
$\widehat L_{\mathrm{opt}}^{(a)}(T,C)$ of the dense Transformer, STEM, PKM,
and MoE under matched parameter and compute budgets. At each budget, the
heatmap shows $\min_a\widehat L_{\mathrm{opt}}^{(a)}(T,C)$, while the colored
boundaries identify the architecture attaining this minimum. This
separates the question of how each architecture allocates its capacity from
the question of which achieves the lowest predicted loss.

The comparison reveals distinct resource regimes. The dense Transformer is
favored when compute is abundant relative to parameter capacity. As parameter
capacity increases relative to compute, the preferred architecture shifts
toward sparse alternatives: MoE occupies an intermediate regime, whereas PKM
is favored across a broader capacity-rich, compute-constrained region. Thus,
the architecture ranking depends on the specific combination of compute and
parameter-capacity budgets, rather than defining a single ordering across all
scales.

The sparsity analysis uses predicted loss together with the approximation
$C\approx6ND$, which counts active FLOPs but does not capture the reduction in
hardware efficiency that accompany parameters sparsity increase. Changes to the memory
module, expert design, or training configuration could also shift both the
preferred sparsity and the architecture ranking. We therefore treat these
results as loss-based comparisons of the studied architecture variants, distinct from the hardware considerations discussed in
Section~\ref{sec:hardware-implications}.

\subsection{Optimal Sparsity along Coupled Resource Scaling}
\label{sec:coupled-resource-scaling}

We next consider a coupled scaling scenario in which compute and parameter
capacity increase together. Figure~\ref{fig:h100-coupled-scaling} evaluates a
scaling path parameterized by $n$, with both the number of NVIDIA H100 80 GB
GPUs and training duration proportional to $n$. We calibrate both budgets to
the Qwen3-235B MoE configuration benchmarked with Megatron Core on 256 H100
GPUs in BF16 precision
(\citealp{nvidia2026megatronmoe}; details in Appendix~\ref{sec:coupled-scaling-hardware-assumptions}). This configuration corresponds to approximately 0.92 billion
model parameters per allocated GPU and a measured throughput of 320 TFLOP/s per
GPU, or 32.3\% Model FLOP Utilization (MFU). With $16n$ GPUs trained
for $n$ days, the resulting budgets are
\begin{equation}
C(n)=4.424\times10^{20}n^2\ \mathrm{FLOP},\qquad
T(n)=14.69n\times10^9\ \text{parameters}.
\label{eq:coupled-resource-budgets}
\end{equation}
The compute budget covers both training and operation of the model. In
addition to training on $D$ tokens, each model processes
$D_{\mathrm{op}}(n)=n\times100$B tokens during its operational lifetime,
at a cost of $2N$ FLOP per token. The budget is therefore split as
\begin{equation}
C(n)=6ND+2ND_{\mathrm{op}}(n),
\label{eq:coupled-training-operation-budget}
\end{equation}

Figure~\ref{fig:h100-coupled-optimal-loss} shows decreasing predicted loss
for all four architectures, with PKM lowest across the entire displayed
range, followed by MoE; the dense Transformer and STEM curves cross near
$n=16$. Along this path, the loss-optimal dense Transformer uses only about
$1/16$ of $T(n)$, placing the path in the capacity-rich regime of
Section~\ref{sec:architecture-budget-comparison}, where sparse architectures
are favored.

Figure~\ref{fig:h100-coupled-optimal-sparsity} shows that the optimal ratio
$T(n)/N_{\mathrm{opt}}(T(n),C(n))$ increases with $n$ and is consistently
higher for PKM (35 to 160) than for MoE (33 to 84). Fewer active parameters
also lower the operational cost $2ND_{\mathrm{op}}$, which explains part of
the sparse advantage over the dense Transformer: without this cost
(Appendix~\ref{sec:additional-results},
Figure~\ref{fig:coupled-training-only}), PKM remains lowest, but its
advantage over the dense Transformer at $n=128$ shrinks from 0.062 to 0.016
and MoE falls behind the dense Transformer for $n\geq16$, whereas PKM's
advantage over MoE is largely unchanged.

The curves are extrapolations of the fitted
laws under the assumed budget conversion; the hardware interpretation and
its limitations are discussed in Section~\ref{sec:hardware-implications}.

\begin{center}
\begin{minipage}{\textwidth}
\centering
\refstepcounter{figure}\label{fig:h100-coupled-scaling}
\begin{minipage}[t]{0.49\textwidth}
\vspace{0pt}
\centering
\refstepcounter{figurepanel}\label{fig:h100-coupled-optimal-loss}
{\small\textbf{(\alph{figurepanel})} Lowest predicted loss
$\widehat L_{\mathrm{opt}}$\par}
\smallskip
\includegraphics[width=0.98\linewidth,trim=0bp 4bp 10bp 20bp,clip]{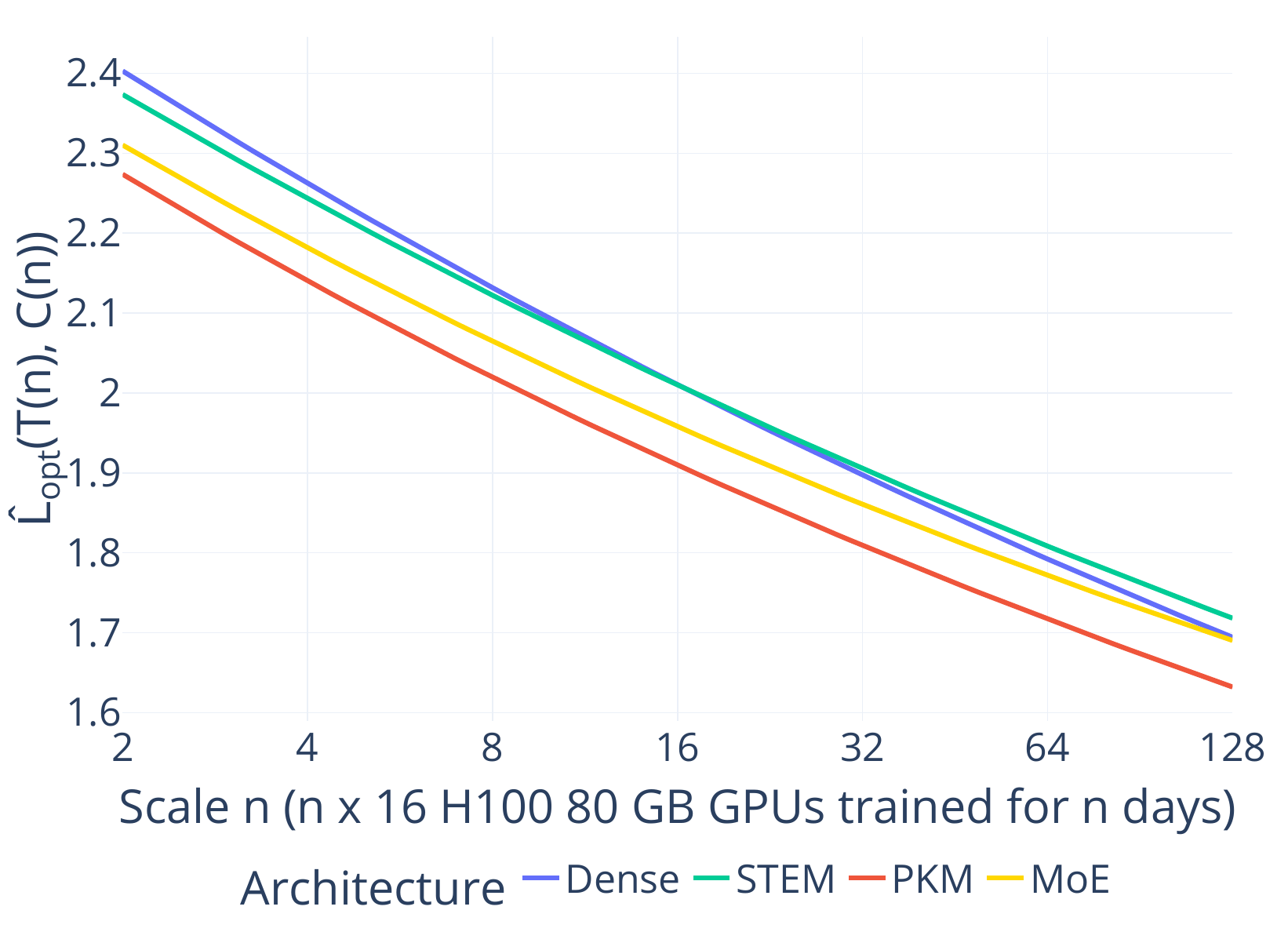}
\end{minipage}\hfill
\begin{minipage}[t]{0.49\textwidth}
\vspace{0pt}
\centering
\refstepcounter{figurepanel}\label{fig:h100-coupled-optimal-sparsity}
{\small\textbf{(\alph{figurepanel})} Optimal total-to-active ratio
$T/N_{\mathrm{opt}}$\par}
\smallskip
\includegraphics[width=0.98\linewidth,trim=0bp 4bp 10bp 20bp,clip]{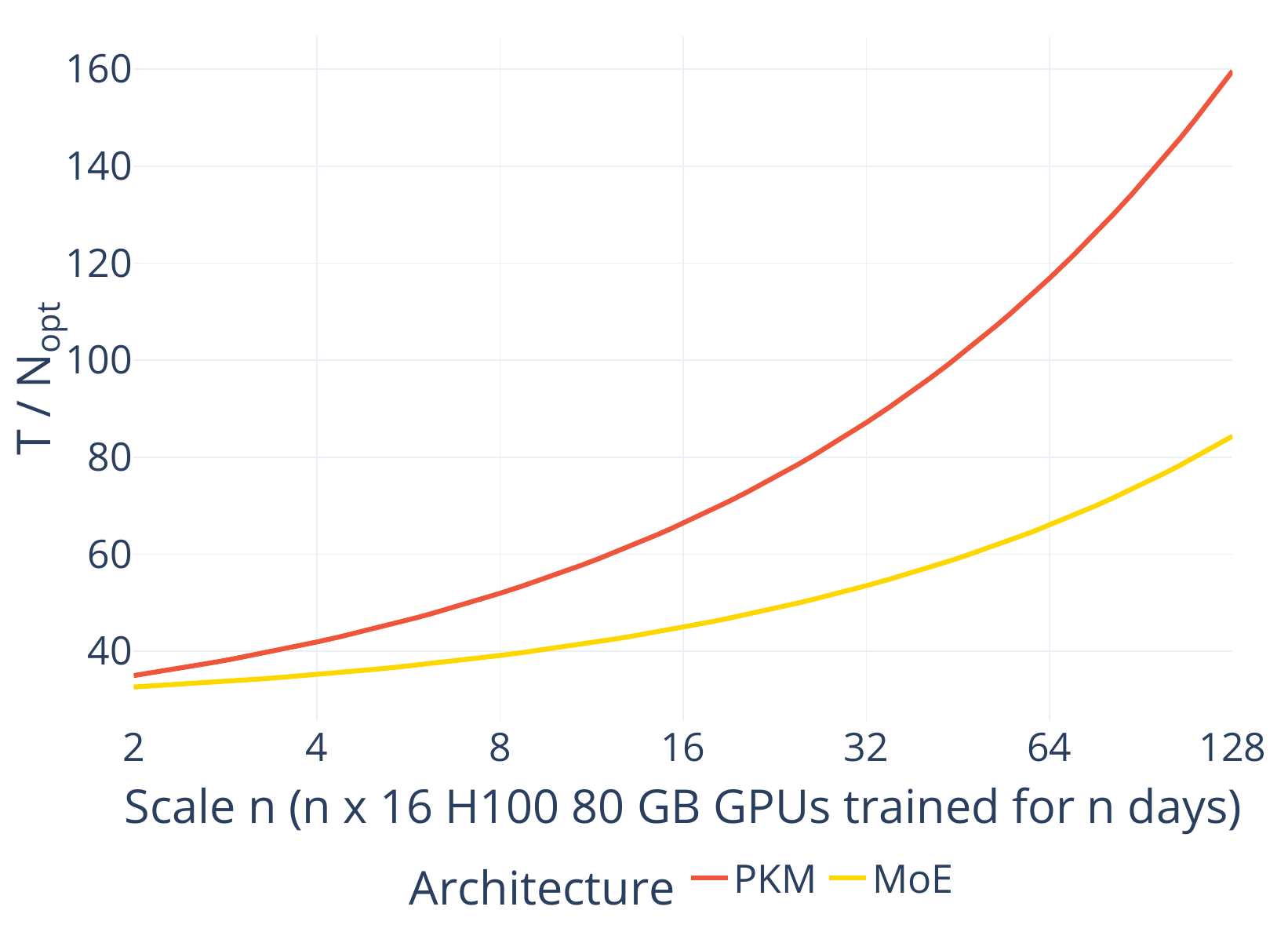}
\end{minipage}
\par\medskip
{\small\raggedright\textbf{Figure \thefigure.} \emph{Optimal loss and sparse
allocation along coupled GPU-and-time scaling.} Budgets follow
Equations~\ref{eq:coupled-resource-budgets} and~\ref{eq:coupled-training-operation-budget} ($16n$ H100 GPUs trained for $n$ days,
$D_{\mathrm{op}}(n)=n\times100$B tokens). (a) Lowest predicted loss $\widehat L_{\mathrm{opt}}(T(n),C(n))$ of each
architecture. (b) Optimal ratio $T(n)/N_{\mathrm{opt}}(T(n),C(n))$ of PKM and MoE, each optimized independently. Stability of the ranking under resampling: Appendix~\ref{sec:stability}. \par}
\end{minipage}
\end{center}

\subsection{Cost of Sparsity on Resource-Constrained Devices}
\label{sec:hardware-implications}

\begin{wrapfigure}[14]{r}{0.4\columnwidth}
\vspace{-\intextsep}
\centering
\refstepcounter{figure}\label{fig:plot-mobile-inference}
\includegraphics[width=\linewidth,trim=3.0bp 3.0bp 3.7bp 3.0bp,clip]{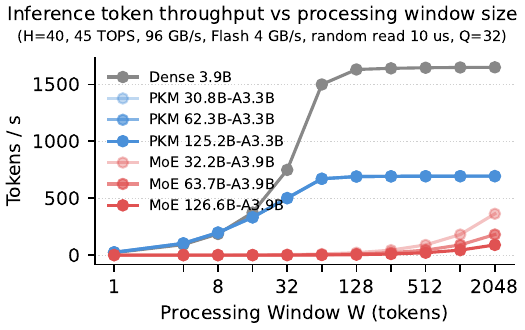}
\par\smallskip
{\small\raggedright\textbf{Figure \thefigure.}
\emph{Mobile inference: tokens/s versus processing window $W$.}
PKM (blue) and MoE (red) at $T/N\in\{8, 16, 32\}$;
dense (grey) for reference.\par}
\end{wrapfigure}

We model inference token throughput of MoE and PKM on a mobile device with limited DRAM and flash storage using a roofline model as described in Appendix~\ref{sec:appendix-hardware}.

On mobile devices, DRAM capacity caps $N$: the active-parameter count sets
the static memory footprint for inference. Scaling model capacity beyond this
budget pushes the additional parameters into flash (4\,GB/s, roughly $17\times$ slower than
DRAM). Both PKM and MoE are flash memory bound, but MoE requires orders of magnitude more flash traffic. For the approximately active-parameter-matched
configurations in Figure~\ref{fig:plot-mobile-inference} (3.3B for PKM and 3.93B for MoE), there is about a $260\times$
gap per distinct load ($78.6$M versus $0.3$M parameters; derivation in
Appendix~\ref{sec:appendix-hardware}). At $W{=}512$, the roofline model
estimates a $7.61\times$ PKM throughput speedup at expansion, $T/N{=}8$, increasing to $30.45\times$ at $T/N{=}32$. This increase
reflects PKM's capacity-independent retrieval cost, whereas MoE loads more
distinct experts as total capacity grows.

\section{Conclusion and Discussion}
\label{sec:discussion}

We developed scaling laws coupling active parameters, memory capacity, and
training data, and used them to derive resource-dependent allocation guidance.
The fitted comparisons favor PKM when capacity is abundant relative to compute,
while MoE and dense models become preferable in other regimes. Together with the
roofline analysis, these results motivate evaluating sparse architectures through
both loss scaling and memory-access costs.

These conclusions concern the architecture variants and training setup studied
here; predictions beyond the measured range require validation at larger scales.
Finer expert granularity~\citep{granularity}, memory-layer
expansion~\citep{ultramem}, or $n$-gram addressing~\citep{engram} could alter both
optimal sparsity and architecture rankings. Extending the analysis to these
designs and other training corpora is a natural next step.

Our compute approximation, $C\approx6ND$, omits memory access, routing,
communication, hardware utilization, and implementation efficiency. Equal FLOPs
therefore need not imply equal training time, especially at high sparsity.
The complementary roofline estimates depend on hardware and caching assumptions;
validating these predictions with end-to-end measurements is necessary to
establish when the predicted loss and throughput advantages translate into
practical gains.

\clearpage
\flushbottom

\bibliography{iclr2026_conference}
\bibliographystyle{iclr2026_conference}

\appendix
\clearpage
\addtocontents{toc}{\protect\setcounter{tocdepth}{2}}
\renewcommand{\contentsname}{Appendix Contents}
\tableofcontents
\setcounter{tocdepth}{3}
\section{Models and Training Hyperparameters}
\label{sec:model-training-details}

This appendix specifies the common backbone, architecture-specific modules,
and training configuration used for the dense Transformer, product-key memory (PKM),
Scaling Transformers with Embedding Modules (STEM), and mixture-of-experts
(MoE) experiments.

\subsection{Shared H-family backbone}

All four families use the same decoder-only Qwen3 backbone, parameterized by
an integer $H$.  Both the number of decoder blocks and the number of query
attention heads equal $H$.  The hidden width is $d=64H$, each attention head
has width 64, grouped-query attention uses $H/2$ key/value heads, and the
ordinary dense feed-forward width is
$d_{\mathrm{ff}}=4d=256H$.  For example, H12 has 12 blocks, $d=768$, and
$d_{\mathrm{ff}}=3072$, whereas H24 has 24 blocks, $d=1536$, and
$d_{\mathrm{ff}}=6144$.  The run names are H12, H16, H20, H24, and H32.
For a sparse model, this name identifies the shared backbone.

The GPT-NeoX vocabulary is padded from 50,257 to 50,432 model entries.  Input
embeddings and the language-model head are not tied.  Every block uses full
causal self-attention with rotary position embeddings
($\theta=10{,}000$), no sliding-window attention, and no attention dropout.
Qwen3 RMSNorm ($\epsilon=10^{-6}$) is applied before attention and before the
feed-forward module, with a final RMSNorm after the decoder stack; queries and
keys additionally receive per-head RMSNorm.  Attention uses FlashAttention~2.
The ordinary feed-forward modules are bias-free SwiGLUs,
\[
  \operatorname{FFN}(x)=W_{\mathrm{down}}
  \left[\operatorname{SiLU}(W_{\mathrm{gate}}x)\odot
  W_{\mathrm{up}}x\right].
\]

\subsection{Architecture-specific modules}

\paragraph{Dense Transformer.}
The dense model leaves the shared backbone unchanged: every block contains
the standard width-$4d$ SwiGLU above.

\paragraph{Product-key memory.}
PKM replaces the dense feed-forward module in zero-based blocks
$4,8,\ldots,H-4$.  This gives 2, 3, 4, 5, and 7 memory layers for H12, H16,
H20, H24, and H32, respectively; all other blocks retain the standard
width-$4d$ SwiGLU.  In a memory block, the learned lookup replaces the dense
up-projection:
\[
  \operatorname{PKM}(x)=W_{\mathrm{down}}
  \left[\operatorname{SiLU}(W_{\mathrm{gate}}x)\odot m(x)\right],
\]
where the gate and retrieved memory vector both have width $d$.  Each memory
block owns an independent value bank.

The depth-two router uses two product-key factors of width $b$.  A
factor pair indexes one physical bank containing
$b^2$ rows, each of width $d$.  The capacity sweep uses
$b\in\{448,704,1024,1536,2048,3072\}$, or 0.200704M, 0.495616M,
1.048576M, 2.359296M, 4.194304M, and 9.437184M records per memory layer.  If
$L_{\mathrm{mem}}$ is the number of memory layers, their value parameters
total $N_{\mathrm{mem}}=L_{\mathrm{mem}}b^2d$.

Routing uses four memory heads.  For each of the two factors, every head
projects the block input to width $d/2=32H$, applies BatchNorm to the projected
queries, and scores a shared table of $b$ RMS-normalized keys with scaled dot
products.  The router retains the best 32 keys from each factor, forms their
$32\times32$ Cartesian combinations, and keeps the best 32 complete addresses
per head.  A token therefore reads at most $4\times32=128$ value rows, and the
per-head softmax-weighted values are summed into $m(x)$.

\paragraph{Scaling Transformers with Embedding Modules.}
STEM uses the same memory-modulated feed-forward construction but replaces
learned routing by the deterministic lookup
$m(x)=\mathcal{M}[\texttt{token\_id}]$.

Each selected block owns a $V\times d_{\mathrm{ff}}$ table with
$V=50{,}432$, and one full-width row is read for each token.  The
context-dependent gate $\operatorname{SiLU}(W_{\mathrm{gate}}x)$ modulates
this context-independent token vector before the down-projection.  STEM layers
occupy zero-based indices $2,4,\ldots,H-2$, every second block except the first
and last, for a total of $H/2-1$ token-memory layers.  Retrieved vectors receive
no additional normalization.

\paragraph{Mixture of experts.}
MoE replaces the feed-forward module in every decoder block with token-choice
MoE.  Each expert is a complete bias-free SwiGLU of width
$d_{\mathrm{ff}}=4d$, so one active expert has the same feed-forward width as
the dense Transformer.  We sweep $E\in\{2,4,8,16\}$ experts per layer and always use top-1
routing; expert granularity is one, each layer holds $E$ experts, and there is
no shared expert.

\subsection{Data, optimization, and training schedule}

Models are trained from scratch on packed sequences of 2,048 tokens from the
DCLM-Edu corpus, using next-token cross-entropy and bfloat16 arithmetic.  The tokenizer uses the
GPT-NeoX vocabulary described above~\citep{black2022gptneox}.  Final
cross-entropy is evaluated on the same fixed 50M-token held-out window for
every model family.

Optimization uses AdamW with $\beta_1=0.9$, $\beta_2=0.95$, weight decay 0.1
for decay-eligible backbone parameters, and gradient-norm clipping at 1.0.
PKM router parameters and PKM/STEM value tables are excluded from weight
decay.  The learning-rate schedule is warmup--stable--decay.  Warmup duration,
in billions of tokens, equals the dense-equivalent model size in billions:
0.183, 0.355, 0.620, 1.004, and 2.219B tokens for H12, H16, H20, H24, and H32.
The stable phase is followed by a linear decay to zero over the final 20\% of
each run.  Multiple training horizons are obtained from a shared stable-phase
parent.

\subsection{Learning-rate selection}
\label{sec:learning-rate-selection}

For every model configuration, we tuned the peak learning rate on a base-2
logarithmic grid, $\eta=2^{x_{\mathrm{lr}}}$, and selected the value with the
lowest loss at a training budget of 16B tokens. The selected learning rate was
then used to train that configuration at all other token budgets. Training uses
a continuously increasing batch size
(Appendix~\ref{sec:batch-size-schedule}), which compensates for the increasing
token budget and allows all training endpoints of a model configuration to
share the same constant learning rate.

\subsection{Batch-size schedule}
\label{sec:batch-size-schedule}

The batch size increases continuously during training.  Step
Law~\citep{steplaw} finds the optimal batch size to depend primarily on the
number of training tokens and to be largely invariant to model size, and gives
$0.58\,D^{0.571}$ tokens as the optimal constant batch for a run of $D$
tokens.  We use this value as the target average batch per optimizer update
and, for every model configuration, set
\[
  B(D)=1.35\,D^{0.571},
\]
where $B$ is the batch size in tokens and $D$ is the current number of
training tokens.  A batch $c\,D^{a}$ processes $D$ tokens in
$D^{1-a}/\bigl(c\,(1-a)\bigr)$ updates, so its average batch per update is
$c\,(1-a)\,D^{a}$, and $c=0.58/(1-0.571)\approx1.35$ makes this average equal
to the Step Law value.  The batch is realized by gradient accumulation with a
fixed per-GPU microbatch and is rounded up to a multiple of 131,072 tokens (64
sequences).  The rounding raises the realized
average batch above the Step Law value, by 32\% for a 16B-token run and 14\%
for a 128B-token run.  Table~\ref{tab:batch-schedule} gives the realized
average batch at every training budget used in this paper, and
Table~\ref{tab:batch-steps} gives the numbers of training tokens at which the
batch increases.

\begin{center}
\begin{minipage}{\textwidth}
\centering
\refstepcounter{table}\label{tab:batch-schedule}
\textbf{Table \thetable.} \emph{Realized average batch size at the training budgets used in
this paper.}
\\{\small Batch sizes are in tokens. ``Average batch'' is the realized average batch per
optimizer update over a run of $D$ tokens, ``Step Law'' is the constant-batch value
$0.58\,D^{0.571}$, and ``Updates'' is the number of optimizer updates in the run.}%

\resizebox{0.43\textwidth}{!}{%
\begin{tabular}{rrrr}
\hline
$D$ (B tokens) & Average batch & Step Law & Updates \\
\hline
1 & 169,988 & 79,878 & 5,884 \\
2 & 211,062 & 118,663 & 9,476 \\
3 & 249,615 & 149,577 & 12,020 \\
4 & 278,177 & 176,280 & 14,380 \\
6 & 329,780 & 222,204 & 18,195 \\
8 & 375,569 & 261,874 & 21,302 \\
12 & 450,071 & 330,096 & 26,664 \\
16 & 514,249 & 389,028 & 31,114 \\
24 & 624,303 & 490,375 & 38,443 \\
32 & 717,989 & 577,921 & 44,569 \\
48 & 877,671 & 728,479 & 54,692 \\
64 & 1,015,108 & 858,533 & 63,049 \\
96 & 1,249,435 & 1,082,193 & 76,836 \\
128 & 1,450,765 & 1,275,396 & 88,230 \\
\hline
\end{tabular}%
}
\end{minipage}
\end{center}

\begin{center}
\begin{minipage}{\textwidth}
\centering
\refstepcounter{table}\label{tab:batch-steps}
\textbf{Table \thetable.} \emph{Training tokens at which the batch size increases.}
\\{\small The batch grows in steps of 131,072 tokens (64 sequences). Each pair gives the
number of training tokens $D$, in billions, from which the listed batch size in tokens is
used. At $D=1$B the batch is 262,144 tokens.}%

\resizebox{0.6\textwidth}{!}{%
\begin{tabular}{rr@{\qquad}rr@{\qquad}rr}
\hline
$D$ (B tokens) & Batch & $D$ (B tokens) & Batch & $D$ (B tokens) & Batch \\
\hline
1.8 & 393,216 & 25.4 & 1,310,720 & 69.6 & 2,228,224 \\
3.7 & 524,288 & 30.6 & 1,441,792 & 77.4 & 2,359,296 \\
6.1 & 655,360 & 36.1 & 1,572,864 & 85.6 & 2,490,368 \\
9.1 & 786,432 & 42.1 & 1,703,936 & 94.1 & 2,621,440 \\
12.5 & 917,504 & 48.4 & 1,835,008 & 103.0 & 2,752,512 \\
16.4 & 1,048,576 & 55.1 & 1,966,080 & 112.1 & 2,883,584 \\
20.7 & 1,179,648 & 62.2 & 2,097,152 & 121.7 & 3,014,656 \\
\hline
\end{tabular}%
}
\end{minipage}
\end{center}

\subsection{Measured model configurations}

\begin{center}
\begin{minipage}{\textwidth}
\centering
\textbf{Table \the\numexpr\value{table}+1\relax.} \emph{All 44 model configurations and their 421 measured endpoints in the
fitting dataset.}
\\{\small Parameters follow the non-embedding accounting of
Section~\ref{sec:model-families-and-variables}, with $N$ and $M$ in billions. \texttt{memc} is the
number of lookup records per memory layer in millions and \texttt{memv} their width. Each entry
of the last column is a measured $D/C$ pair with $C=6ND$. A dash means not applicable. STEM has
token-memory parameters, but its $M$ is dashed because memory capacity was not varied
independently.}%
\end{minipage}
\end{center}
\begingroup
\scriptsize
\setlength{\tabcolsep}{2.2pt}
\renewcommand{\arraystretch}{1.0}
\setlength{\LTpre}{0pt}
\begin{longtable}{@{}lrrrrrrr>{\raggedright\arraybackslash}p{0.365\textwidth}@{}}
\hline
Arch. & $H$ & $b$ & $E$ & \texttt{memc} & \texttt{memv} &
$N$ & $M$ & Measured $D/C$ (B/EF)\label{tab:model-configurations} \\
\hline
\endfirsthead
\multicolumn{9}{c}{\textbf{Table \thetable.} \emph{(continued)}}\\
\hline
Arch. & $H$ & $b$ & $E$ & \texttt{memc} & \texttt{memv} &
$N$ & $M$ & Measured $D/C$ (B/EF) \\
\hline
\endhead
\hline
\multicolumn{9}{r}{Continued on next page}\\
\endfoot
\hline
\endlastfoot
PKM & 12 & 448 & -- & 0.200704 & 768 & 0.100009 & 0.308281 & \mbox{2/1.200},\allowbreak\ \mbox{3/1.800},\allowbreak\ \mbox{4/2.400},\allowbreak\ \mbox{6/3.600},\allowbreak\ \mbox{8/4.800},\allowbreak\ \mbox{12/7.201},\allowbreak\ \mbox{16/9.601},\allowbreak\ \mbox{24/14.401},\allowbreak\ \mbox{32/19.202},\allowbreak\ \mbox{48/28.803},\allowbreak\ \mbox{64/38.403},\allowbreak\ \mbox{96/57.605} \\
PKM & 12 & 704 & -- & 0.495616 & 768 & 0.100402 & 0.761266 & \mbox{2/1.205},\allowbreak\ \mbox{3/1.807},\allowbreak\ \mbox{4/2.410},\allowbreak\ \mbox{6/3.614},\allowbreak\ \mbox{8/4.819},\allowbreak\ \mbox{12/7.229},\allowbreak\ \mbox{16/9.639},\allowbreak\ \mbox{24/14.458},\allowbreak\ \mbox{32/19.277},\allowbreak\ \mbox{48/28.916},\allowbreak\ \mbox{64/38.554},\allowbreak\ \mbox{96/57.832} \\
PKM & 12 & 1024 & -- & 1.048576 & 768 & 0.100894 & 1.610613 & \mbox{2/1.211},\allowbreak\ \mbox{3/1.816},\allowbreak\ \mbox{4/2.421},\allowbreak\ \mbox{6/3.632},\allowbreak\ \mbox{8/4.843},\allowbreak\ \mbox{12/7.264},\allowbreak\ \mbox{16/9.686},\allowbreak\ \mbox{24/14.529},\allowbreak\ \mbox{32/19.372},\allowbreak\ \mbox{48/29.057},\allowbreak\ \mbox{64/38.743},\allowbreak\ \mbox{96/58.115} \\
PKM & 12 & 1536 & -- & 2.359296 & 768 & 0.101680 & 3.623879 & \mbox{2/1.220},\allowbreak\ \mbox{3/1.830},\allowbreak\ \mbox{4/2.440},\allowbreak\ \mbox{6/3.660},\allowbreak\ \mbox{8/4.881},\allowbreak\ \mbox{12/7.321},\allowbreak\ \mbox{16/9.761},\allowbreak\ \mbox{24/14.642},\allowbreak\ \mbox{32/19.523},\allowbreak\ \mbox{48/29.284},\allowbreak\ \mbox{64/39.045},\allowbreak\ \mbox{96/58.568} \\
PKM & 12 & 2048 & -- & 4.194304 & 768 & 0.102467 & 6.442451 & \mbox{2/1.230},\allowbreak\ \mbox{3/1.844},\allowbreak\ \mbox{4/2.459},\allowbreak\ \mbox{6/3.689},\allowbreak\ \mbox{8/4.918},\allowbreak\ \mbox{12/7.378},\allowbreak\ \mbox{16/9.837},\allowbreak\ \mbox{24/14.755},\allowbreak\ \mbox{32/19.674},\allowbreak\ \mbox{48/29.510},\allowbreak\ \mbox{64/39.347} \\
PKM & 12 & 3072 & -- & 9.437184 & 768 & 0.104039 & 14.495515 & \mbox{4/2.497},\allowbreak\ \mbox{6/3.745},\allowbreak\ \mbox{8/4.994},\allowbreak\ \mbox{12/7.491},\allowbreak\ \mbox{16/9.988},\allowbreak\ \mbox{24/14.982},\allowbreak\ \mbox{32/19.976},\allowbreak\ \mbox{48/29.963},\allowbreak\ \mbox{64/39.951} \\
PKM & 16 & 448 & -- & 0.200704 & 1024 & 0.234615 & 0.616563 & \mbox{2/2.815},\allowbreak\ \mbox{3/4.223},\allowbreak\ \mbox{4/5.631},\allowbreak\ \mbox{6/8.446},\allowbreak\ \mbox{8/11.262},\allowbreak\ \mbox{12/16.892},\allowbreak\ \mbox{16/22.523},\allowbreak\ \mbox{24/33.785},\allowbreak\ \mbox{32/45.046},\allowbreak\ \mbox{48/67.569},\allowbreak\ \mbox{64/90.092} \\
PKM & 16 & 704 & -- & 0.495616 & 1024 & 0.235402 & 1.522532 & \mbox{2/2.825},\allowbreak\ \mbox{3/4.237},\allowbreak\ \mbox{4/5.650},\allowbreak\ \mbox{6/8.474},\allowbreak\ \mbox{8/11.299},\allowbreak\ \mbox{12/16.949},\allowbreak\ \mbox{16/22.599},\allowbreak\ \mbox{24/33.898},\allowbreak\ \mbox{32/45.197},\allowbreak\ \mbox{48/67.796},\allowbreak\ \mbox{64/90.394} \\
PKM & 16 & 1024 & -- & 1.048576 & 1024 & 0.236385 & 3.221225 & \mbox{2/2.837},\allowbreak\ \mbox{3/4.255},\allowbreak\ \mbox{4/5.673},\allowbreak\ \mbox{6/8.510},\allowbreak\ \mbox{8/11.346},\allowbreak\ \mbox{12/17.020},\allowbreak\ \mbox{16/22.693},\allowbreak\ \mbox{24/34.039},\allowbreak\ \mbox{32/45.386},\allowbreak\ \mbox{48/68.079},\allowbreak\ \mbox{64/90.772} \\
PKM & 16 & 1536 & -- & 2.359296 & 1024 & 0.237958 & 7.247757 & \mbox{2/2.855},\allowbreak\ \mbox{3/4.283},\allowbreak\ \mbox{4/5.711},\allowbreak\ \mbox{6/8.566},\allowbreak\ \mbox{8/11.422},\allowbreak\ \mbox{12/17.133},\allowbreak\ \mbox{16/22.844},\allowbreak\ \mbox{24/34.266},\allowbreak\ \mbox{32/45.688},\allowbreak\ \mbox{48/68.532},\allowbreak\ \mbox{64/91.376} \\
PKM & 16 & 2048 & -- & 4.194304 & 1024 & 0.239530 & 12.884902 & \mbox{4/5.749},\allowbreak\ \mbox{6/8.623},\allowbreak\ \mbox{8/11.497},\allowbreak\ \mbox{12/17.246},\allowbreak\ \mbox{16/22.995},\allowbreak\ \mbox{24/34.492},\allowbreak\ \mbox{32/45.990},\allowbreak\ \mbox{48/68.985},\allowbreak\ \mbox{64/91.980} \\
PKM & 20 & 448 & -- & 0.200704 & 1280 & 0.455246 & 1.027604 & \mbox{4/10.926},\allowbreak\ \mbox{6/16.389},\allowbreak\ \mbox{8/21.852},\allowbreak\ \mbox{12/32.778},\allowbreak\ \mbox{16/43.704},\allowbreak\ \mbox{24/65.555},\allowbreak\ \mbox{32/87.407},\allowbreak\ \mbox{48/131.111},\allowbreak\ \mbox{64/174.814} \\
PKM & 20 & 704 & -- & 0.495616 & 1280 & 0.456557 & 2.537554 & \mbox{4/10.957},\allowbreak\ \mbox{6/16.436},\allowbreak\ \mbox{8/21.915},\allowbreak\ \mbox{12/32.872},\allowbreak\ \mbox{16/43.829},\allowbreak\ \mbox{24/65.744},\allowbreak\ \mbox{32/87.659},\allowbreak\ \mbox{48/131.488},\allowbreak\ \mbox{64/175.318} \\
PKM & 20 & 1024 & -- & 1.048576 & 1280 & 0.458195 & 5.368709 & \mbox{4/10.997},\allowbreak\ \mbox{6/16.495},\allowbreak\ \mbox{8/21.993},\allowbreak\ \mbox{12/32.990},\allowbreak\ \mbox{16/43.987},\allowbreak\ \mbox{24/65.980},\allowbreak\ \mbox{32/87.973},\allowbreak\ \mbox{48/131.960},\allowbreak\ \mbox{64/175.947} \\
PKM & 20 & 1536 & -- & 2.359296 & 1280 & 0.460817 & 12.079596 & \mbox{4/11.060},\allowbreak\ \mbox{6/16.589},\allowbreak\ \mbox{8/22.119},\allowbreak\ \mbox{12/33.179},\allowbreak\ \mbox{16/44.238},\allowbreak\ \mbox{24/66.358},\allowbreak\ \mbox{32/88.477},\allowbreak\ \mbox{48/132.715},\allowbreak\ \mbox{64/176.954} \\
PKM & 20 & 2048 & -- & 4.194304 & 1280 & 0.463438 & 21.474836 & \mbox{8/22.245},\allowbreak\ \mbox{12/33.368},\allowbreak\ \mbox{16/44.490},\allowbreak\ \mbox{24/66.735},\allowbreak\ \mbox{32/88.980},\allowbreak\ \mbox{48/133.470},\allowbreak\ \mbox{64/177.960} \\
PKM & 24 & 448 & -- & 0.200704 & 1536 & 0.783135 & 1.541407 & \mbox{8/37.590},\allowbreak\ \mbox{12/56.386},\allowbreak\ \mbox{16/75.181},\allowbreak\ \mbox{24/112.771},\allowbreak\ \mbox{32/150.362},\allowbreak\ \mbox{48/225.543},\allowbreak\ \mbox{64/300.724} \\
PKM & 24 & 704 & -- & 0.495616 & 1536 & 0.785101 & 3.806331 & \mbox{8/37.685},\allowbreak\ \mbox{12/56.527},\allowbreak\ \mbox{16/75.370},\allowbreak\ \mbox{24/113.055},\allowbreak\ \mbox{32/150.739},\allowbreak\ \mbox{48/226.109},\allowbreak\ \mbox{64/301.479} \\
PKM & 24 & 1024 & -- & 1.048576 & 1536 & 0.787559 & 8.053064 & \mbox{8/37.803},\allowbreak\ \mbox{12/56.704},\allowbreak\ \mbox{16/75.606},\allowbreak\ \mbox{24/113.408},\allowbreak\ \mbox{32/151.211},\allowbreak\ \mbox{48/226.817},\allowbreak\ \mbox{64/302.423} \\
PKM & 24 & 1536 & -- & 2.359296 & 1536 & 0.791491 & 18.119393 & \mbox{8/37.992},\allowbreak\ \mbox{12/56.987},\allowbreak\ \mbox{16/75.983},\allowbreak\ \mbox{24/113.975},\allowbreak\ \mbox{32/151.966},\allowbreak\ \mbox{48/227.949},\allowbreak\ \mbox{64/303.932} \\
\hline
MoE & 12 & -- & 2 & -- & -- & 0.106207 & 0.084935 & \mbox{2/1.274},\allowbreak\ \mbox{3/1.912},\allowbreak\ \mbox{4/2.549},\allowbreak\ \mbox{6/3.823},\allowbreak\ \mbox{8/5.098},\allowbreak\ \mbox{12/7.647},\allowbreak\ \mbox{16/10.196},\allowbreak\ \mbox{24/15.294},\allowbreak\ \mbox{32/20.392},\allowbreak\ \mbox{48/30.588},\allowbreak\ \mbox{64/40.784} \\
MoE & 12 & -- & 4 & -- & -- & 0.106226 & 0.254804 & \mbox{2/1.275},\allowbreak\ \mbox{3/1.912},\allowbreak\ \mbox{4/2.549},\allowbreak\ \mbox{6/3.824},\allowbreak\ \mbox{8/5.099},\allowbreak\ \mbox{12/7.648},\allowbreak\ \mbox{16/10.198},\allowbreak\ \mbox{24/15.297},\allowbreak\ \mbox{32/20.395},\allowbreak\ \mbox{48/30.593},\allowbreak\ \mbox{64/40.791},\allowbreak\ \mbox{96/61.186} \\
MoE & 12 & -- & 8 & -- & -- & 0.106263 & 0.594543 & \mbox{2/1.275},\allowbreak\ \mbox{3/1.913},\allowbreak\ \mbox{4/2.550},\allowbreak\ \mbox{6/3.825},\allowbreak\ \mbox{8/5.101},\allowbreak\ \mbox{12/7.651},\allowbreak\ \mbox{16/10.201},\allowbreak\ \mbox{24/15.302},\allowbreak\ \mbox{32/20.402},\allowbreak\ \mbox{48/30.604},\allowbreak\ \mbox{64/40.805},\allowbreak\ \mbox{96/61.207} \\
MoE & 12 & -- & 16 & -- & -- & 0.106337 & 1.274020 & \mbox{2/1.276},\allowbreak\ \mbox{3/1.914},\allowbreak\ \mbox{4/2.552},\allowbreak\ \mbox{6/3.828},\allowbreak\ \mbox{8/5.104},\allowbreak\ \mbox{12/7.656},\allowbreak\ \mbox{16/10.208},\allowbreak\ \mbox{24/15.312},\allowbreak\ \mbox{32/20.417},\allowbreak\ \mbox{48/30.625},\allowbreak\ \mbox{64/40.833},\allowbreak\ \mbox{96/61.250} \\
MoE & 16 & -- & 2 & -- & -- & 0.251727 & 0.201327 & \mbox{2/3.021},\allowbreak\ \mbox{3/4.531},\allowbreak\ \mbox{4/6.041},\allowbreak\ \mbox{6/9.062},\allowbreak\ \mbox{8/12.083},\allowbreak\ \mbox{12/18.124},\allowbreak\ \mbox{16/24.166},\allowbreak\ \mbox{24/36.249},\allowbreak\ \mbox{32/48.332} \\
MoE & 16 & -- & 4 & -- & -- & 0.251760 & 0.603980 & \mbox{2/3.021},\allowbreak\ \mbox{3/4.532},\allowbreak\ \mbox{4/6.042},\allowbreak\ \mbox{6/9.063},\allowbreak\ \mbox{8/12.084},\allowbreak\ \mbox{12/18.127},\allowbreak\ \mbox{16/24.169},\allowbreak\ \mbox{24/36.253},\allowbreak\ \mbox{32/48.338},\allowbreak\ \mbox{48/72.507},\allowbreak\ \mbox{64/96.676} \\
MoE & 16 & -- & 8 & -- & -- & 0.251825 & 1.409286 & \mbox{2/3.022},\allowbreak\ \mbox{3/4.533},\allowbreak\ \mbox{4/6.044},\allowbreak\ \mbox{6/9.066},\allowbreak\ \mbox{8/12.088},\allowbreak\ \mbox{12/18.131},\allowbreak\ \mbox{16/24.175},\allowbreak\ \mbox{24/36.263},\allowbreak\ \mbox{32/48.350},\allowbreak\ \mbox{48/72.526},\allowbreak\ \mbox{64/96.701} \\
MoE & 16 & -- & 16 & -- & -- & 0.251956 & 3.019899 & \mbox{16/24.188},\allowbreak\ \mbox{24/36.282},\allowbreak\ \mbox{32/48.376},\allowbreak\ \mbox{48/72.563},\allowbreak\ \mbox{64/96.751} \\
MoE & 20 & -- & 2 & -- & -- & 0.491626 & 0.393216 & \mbox{4/11.799},\allowbreak\ \mbox{6/17.699},\allowbreak\ \mbox{8/23.598},\allowbreak\ \mbox{12/35.397},\allowbreak\ \mbox{16/47.196},\allowbreak\ \mbox{24/70.794},\allowbreak\ \mbox{32/94.392},\allowbreak\ \mbox{48/141.588} \\
MoE & 20 & -- & 4 & -- & -- & 0.491677 & 1.179648 & \mbox{4/11.800},\allowbreak\ \mbox{6/17.700},\allowbreak\ \mbox{8/23.601},\allowbreak\ \mbox{12/35.401},\allowbreak\ \mbox{16/47.201},\allowbreak\ \mbox{24/70.802},\allowbreak\ \mbox{32/94.402},\allowbreak\ \mbox{48/141.603} \\
MoE & 20 & -- & 8 & -- & -- & 0.491780 & 2.752512 & \mbox{4/11.803},\allowbreak\ \mbox{6/17.704},\allowbreak\ \mbox{8/23.605},\allowbreak\ \mbox{12/35.408},\allowbreak\ \mbox{16/47.211},\allowbreak\ \mbox{24/70.816},\allowbreak\ \mbox{32/94.422},\allowbreak\ \mbox{48/141.633} \\
MoE & 20 & -- & 16 & -- & -- & 0.491985 & 5.898240 & \mbox{16/47.231},\allowbreak\ \mbox{24/70.846},\allowbreak\ \mbox{32/94.461},\allowbreak\ \mbox{48/141.692},\allowbreak\ \mbox{64/188.922} \\
MoE & 24 & -- & 2 & -- & -- & 0.849499 & 0.679477 & \mbox{8/40.776} \\
MoE & 24 & -- & 4 & -- & -- & 0.849572 & 2.038432 & \mbox{8/40.779},\allowbreak\ \mbox{12/61.169},\allowbreak\ \mbox{16/81.559},\allowbreak\ \mbox{24/122.338},\allowbreak\ \mbox{32/163.118} \\
MoE & 24 & -- & 8 & -- & -- & 0.849720 & 4.756341 & \mbox{8/40.787},\allowbreak\ \mbox{12/61.180},\allowbreak\ \mbox{16/81.573},\allowbreak\ \mbox{24/122.360},\allowbreak\ \mbox{32/163.146},\allowbreak\ \mbox{48/244.719} \\
\hline
Dense & 12 & -- & -- & -- & -- & 0.106189 & 0 & \mbox{1/0.637},\allowbreak\ \mbox{2/1.274},\allowbreak\ \mbox{3/1.911},\allowbreak\ \mbox{4/2.549},\allowbreak\ \mbox{6/3.823},\allowbreak\ \mbox{8/5.097},\allowbreak\ \mbox{12/7.646},\allowbreak\ \mbox{16/10.194},\allowbreak\ \mbox{24/15.291},\allowbreak\ \mbox{32/20.388},\allowbreak\ \mbox{48/30.582},\allowbreak\ \mbox{64/40.777},\allowbreak\ \mbox{96/61.165} \\
Dense & 16 & -- & -- & -- & -- & 0.251694 & 0 & \mbox{1/1.510},\allowbreak\ \mbox{2/3.020},\allowbreak\ \mbox{3/4.530},\allowbreak\ \mbox{4/6.041},\allowbreak\ \mbox{6/9.061},\allowbreak\ \mbox{8/12.081},\allowbreak\ \mbox{12/18.122},\allowbreak\ \mbox{16/24.163},\allowbreak\ \mbox{24/36.244},\allowbreak\ \mbox{32/48.325},\allowbreak\ \mbox{48/72.488},\allowbreak\ \mbox{64/96.651},\allowbreak\ \mbox{96/144.976} \\
Dense & 20 & -- & -- & -- & -- & 0.491575 & 0 & \mbox{2/5.899},\allowbreak\ \mbox{3/8.848},\allowbreak\ \mbox{4/11.798},\allowbreak\ \mbox{6/17.697},\allowbreak\ \mbox{8/23.596},\allowbreak\ \mbox{12/35.393},\allowbreak\ \mbox{16/47.191},\allowbreak\ \mbox{24/70.787},\allowbreak\ \mbox{32/94.382},\allowbreak\ \mbox{48/141.574},\allowbreak\ \mbox{64/188.765},\allowbreak\ \mbox{96/283.147} \\
Dense & 24 & -- & -- & -- & -- & 0.849425 & 0 & \mbox{3/15.290},\allowbreak\ \mbox{4/20.386},\allowbreak\ \mbox{6/30.579},\allowbreak\ \mbox{8/40.772},\allowbreak\ \mbox{12/61.159},\allowbreak\ \mbox{16/81.545},\allowbreak\ \mbox{24/122.317},\allowbreak\ \mbox{32/163.090},\allowbreak\ \mbox{48/244.634},\allowbreak\ \mbox{64/326.179} \\
Dense & 32 & -- & -- & -- & -- & 2.013403 & 0 & \mbox{4/48.322},\allowbreak\ \mbox{6/72.483},\allowbreak\ \mbox{8/96.643},\allowbreak\ \mbox{12/144.965},\allowbreak\ \mbox{16/193.287},\allowbreak\ \mbox{24/289.930},\allowbreak\ \mbox{32/386.573},\allowbreak\ \mbox{48/579.860},\allowbreak\ \mbox{64/773.147} \\
\hline
STEM & 12 & -- & -- & 0.050432 & 3072 & 0.094408 & -- & \mbox{1/0.566},\allowbreak\ \mbox{2/1.133},\allowbreak\ \mbox{3/1.699},\allowbreak\ \mbox{4/2.266},\allowbreak\ \mbox{6/3.399},\allowbreak\ \mbox{8/4.532},\allowbreak\ \mbox{12/6.797},\allowbreak\ \mbox{16/9.063},\allowbreak\ \mbox{24/13.595},\allowbreak\ \mbox{32/18.126},\allowbreak\ \mbox{48/27.189},\allowbreak\ \mbox{64/36.253},\allowbreak\ \mbox{96/54.379},\allowbreak\ \mbox{128/72.505} \\
STEM & 16 & -- & -- & 0.050432 & 4096 & 0.222363 & -- & \mbox{1/1.334},\allowbreak\ \mbox{2/2.668},\allowbreak\ \mbox{3/4.003},\allowbreak\ \mbox{4/5.337},\allowbreak\ \mbox{6/8.005},\allowbreak\ \mbox{8/10.673},\allowbreak\ \mbox{12/16.010},\allowbreak\ \mbox{16/21.347},\allowbreak\ \mbox{24/32.020},\allowbreak\ \mbox{32/42.694},\allowbreak\ \mbox{48/64.040},\allowbreak\ \mbox{64/85.387},\allowbreak\ \mbox{96/128.081} \\
STEM & 20 & -- & -- & 0.050432 & 5120 & 0.432639 & -- & \mbox{2/5.192},\allowbreak\ \mbox{3/7.787},\allowbreak\ \mbox{4/10.383},\allowbreak\ \mbox{6/15.575},\allowbreak\ \mbox{8/20.767},\allowbreak\ \mbox{12/31.150},\allowbreak\ \mbox{16/41.533},\allowbreak\ \mbox{24/62.300},\allowbreak\ \mbox{32/83.067},\allowbreak\ \mbox{48/124.600},\allowbreak\ \mbox{64/166.133} \\
STEM & 24 & -- & -- & 0.050432 & 6144 & 0.745683 & -- & \mbox{3/13.422},\allowbreak\ \mbox{4/17.896},\allowbreak\ \mbox{6/26.845},\allowbreak\ \mbox{8/35.793},\allowbreak\ \mbox{12/53.689},\allowbreak\ \mbox{16/71.586},\allowbreak\ \mbox{24/107.378},\allowbreak\ \mbox{32/143.171},\allowbreak\ \mbox{48/214.757},\allowbreak\ \mbox{64/286.342} \\
\end{longtable}
\endgroup

\section{Fitting scaling laws functions}
\label{sec:fitting-scaling-laws}

\subsection{Optimization of Formula Coefficients}
For all four architectures, coefficients are fitted independently by nonlinear
least squares. Throughout the fitting procedure,
$N$, $D$, $T$, and $M$ are fixed-unit raw counts, and $\log$ denotes the
natural logarithm. For observation $i$, with predictors
$\mathbf{x}_i=(N_i,D_i)$ or $(N_i,D_i,M_i)$, the optimized residual is
\[
r_i(\boldsymbol{\theta})
=\sqrt{w_i}\left[
\log \widehat L(\mathbf{x}_i;\boldsymbol{\theta})-\log L_i
\right].
\]
The weight $w_i$ is proportional to the inverse number of observations in the
corresponding architecture lineage and is normalized to have mean one. Thus,
every lineage contributes equal total weight even when token-budget coverage
differs across lineages. Full-data fits use the soft-$L_1$ loss with
$f_{\mathrm{scale}}=10^{-3}$; validation fits use the Huber loss with the same
scale. Optimization is performed in log-loss space.

The scale variables $N$, $D$, and $M$ are observed inputs and are not optimized.
For the PKM and MoE three-variable fits, the positive amplitudes are represented
as $A=\exp(a)$ and $B=\exp(b)$, where $a$ and $b$ are unconstrained optimizer
coordinates. The positive exponents are represented as
$\alpha=\operatorname{softplus}(a_{\alpha})+\epsilon$,
$\beta=\operatorname{softplus}(a_{\beta})+\epsilon$, and
$\kappa=\operatorname{softplus}(a_{\kappa})+\epsilon$, with
$\operatorname{softplus}(x)=\log(1+\exp x)$ and $\epsilon=10^{-8}$. The
interaction coefficients $\lambda$, $\mu$, $\zeta$, and $\xi$ are optimized
directly and may have either sign. The Dense Transformer and STEM
fits instead use exponential transforms for $A$, $B$, $\alpha$,
and $\beta$, while retaining signed $\lambda$ and $\xi$.

The Dense Transformer and STEM fits use the same robust least-squares
objective, with the auxiliary coordinates
\[
a=\log A,\qquad b=\log B,\qquad
a_{\alpha}=\log\alpha,\qquad a_{\beta}=\log\beta,
\]
with bounds
\[
a,b\in[-60,100],\qquad
a_{\alpha},a_{\beta}\in[-6,3],
\]
\[
\lambda\in[-5,5],\qquad \xi\in[-0.5,0.5].
\]

\subsubsection{Capacity--Token Coupling and Local Responses}
\label{sec:capacity-token-analysis}

The parameterization introduces coupling between the amplitude and exponent
coefficients. In particular, the data-dependent term can be written
equivalently as
\[
\frac{B N^{-\lambda}M^{-\mu}}
{D^{\beta+\zeta\log M+\xi\log N}}
=B D^{-\beta}
N^{-(\lambda+\xi\log D)}
M^{-(\mu+\zeta\log D)}.
\]
Therefore, at a given token budget, the effective exponent of $N$ is
$\lambda+\xi\log D$, while the effective exponent of $M$ is
$\mu+\zeta\log D$. Over a finite range of $D$, an increase in $\lambda$ can be
partly offset by a decrease in $\xi$, and the same compensation can occur
between $\mu$ and $\zeta$, without substantially changing the predictions.
Consequently, individual fitted coefficients should not be interpreted as
independent causal effects. We instead evaluate the combined amplitude,
$q_D(N,M)$, effective exponents, local scaling responses, and held-out
predictions. The fitted full-data coefficients of the reported laws are shown
in Table~\ref{tab:selected-law-coefficients}. PKM and MoE use the selected P6F law
(Equation~\ref{eq:memory-interaction-law};
Appendix~\ref{sec:fitted-formulas}); the joint PKM--MoE fit uses its P6F
working reference; and the Dense Transformer and STEM use vP6F
(Equation~\ref{eq:two-variable-interaction-law}). The table includes every learned amplitude,
exponent, and interaction parameter. A dash indicates that
the parameter is absent from the two-variable law.

\begin{center}
\begin{minipage}{\textwidth}
\centering
\refstepcounter{table}\label{tab:selected-law-coefficients}
\textbf{Table \thetable.} \emph{Full-data fitted coefficients of the reported scaling laws.}

\resizebox{\textwidth}{!}{%
\begin{tabular}{llccccccccc}
\hline
Architecture & Formula & $A$ & $B$ & $\alpha$ & $\beta$ & $\kappa$ & $\lambda$ & $\mu$ & $\zeta$ & $\xi$ \\
\hline
PKM & P6F & $1.96\mathrm{e}1$ & $2.02\mathrm{e}4$ & $9.29\mathrm{e}{-2}$ & $6.36\mathrm{e}{-1}$ & $1.60\mathrm{e}{-2}$ & $5.41\mathrm{e}{-1}$ & $-2.63\mathrm{e}{-1}$ & $1.07\mathrm{e}{-2}$ & $-3.19\mathrm{e}{-2}$ \\
MoE & P6F & $1.42\mathrm{e}1$ & $7.30\mathrm{e}3$ & $7.47\mathrm{e}{-2}$ & $5.51\mathrm{e}{-1}$ & $1.73\mathrm{e}{-2}$ & $2.94\mathrm{e}{-1}$ & $-9.86\mathrm{e}{-2}$ & $2.34\mathrm{e}{-3}$ & $-1.77\mathrm{e}{-2}$ \\
Joint PKM and MoE & P6F & $1.79\mathrm{e}1$ & $6.12\mathrm{e}4$ & $9.29\mathrm{e}{-2}$ & $6.72\mathrm{e}{-1}$ & $1.20\mathrm{e}{-2}$ & $5.80\mathrm{e}{-1}$ & $-2.50\mathrm{e}{-1}$ & $1.08\mathrm{e}{-2}$ & $-3.36\mathrm{e}{-2}$ \\
Dense Transformer & vP6F & $1.32\mathrm{e}1$ & $6.35\mathrm{e}2$ & $8.49\mathrm{e}{-2}$ & $4.71\mathrm{e}{-1}$ & -- & $1.11\mathrm{e}{-2}$ & -- & -- & $-8.00\mathrm{e}{-3}$ \\
STEM & vP6F & $1.16\mathrm{e}1$ & $1.08\mathrm{e}3$ & $8.03\mathrm{e}{-2}$ & $4.80\mathrm{e}{-1}$ & -- & $5.85\mathrm{e}{-2}$ & -- & -- & $-9.46\mathrm{e}{-3}$ \\
\hline
\end{tabular}%
}
\end{minipage}
\end{center}

All numerical ranges below evaluate the single PKM coefficient vector in
Table~\ref{tab:selected-law-coefficients} over the observed configurations;
they are not uncertainty intervals.

The finite-model component is
\[
U(N,M)=A N^{-\alpha}M^{-\kappa}.
\]
Because $\alpha>0$ and $\kappa>0$, $U$ decreases monotonically with either
$N$ or $M$ when the other coordinate is fixed. Across the 20 observed
$(N,M)$ configurations, $U$ ranges from 1.994 to 2.580. Its minimum occurs at
$N=791.5$ million and $M=18.119$ billion, whereas its maximum occurs at
$N=100.0$ million and $M=0.308$ billion. Doubling either capacity coordinate
gives
\[
\frac{U(2N,M)}{U(N,M)}=2^{-\alpha}=0.9376,
\qquad
\frac{U(N,2M)}{U(N,M)}=2^{-\kappa}=0.9889.
\]
Thus, within this fitted component, doubling $N$ reduces $U$ by 6.24\%, while
doubling $M$ reduces it by 1.11\%. These percentages concern $U$ alone, not
total predicted loss.

For the data-dependent component, define
\[
V_0(N,M)=B N^{-\lambda}M^{-\mu},
\qquad
V(N,D,M)=V_0(N,M)D^{-q_D(N,M)}.
\]
Across the same 20 configurations, $V_0(N,M)$ ranges from 82.1 to 441.4, a
5.4-fold variation in the prefactor of $V$, while $q_D(N,M)$ ranges from 0.211
at $N=783.1$ million and $M=1.541$ billion to 0.299 at $N=104.0$ million and
$M=14.496$ billion. Increasing $N$ lowers both $V_0$ and $q_D$, whereas
increasing $M$ raises both. Their effects on $V$ oppose each other: $V_0$ is the
multiplicative prefactor of $D^{-q_D}$, while a larger $q_D$ makes
$D^{-q_D}$ smaller. At fixed $D$, the net dependence of $V$ on $N$ and $M$
is therefore governed by the effective exponents
$\lambda+\xi\log D$ and $\mu+\zeta\log D$, respectively.

At fixed supported $N$ and $M$, doubling the training budget gives
\[
\frac{V(N,2D,M)}{V(N,D,M)}=2^{-q_D(N,M)}.
\]
The proportional reduction is therefore
$R_D(N,M)=1-2^{-q_D(N,M)}$, with
$\partial R_D/\partial q_D=(\log 2)2^{-q_D}>0$. Evaluating the endpoints of
the fitted $q_D$ range gives reductions of 13.6\% and 18.7\%.

To measure how doubling $N$ or $M$ changes this percentage benefit of doubling
$D$, we compare the corresponding reductions $R_D$. Doubling either capacity
coordinate changes $q_D$ according to
\[
q_D(2N,M)=q_D(N,M)+\xi\log 2,
\qquad
q_D(N,2M)=q_D(N,M)+\zeta\log 2.
\]
The relative changes in the data-doubling benefit are therefore
\[
\frac{R_D(2N,M)}{R_D(N,M)}
=\frac{1-2^{-[q_D(N,M)+\xi\log 2]}}
{1-2^{-q_D(N,M)}},
\]
\[
\frac{R_D(N,2M)}{R_D(N,M)}
=\frac{1-2^{-[q_D(N,M)+\zeta\log 2]}}
{1-2^{-q_D(N,M)}}.
\]
With the fitted coefficients, the first ratio ranges from 0.902 to 0.933:
doubling $N$ makes the reduction in $V$ obtained from doubling $D$
approximately 6.7--9.8\% smaller. The second ratio ranges from 1.022 to 1.033:
doubling $M$ makes that reduction approximately 2.2--3.3\% larger. Each is a
range because the same fixed change in $q_D$ is applied to different initial
$q_D(N,M)$ values across the observed configurations. These ratios compare
percentage reductions in $V$ and do not describe changes in total predicted
loss. Their directions are consistent across all 47 validation fits: every
fitted $\xi$ is negative and every fitted $\zeta$ is positive.

Finally, the response of total predicted loss $\widehat L=U+V$ depends on the
relative sizes and responses of both components. Its local log-log responses
are
\[
\mathcal E_N
=-\frac{\partial\log\widehat L}{\partial\log N}
=\frac{\alpha U+(\lambda+\xi\log D)V}{U+V},
\]
\[
\mathcal E_M
=-\frac{\partial\log\widehat L}{\partial\log M}
=\frac{\kappa U+(\mu+\zeta\log D)V}{U+V},
\qquad
\mathcal E_D
=-\frac{\partial\log\widehat L}{\partial\log D}
=\frac{q_DV}{U+V}.
\]
In the fitted decomposition, $U$ constitutes 73.7--91.7\% of predicted loss
over the observed data. Evaluating the single full-data fit of the selected PKM
scaling law at all 192 observed $(N,D,M)$ endpoints, which span
$N=100.0$--$791.5$ million active non-embedding parameters,
$M=0.308$--$21.475$ billion inactive parameters, and $D=2$--$96$ billion
training tokens, total predicted loss decreases locally as any of $N$, $M$, or
$D$ increases. These are the marginal ranges of the observed coordinates; the
192 endpoints do not form a complete Cartesian grid over them. The following
ranges are the minimum and maximum responses across those endpoints, not
uncertainty intervals or ranges across validation fits. Holding the other
coordinates fixed, a small 1\% increase
reduces predicted loss by approximately 0.020--0.063\% for $N$,
0.003--0.015\% for $M$, and 0.022--0.069\% for $D$.
Thus, the fitted loss is less locally sensitive to additional memory capacity
than to active parameters or training tokens. These quantities are local
sensitivities of the fitted function, not finite-doubling effects or causal
estimates.

\subsection{Evaluation of Fitted Formulas}
To assess both goodness of fit and predictive generalization, we evaluate each
fitted scaling law on the complete dataset and under a suite of held-out
validation tests. Each endpoint is one measured final cross-entropy value
$L_i$, associated with scale coordinates $(N_i,D_i)$ or $(N_i,D_i,M_i)$.

For held-out evaluation, each validation fold $k$ partitions the observed
endpoints into a training set $\mathcal T_k$ and a disjoint test set
$\mathcal S_k$. The formula coefficients are refitted using only
$\mathcal T_k$, after which prediction error is evaluated only on
$\mathcal S_k$. For any evaluation set $\mathcal S$, let
$e_i=\widehat L_i-L_i$ and let $\widetilde w_i(\mathcal S)$ denote the
normalized reporting weight of endpoint $i$, such that
$\sum_{i\in\mathcal S}\widetilde w_i(\mathcal S)=1$. We report
\[
\operatorname{MAE}(\mathcal S)
=\sum_{i\in\mathcal S}\widetilde w_i(\mathcal S)|e_i|,
\qquad
\operatorname{RMSE}(\mathcal S)
=\sqrt{\sum_{i\in\mathcal S}\widetilde w_i(\mathcal S)e_i^2}.
\]
Lower values are better. MAE measures the typical absolute prediction error,
whereas RMSE gives greater weight to large errors. Both are evaluated in the
original cross-entropy scale. For PKM and MoE, let $\mathcal G(\mathcal S)$ be the set
of model lineages represented in $\mathcal S$, let $g(i)$ be the lineage of
endpoint $i$, and let
$n_g(\mathcal S)=|\{i\in\mathcal S:g(i)=g\}|$. The reporting weights are
\[
\widetilde w_i(\mathcal S)
=\frac{1}{|\mathcal G(\mathcal S)|\,n_{g(i)}(\mathcal S)}.
\]
Each represented sparse-model lineage therefore contributes equal total mass within
the evaluation set, irrespective of how many token-budget endpoints it
contains. For the Dense Transformer and STEM tables,
$\widetilde w_i(\mathcal S)=1/|\mathcal S|$, so the reported metrics reduce to
ordinary endpoint-wise MAE and RMSE.

For a validation family containing $K$ folds, we report the arithmetic means
of the fold-level errors,
\[
\operatorname{MAE}_{\mathrm{family}}
=\frac{1}{K}\sum_{k=1}^{K}\operatorname{MAE}(\mathcal S_k),
\qquad
\operatorname{RMSE}_{\mathrm{family}}
=\frac{1}{K}\sum_{k=1}^{K}\operatorname{RMSE}(\mathcal S_k),
\]
over all declared folds. Folds whose refits fail a validity check below are
included in these averages, and the resulting family value is flagged in the tables.

Fit validity is determined by fixed numerical and structural checks, not by
the magnitude of MAE or RMSE. Every invalid P3-or-later result reported in the
PKM and MoE tables is explained by one or both of two checks: (i) the optimizer's
first-order optimality measure for the retained best-objective fit is not
finite or exceeds $10^{-3}$; and (ii) increasing either $N$ or $D$ causes
predicted loss to increase beyond numerical tolerance somewhere within the
observed coordinate range.

For every table in this section, a family metric is computed when every
declared fold has finite test-set predictions and finite MAE and RMSE. A fold
whose refit fails a validity check above still contributes its errors, and the
family value is then flagged by a subscript that names the failed check.
A large prediction error does not by itself make a fit invalid. Likewise, a large
condition number, strong parameter correlation, bootstrap instability, or failure
to improve a parent formula's held-out error is diagnostic evidence but does not
alone flag a table entry.

For the validation definitions below, capacity refers to the number of inactive
parameters $M$. A capacity family groups corresponding sparse-capacity
configurations across active-model sizes, although their exact values of $M$
may differ. A model lineage contains all token-budget endpoints associated with
one fixed observed $(N,M)$ configuration. We report the following full-data
and held-out evaluations:

\emph{Full.} ``Full'' denotes the in-sample result obtained by fitting and
evaluating on the complete dataset. Every other reported evaluation is a
genuinely held-out result from a refitted validation fold or the mean across
all folds in that family.

\emph{Leave lowest losses out (LLLO$_p$).} For a held-out percentage $p$, all
$n$ endpoints are ordered from lowest to highest observed cross-entropy. If two
endpoints have exactly the same loss, a fixed endpoint identifier is used only
to make their ordering reproducible. The lowest-loss
$\lceil (p/100)n\rceil$ endpoints form the test set, and the complementary
higher-loss endpoints form the training set; each partition is therefore
disjoint and exhaustive. We evaluate $p\in\{10,20,30,40,50\}$ and report the
five LLLO$_p$ tests separately. Increasing $p$ measures progressively longer
extrapolation from worse observed losses toward the best observed endpoints.
Because LLLO$_p$ is an endpoint-level split, its training and test sets may
contain endpoints from the same model lineage.

\emph{Leave one model out (LOMO).} Each fold removes one complete lineage,
refits on all other lineages, and predicts every endpoint of the excluded
model configuration.

\emph{Leave one size out (LOSO).} Each fold removes one complete active-model-
size family, represented by its measured non-embedding active-parameter counts
$N$, refits on the remaining size families, and predicts the excluded family.

\emph{Leave one capacity out (LOCO).} Each fold removes one complete capacity
family for inactive parameters across all available active-model sizes and
token budgets. A capacity family can correspond to different absolute values
of $M$ at different active-model sizes; all reported capacities use the
measured values of $M$.

\emph{Capacity-forward validation (CF).} The observed capacity families for
inactive parameters are ordered from smallest to largest. For each target after
the first two levels, the formula is fitted using only lower-capacity families
and evaluated at the target family. This tests forward extrapolation in $M$.

\emph{Future-token validation (FT).} We first identify token budgets observed
in at least 80\% of the model lineages and select the two largest eligible
budgets. The formula is fitted on all rows below the smaller selected budget
and evaluated jointly on the two selected tail budgets.

\emph{Unseen-capacity future-token validation (UCF).} For each observed
capacity family for inactive parameters, the training set excludes that family 
and contains only rows below the smaller of the same two selected tail budgets. The test set
contains the excluded capacity family for inactive parameters at those two
budgets. UCF therefore tests simultaneous extrapolation in $M$ and training
tokens $D$.

\emph{Largest-corner validation (LC).} The test configuration uses the largest
observed active-model-size family and the largest capacity in inactive parameters
available at that size having observations at both selected tail budgets.
Training is restricted to rows belonging to lower active-model-size families,
lower capacity families for inactive parameters, and lower token budgets. This
evaluates extrapolation to the largest jointly supported $(N,M,D)$ corner.

Formula selection considers these held-out MAE and RMSE results together with
the full-data fit; it is not based on the full-data error alone.

\subsection{Fitted Formulas}
\label{sec:fitted-formulas}
 We evaluated the following candidate functional forms. Reused coefficient
symbols denote analogous roles across candidates, not shared numerical values:
every formula is refitted independently. The labels are experimental names
used to track the progression of fitted hypotheses.

\emph{Three-variable laws $L(N,D,M)$.}
For compactness, define
\[
U_M(N,M)=\frac{A M^{-\kappa}}{N^\alpha},
\qquad
q_M(M)=\beta+\zeta\log M,
\]
\[
q_{NM}(N,M)=\beta+\zeta\log M+\xi\log N,
\qquad
q_N(M)=\alpha+\gamma\log M.
\]
The fitted candidates are
\begin{align*}
\text{P0:}\quad
\widehat L_0(N,D,M)
&=E+\frac{A}{N^\alpha}+\frac{B}{D^\beta},\\
\text{P1A:}\quad
\widehat L_{1A}(N,D,M)
&=E+\frac{A}{N^\alpha}+\frac{B}{D^\beta}
  +\frac{G}{M^\eta},\\
\text{P1B:}\quad
\widehat L_{1B}(N,D,M)
&=E+U_M(N,M)+\frac{B}{D^\beta},\\
\text{P1C:}\quad
\widehat L_{1C}(N,D,M)
&=E+\frac{A}{N^\alpha}+\frac{B M^{-\omega}}{D^\beta},\\
\text{P2B:}\quad
\widehat L_{2B}(N,D,M)
&=E+U_M(N,M)+\frac{B M^{-\omega}}{D^\beta},\\
\text{P3A:}\quad
\widehat L_{3A}(N,D,M)
&=E+\frac{A M^{-\kappa}}{N^{q_N(M)}}+\frac{B}{D^\beta},\\
\text{P3B:}\quad
\widehat L_{3B}(N,D,M)
&=E+U_M(N,M)+\frac{B}{D^{q_M(M)}},\\
\text{P3C:}\quad
\widehat L_{3C}(N,D,M)
&=E+\frac{A M^{-\kappa}}{N^{q_N(M)}}
  +\frac{B}{D^{q_M(M)}},\\
\text{P4:}\quad
\widehat L_4(N,D,M)
&=E+\left[U_M(N,M)+\frac{B}{D^{q_M(M)}}\right]^\rho,\\
\text{P5:}\quad
\widehat L_5(N,D,M)
&=E+U_M(N,M)+\frac{B N^{-\lambda}}{D^{q_M(M)}}.
\end{align*}
The P0 baseline is included in this group because it was fitted to the same
three-coordinate dataset, although its prediction is independent of $M$.

The candidates with a joint model--capacity-dependent token exponent are
\begin{align*}
\text{P6B:}\quad
\widehat L_{6B}(N,D,M)
&=E+U_M(N,M)+\frac{B N^{-\lambda}}{D^{q_{NM}(N,M)}},\\
\text{P6C:}\quad
\widehat L_{6C}(N,D,M)
&=E+U_M(N,M)+\frac{B M^{-\mu}}{D^{q_{NM}(N,M)}},\\
\text{P6D:}\quad
\widehat L_{6D}(N,D,M)
&=E+U_M(N,M)
  +\frac{B N^{-\lambda}M^{-\mu}}{D^{q_{NM}(N,M)}},\\
\text{P6F:}\quad
\widehat L_{6F}(N,D,M)
&=U_M(N,M)
  +\frac{B N^{-\lambda}M^{-\mu}}{D^{q_{NM}(N,M)}},\\
\text{P6G:}\quad
\widehat L_{6G}(N,D,M)
&=U_M(N,M)+\frac{B(N+M)^{-\mu}}{D^{q_{NM}(N,M)}},\\
\text{P6I:}\quad
\widehat L_{6I}(N,D,M)
&=U_M(N,M)+\frac{B(NM)^{-\mu}}{D^{q_{NM}(N,M)}},\\
\text{P6J:}\quad
\widehat L_{6J}(N,D,M)
&=U_M(N,M)
  +\frac{B\left(N/(N+M)\right)^{-\mu}}{D^{q_{NM}(N,M)}},\\
\text{P6K:}\quad
\widehat L_{6K}(N,D,M)
&=U_M(N,M)
  +\frac{B\left(N/M\right)^{-\mu}}{D^{q_{NM}(N,M)}}.
\end{align*}
Thus, P6F removes P6D's separately fitted constant floor, while P6G, P6I,
P6J, and P6K replace its two-factor data-term amplitude with a single
capacity coordinate.

\emph{Removal of the constant floor.} P6F differs from P6D only by the
standalone floor $E$. Fitted to the PKM data, P6D drives $E$ to numerically
zero ($E\approx10^{-9}$); removing it leaves the full-data error unchanged
(the MAE changes by $4\times10^{-12}$) and the LLLO errors identical or lower
(Table~\ref{tab:pkm-rmse}). The floor is not identifiable from these data: it
trades off against the model-size-dependent data amplitude $BN^{-\lambda}$.
P6C, which lacks that amplitude, retains $E=1.52$, whereas P6B and P6D, which
include it, drive $E$ to zero. Over the measured range the losses are far from
saturating, so a constant cannot be separated from the slowly decaying model-
and data-dependent terms. Farseer's model-dependent term likewise absorbs the
constant~\citep{steplaw2}. $E=0$ is therefore a parameterization choice for the
fitted range, not a claim that the irreducible loss is zero.

Finally, define the positive learned functions
\[
B_7(N)=\exp\!\left(b_0+b_N\log N\right),
\qquad
q_7(N,M)=\exp\!\left(a_0+a_N\log N+a_M\log M\right).
\]
The Farseer-style three-variable candidate is
\[
\text{P7:}\qquad
\widehat L_7(N,D,M)
=E+U_M(N,M)+\frac{B_7(N)}{D^{q_7(N,M)}}.
\]

\emph{Two-variable laws $L(N,D)$.}
Three mathematical forms were evaluated for the Dense Transformer
and STEM, with all coefficients fitted independently for each architecture.
We retain the established Chinchilla name for the two-variable P0 baseline and
use the prefix vP6F only to distinguish that form from the similarly named
three-variable candidate. For the normalized
coordinates
\[
n=\frac{N}{N_{\mathrm{ref}}},
\qquad
d=\frac{D}{D_{\mathrm{ref}}}, 
\]
the Chinchilla baseline is
\[
\text{Chinchilla:}\qquad
\widehat L_{\mathrm C}(N,D)
=E+A n^{-\alpha}+B d^{-\beta}.
\]
The memory-free joint-interaction law is evaluated on raw $N$ and $D$:
\[
q_D(N)=\beta+\xi\log N,
\qquad
\text{vP6F:}\qquad
\widehat L_{\mathrm{vP6F}}(N,D)
=\frac{A}{N^\alpha}+\frac{B N^{-\lambda}}{D^{q_D(N)}}.
\]
For Farseer, define
\[
F_U(N)=\exp\!\left(a_3 n^\gamma+b_3\right),
\qquad
B_{\mathrm F}(N)=\exp\!\left(a_2 n^\beta+b_2\right),
\]
\[
A_{\mathrm F}(N)=\exp\!\left(a_1 n^\alpha+b_1\right).
\]
Its fitted form is
\[
\text{Farseer:}\qquad
\widehat L_{\mathrm F}(N,D)
=F_U(N)+B_{\mathrm F}(N)D^{-A_{\mathrm F}(N)}.
\]

\subsection{Fitted-Formula Evaluation Results}
\label{sec:fitted-formula-results}
MAE and RMSE are reported in separate tables in the original cross-entropy
scale using a common $10^{-3}$ exponent and a three-digit mantissa; lower
values are better. The RMSE tables remain in this appendix, while the
corresponding MAE tables are collected in
Appendix~\ref{sec:additional-results}. Under the validity rules specified above, an invalid full fit or held-out family
is still reported, followed by a subscript that names the failed validity condition: i for condition~(i), ii for condition~(ii), and i,ii for failures of both kinds within that family; flagged values are set in gray, as in \textcolor{gray}{24.8}\textsubscript{i,ii}. A flagged held-out value
averages all declared folds, including the invalid ones; valid-subset averages
are not substituted. A flagged full-data value is the error of the invalid full
fit. The lowest unflagged value in each evaluation column is bolded, including
all formulas tied at the displayed precision; flagged values are never bolded.
The datasets differ across architectures, so values should be compared between
formulas within an architecture, not directly between architectures.

\emph{PKM.} The PKM fits use 192 endpoints. The complete validation suite
contains 47 folds, including the five LLLO fractions. The corresponding
full-data and LLLO RMSE trends are visualized in
Figure~\ref{fig:pkm-full-lllo}, and the selected P6F law's LLLO$_{30}$
endpoint predictions are shown in
Figure~\ref{fig:pkm-p6f-lllo30-observed-predicted}.

\begin{center}
\begin{minipage}{\textwidth}
\centering
\refstepcounter{table}\label{tab:pkm-rmse}
\textbf{Table \thetable.} \emph{PKM RMSE (all values in $10^{-3}$ cross-entropy units).}
\\{\small Gray values come from fits that failed validity condition~(i) or (ii) (Appendix~\ref{sec:fitted-formula-results}).}%

\resizebox{\textwidth}{!}{%
\begin{tabular}{l*{13}{c}}
\hline
Formula & \multicolumn{6}{c}{Full-data and LLLO} &
\multicolumn{7}{c}{Grouped and directional holdouts} \\
\cline{2-7}\cline{8-14}
& Full & LLLO$_{10}$ & LLLO$_{20}$ & LLLO$_{30}$ & LLLO$_{40}$ & LLLO$_{50}$
& LOMO & LOSO & LOCO & CF & FT & UCF & LC \\
\hline
P0 & 36.7 & 45.1 & 54.3 & 62.6 & 67.9 & 116 & 42.9 & 41.8 & 46.3 & 52.5 & 48.0 & 88.6 & 70.8 \\
P1A & 12.4 & 17.7 & 27.5 & 33.2 & 33.5 & 37.5 & 11.6 & 16.5 & 12.0 & 12.0 & 21.4 & 18.7 & 25.8 \\
P1B & 13.1 & 25.4 & 30.0 & 32.3 & 31.0 & 33.2 & 13.5 & 18.8 & 13.7 & 13.8 & 19.8 & 19.1 & 26.3 \\
P1C & 17.1 & 16.7 & 19.9 & 30.6 & 33.6 & 44.9 & 16.4 & 20.0 & 17.0 & 15.9 & 32.6 & 26.8 & 21.4 \\
P2B & 13.1 & 25.4 & 30.0 & 32.3 & 31.0 & 33.2 & 12.5 & 18.8 & 13.7 & 13.8 & 32.6 & 19.1 & 26.3 \\
P3A & 12.8 & 24.6 & 29.7 & 32.4 & 31.2 & 33.4 & 12.1 & 17.9 & 13.1 & 13.2 & 19.7 & 18.5 & \textcolor{gray}{25.7}\textsubscript{i} \\
P3B & 10.2 & 23.8 & 26.5 & 28.2 & 26.7 & 26.2 & 12.0 & 20.0 & 12.5 & 13.8 & 14.5 & 14.6 & 5.23 \\
P3C & 9.79 & 22.9 & 25.5 & 27.2 & 26.2 & 25.9 & 11.2 & \textcolor{gray}{19.1}\textsubscript{i} & 11.5 & \textcolor{gray}{12.7}\textsubscript{i} & 14.3 & 13.6 & 7.43 \\
P4 & 7.52 & 11.4 & 11.5 & 11.9 & 11.1 & 11.2 & 8.91 & 13.6 & 10.1 & 11.2 & 12.1 & 15.2 & 5.69 \\
P5 & 7.80 & 11.8 & 12.2 & 12.8 & 11.9 & 11.6 & 8.57 & \textcolor{gray}{12.6}\textsubscript{i} & 10.1 & 11.5 & 12.2 & 15.1 & 13.8 \\
P6B & 6.52 & 9.85 & 9.04 & 9.21 & 9.87 & 10.1 & 6.75 & 10.8 & 8.25 & 9.60 & 10.9 & 13.0 & 7.54 \\
P6C & 5.22 & 6.36 & 8.42 & \textcolor{gray}{9.91}\textsubscript{i} & 10.9 & 9.76 & \textcolor{gray}{5.90}\textsubscript{i} & 11.6 & 6.47 & 7.22 & 9.10 & 10.1 & 5.98 \\
P6D & \textbf{4.58} & \textbf{4.32} & \textbf{5.01} & \textbf{5.86} & 11.0 & \textbf{9.27} & \textbf{4.82} & \textbf{8.11} & \textbf{6.08} & \textbf{6.66} & 7.63 & \textcolor{gray}{8.72}\textsubscript{i} & 3.11 \\
P6F & \textbf{4.58} & \textbf{4.32} & \textbf{5.01} & \textbf{5.86} & \textbf{7.39} & \textbf{9.27} & 5.13 & 8.67 & \textbf{6.08} & \textbf{6.66} & 7.63 & \textbf{8.49} & 3.11 \\
P6G & 7.04 & 10.6 & 10.1 & 11.3 & \textcolor{gray}{23.6}\textsubscript{ii} & 12.0 & \textcolor{gray}{24.8}\textsubscript{i,ii} & \textcolor{gray}{13.7}\textsubscript{ii} & \textcolor{gray}{70.7}\textsubscript{i,ii} & 11.0 & 10.3 & \textcolor{gray}{79.0}\textsubscript{i,ii} & \textbf{2.84} \\
P6I & 7.68 & \textcolor{gray}{69.3}\textsubscript{i,ii} & 10.3 & \textcolor{gray}{101}\textsubscript{i,ii} & 24.5 & 12.1 & \textcolor{gray}{39.2}\textsubscript{i,ii} & 17.4 & \textcolor{gray}{33.0}\textsubscript{i,ii} & \textcolor{gray}{15.2}\textsubscript{ii} & 11.0 & \textcolor{gray}{14.5}\textsubscript{i,ii} & 4.07 \\
P6J & \textcolor{gray}{6.21}\textsubscript{ii} & \textcolor{gray}{8.58}\textsubscript{ii} & \textcolor{gray}{8.42}\textsubscript{ii} & \textcolor{gray}{9.49}\textsubscript{ii} & \textcolor{gray}{10.4}\textsubscript{ii} & 11.4 & \textcolor{gray}{7.12}\textsubscript{ii} & \textcolor{gray}{12.4}\textsubscript{ii} & \textcolor{gray}{10.6}\textsubscript{ii} & \textcolor{gray}{8.49}\textsubscript{ii} & \textcolor{gray}{8.82}\textsubscript{ii} & \textcolor{gray}{8.52}\textsubscript{ii} & 3.62 \\
P6K & 5.67 & 6.13 & 6.53 & 8.24 & 9.66 & 11.1 & 7.41 & 10.6 & 7.98 & 7.34 & \textbf{6.79} & 9.63 & \textcolor{gray}{97.1}\textsubscript{i,ii} \\
P7 & 7.48 & 10.9 & 10.9 & 11.5 & 10.9 & 11.4 & 7.93 & 11.6 & 9.65 & 11.1 & 11.5 & 14.3 & 13.6 \\
\hline
\end{tabular}%
}
\end{minipage}
\end{center}

\emph{MoE.} The MoE fits use 124 endpoints across 15 model lineages and 36
validation folds, including the five LLLO fractions. P6F gives valid full-data
and LLLO fits, with LLLO RMSE rising steadily from $5.36\mathrm{e}{-3}$ to
$9.79\mathrm{e}{-3}$ as the held-out fraction grows. These errors indicate a good fit, comparable
to or lower than the LLLO errors reported in other scaling-law papers
(Section~\ref{sec:scaling-predictive-accuracy}). The LOMO, LOCO, CF, and
UCF families contain invalid folds (Table~\ref{tab:moe-rmse}); these holdouts
remove entire lineages or capacity families, which is demanding for the
smaller MoE dataset (124 endpoints and four capacity families, versus 192 and
six for PKM). RMSE for the full-data fit and all five LLLO fractions is shown
in Figure~\ref{fig:moe-full-lllo}, and LLLO$_{30}$ observed-versus-predicted
losses in Figure~\ref{fig:moe-p6f-lllo30-observed-predicted}.

\begin{center}
\begin{minipage}{\textwidth}
\centering
\refstepcounter{table}\label{tab:moe-rmse}
\textbf{Table \thetable.} \emph{MoE RMSE (all values in $10^{-3}$ cross-entropy units).}
\\{\small Gray values come from fits that failed validity condition~(i) or (ii) (Appendix~\ref{sec:fitted-formula-results}).}%

\resizebox{\textwidth}{!}{%
\begin{tabular}{l*{13}{c}}
\hline
Formula & \multicolumn{6}{c}{Full-data and LLLO} &
\multicolumn{7}{c}{Grouped and directional holdouts} \\
\cline{2-7}\cline{8-14}
& Full & LLLO$_{10}$ & LLLO$_{20}$ & LLLO$_{30}$ & LLLO$_{40}$ & LLLO$_{50}$
& LOMO & LOSO & LOCO & CF & FT & UCF & LC \\
\hline
P0 & 23.5 & 22.3 & 22.1 & 28.2 & 47.2 & 52.0 & 25.3 & 22.8 & 29.1 & 32.6 & 45.4 & 40.8 & 45.0 \\
P1A & 10.3 & 13.7 & 15.9 & 22.9 & 33.5 & 37.8 & 12.1 & 12.7 & 11.7 & 12.3 & 16.7 & 17.5 & 26.8 \\
P1B & 10.4 & 14.1 & 15.2 & 25.8 & 29.9 & 35.8 & 12.8 & 12.6 & 11.7 & 12.2 & 16.5 & 16.4 & 20.5 \\
P1C & 13.3 & 11.8 & 15.8 & 20.9 & 40.4 & 43.2 & 12.8 & 15.5 & 14.9 & 14.5 & 22.7 & 22.3 & 22.0 \\
P2B & 10.4 & 11.8 & 15.2 & 25.8 & 29.9 & 35.8 & 10.2 & 12.6 & 12.7 & 12.2 & 22.7 & 16.4 & 22.0 \\
P3A & 10.3 & 14.4 & 15.8 & 25.3 & 29.9 & 35.9 & 9.98 & 12.6 & 11.5 & 12.0 & 16.5 & 16.5 & 25.6 \\
P3B & 7.64 & 6.70 & 8.88 & 17.5 & 17.1 & 23.4 & 9.46 & 12.2 & 10.3 & 13.6 & 7.98 & 9.91 & 14.8 \\
P3C & 7.28 & 6.79 & 8.42 & 15.8 & 16.0 & 22.6 & 8.48 & \textcolor{gray}{12.0}\textsubscript{i} & 10.2 & 13.4 & 7.93 & \textbf{9.62} & 15.9 \\
P4 & 6.09 & 5.99 & 6.48 & 9.98 & 9.50 & 10.7 & 8.30 & 8.85 & \textbf{8.70} & 11.8 & 8.39 & 12.2 & 13.7 \\
P5 & 6.26 & 6.16 & 6.94 & 11.0 & 10.2 & 11.8 & \textbf{6.90} & \textbf{7.15} & 8.91 & 12.1 & 8.41 & 12.4 & 14.6 \\
P6B & 6.10 & 6.04 & 6.18 & 9.09 & 9.46 & 10.4 & \textcolor{gray}{7.38}\textsubscript{i} & 8.65 & 9.18 & \textbf{11.5} & 8.13 & 11.8 & 13.1 \\
P6C & 5.79 & \textbf{5.29} & 6.51 & 9.48 & 10.0 & 11.2 & 7.57 & 8.39 & 9.70 & 12.0 & 7.89 & 10.8 & 14.5 \\
P6D & \textbf{5.78} & 5.36 & 6.17 & 8.49 & \textbf{9.11} & 10.4 & \textcolor{gray}{7.13}\textsubscript{i} & \textcolor{gray}{8.49}\textsubscript{i} & \textcolor{gray}{9.90}\textsubscript{i} & 12.1 & 7.75 & 11.1 & 14.3 \\
P6F & 5.80 & 5.36 & 6.08 & 8.49 & \textbf{9.11} & \textbf{9.79} & \textcolor{gray}{13.4}\textsubscript{i} & 8.56 & \textcolor{gray}{52.2}\textsubscript{i} & \textcolor{gray}{12.2}\textsubscript{i} & 7.67 & \textcolor{gray}{46.7}\textsubscript{i,ii} & 14.3 \\
P6G & 6.27 & 6.52 & 6.16 & 8.71 & 9.39 & 10.1 & 7.43 & \textcolor{gray}{30.6}\textsubscript{i,ii} & \textcolor{gray}{17.3}\textsubscript{ii} & \textcolor{gray}{45.7}\textsubscript{i,ii} & 7.62 & \textcolor{gray}{11.9}\textsubscript{ii} & \textcolor{gray}{16.1}\textsubscript{i,ii} \\
P6I & 6.27 & 6.36 & 6.17 & 9.47 & 18.5 & \textcolor{gray}{101}\textsubscript{i,ii} & \textcolor{gray}{16.3}\textsubscript{i,ii} & \textcolor{gray}{31.9}\textsubscript{i,ii} & \textcolor{gray}{198}\textsubscript{i,ii} & 12.6 & 7.72 & 15.5 & \textcolor{gray}{2.97}\textsubscript{i,ii} \\
P6J & 6.25 & 6.68 & 6.22 & 8.76 & 9.39 & 10.1 & 7.35 & 9.27 & \textcolor{gray}{26.8}\textsubscript{ii} & 18.4 & 7.60 & \textcolor{gray}{13.2}\textsubscript{ii} & \textbf{2.61} \\
P6K & 6.01 & 6.10 & \textbf{5.74} & \textbf{8.48} & 9.42 & 10.1 & 8.54 & 8.59 & 9.78 & 12.3 & \textbf{7.37} & 10.6 & 8.92 \\
P7 & 6.22 & 6.31 & 6.45 & 9.96 & 9.78 & 10.9 & \textcolor{gray}{6.76}\textsubscript{i} & 8.22 & 8.91 & 11.7 & 8.32 & 11.8 & 14.9 \\
\hline
\end{tabular}%
}
\end{minipage}
\end{center}

\emph{Dense Transformer.} The dense-Transformer fits use 57 endpoints and 12
validation folds. Because each two-variable model-size setting defines one
lineage, LOMO is also the available held-out-size test. A separate LOSO family
and the capacity-based LOCO, CF, and UCF tests are not defined for this dataset;
LC denotes the largest $(N,D)$ corner.

\begin{center}
\begin{minipage}{\textwidth}
\centering
\refstepcounter{table}\label{tab:vanilla-rmse}
\textbf{Table \thetable.} \emph{Dense Transformer RMSE (all values in $10^{-3}$ cross-entropy units).}

\resizebox{\textwidth}{!}{%
\begin{tabular}{lccccccccc}
\hline
Formula & Full & LLLO$_{10}$ & LLLO$_{20}$ & LLLO$_{30}$ & LLLO$_{40}$ & LLLO$_{50}$ & LOMO & FT & LC \\
\hline
Chinchilla & 18.8 & 42.2 & 49.4 & 51.7 & 61.4 & 61.1 & 25.6 & 25.8 & 27.3 \\
vP6F & 4.65 & \textbf{6.49} & \textbf{6.24} & \textbf{6.08} & \textbf{9.30} & \textbf{10.6} & \textbf{11.4} & \textbf{6.17} & 2.30 \\
Farseer & \textbf{4.58} & 6.98 & 7.06 & 6.22 & 9.72 & 38.7 & 47.5 & 7.13 & \textbf{1.69} \\
\hline
\end{tabular}%
}
\end{minipage}
\end{center}

\emph{STEM.} The STEM fits use 48 endpoints and 11 validation folds. As for
the Dense Transformer, LOMO is the available held-out-size test,
capacity-based tests are not defined, and LC denotes the largest $(N,D)$
corner.

\begin{center}
\begin{minipage}{\textwidth}
\centering
\refstepcounter{table}\label{tab:stem-rmse}
\textbf{Table \thetable.} \emph{STEM RMSE (all values in $10^{-3}$ cross-entropy units).}

\resizebox{\textwidth}{!}{%
\begin{tabular}{lccccccccc}
\hline
Formula & Full & LLLO$_{10}$ & LLLO$_{20}$ & LLLO$_{30}$ & LLLO$_{40}$ & LLLO$_{50}$ & LOMO & FT & LC \\
\hline
Chinchilla & 13.2 & 26.2 & 29.3 & 34.2 & 35.1 & 36.4 & 16.8 & 17.1 & 25.9 \\
vP6F & 1.76 & \textbf{1.27} & \textbf{2.82} & \textbf{2.88} & 4.45 & \textbf{5.18} & \textbf{4.32} & \textbf{2.31} & \textbf{1.15} \\
Farseer & \textbf{1.61} & 5.12 & 4.44 & 7.86 & \textbf{3.65} & 16.2 & 15.0 & 2.73 & 4.27 \\
\hline
\end{tabular}%
}
\end{minipage}
\end{center}

\phantomsection
\label{sec:two-variable-law-selection}
\emph{Selection of the two-variable law.}
Farseer fits nine coefficients, three more than vP6F. It achieves lower
full-data error for both architectures: for the Dense Transformer, MAE/RMSE
decrease from $3.32/4.65$ to $3.25/4.58$; for STEM, they decrease from
$1.43/1.76$ to $1.15/1.61$, with all values in $10^{-3}$ units. However, vP6F
has lower held-out error in 27 of 32 architecture--metric--family comparisons
reported in the Dense Transformer and STEM RMSE tables above and their MAE
counterparts in Appendix~\ref{sec:additional-results}. The LLLO trends
are shown in Figures~\ref{fig:vanilla-full-lllo}
and~\ref{fig:stem-full-lllo}.

Farseer's lower full-data error but poorer held-out extrapolation, together
with its three additional fitted coefficients, is consistent with overfitting
the observed endpoints, although it does not uniquely prove overfitting. We
therefore retain the more parsimonious and better-generalizing vP6F. Its
structure also matches the selected three-variable P6F law after removing the
$M$-dependent factors, enabling direct comparison between the two- and
three-variable fits.

\emph{Joint Sparse LLM Scaling-Law Formula.} We additionally fit the
three-variable candidate laws jointly to the PKM and MoE observations. The
joint dataset contains 316 endpoints---192 PKM and 124 MoE---across 35
architecture-qualified model lineages. Its validation suite contains 74 folds,
including the five LLLO fractions. The reported values include the full-data
fits and every validation fold but no bootstrap uncertainty estimates.
P6F remains the working reference for this joint fit; 72 of
its 74 validation folds pass the applicable validity checks. The corresponding
full-data and LLLO RMSE trends are visualized in
Figure~\ref{fig:joint-sparse-full-lllo}, and the joint PKM--MoE P6F law's
LLLO$_{30}$ endpoint predictions are shown in
Figure~\ref{fig:joint-sparse-p6f-lllo30-observed-predicted}.

\begin{center}
\begin{minipage}{\textwidth}
\centering
\refstepcounter{table}\label{tab:joint-sparse-rmse}
\textbf{Table \thetable.} \emph{Joint PKM--MoE RMSE (all values in $10^{-3}$ cross-entropy units).}
\\{\small Gray values come from fits that failed validity condition~(i) or (ii) (Appendix~\ref{sec:fitted-formula-results}).}%

\resizebox{\textwidth}{!}{%
\begin{tabular}{l*{13}{c}}
\hline
Formula & \multicolumn{6}{c}{Full-data and LLLO} &
\multicolumn{7}{c}{Grouped and directional holdouts} \\
\cline{2-7}\cline{8-14}
& Full & LLLO$_{10}$ & LLLO$_{20}$ & LLLO$_{30}$ & LLLO$_{40}$ & LLLO$_{50}$
& LOMO & LOSO & LOCO & CF & FT & UCF & LC \\
\hline
P0 & 37.6 & 48.8 & 46.8 & 52.7 & 63.6 & 78.4 & 34.0 & 37.1 & 40.2 & 46.5 & 46.8 & 49.7 & 55.7 \\
P1A & 13.0 & 22.7 & 23.0 & 26.7 & 45.1 & 33.5 & 12.6 & 15.2 & 13.8 & 14.1 & 19.5 & 20.2 & 42.2 \\
P1B & 13.8 & 50.0 & 27.6 & 28.5 & 32.1 & 34.1 & 13.2 & 18.9 & 16.6 & 14.9 & 21.0 & 19.0 & 49.3 \\
P1C & 17.1 & 26.3 & 26.9 & 33.7 & 37.7 & 44.5 & 17.1 & 18.4 & 19.1 & 18.6 & 31.7 & 29.9 & 48.0 \\
P2B & 13.8 & 26.3 & 26.9 & 28.5 & 32.1 & 34.1 & 13.1 & 18.8 & 16.3 & 14.9 & 31.7 & 19.0 & 49.3 \\
P3A & 13.6 & 27.4 & \textcolor{gray}{27.3}\textsubscript{i} & 28.4 & 32.4 & \textcolor{gray}{34.3}\textsubscript{i} & \textcolor{gray}{12.8}\textsubscript{i} & \textcolor{gray}{18.1}\textsubscript{i} & \textcolor{gray}{14.4}\textsubscript{i} & 14.6 & 20.8 & 19.3 & 49.6 \\
P3B & 12.4 & 24.9 & 26.8 & 26.9 & 30.5 & 32.5 & 12.8 & 20.1 & 13.5 & 15.0 & 19.2 & 16.6 & 37.5 \\
P3C & 12.2 & 25.9 & 25.8 & 26.2 & 30.1 & 32.4 & \textcolor{gray}{12.5}\textsubscript{i} & \textcolor{gray}{20.8}\textsubscript{i} & 13.3 & 14.9 & 19.1 & 17.0 & 38.9 \\
P4 & 10.2 & 20.8 & 16.7 & 14.4 & 15.1 & 15.2 & 10.9 & 16.6 & 12.7 & 13.5 & 16.1 & 17.3 & 37.0 \\
P5 & 10.4 & 19.2 & 17.3 & 14.9 & 15.5 & 15.3 & 10.9 & 13.6 & 12.7 & 13.7 & 16.4 & 16.6 & 37.2 \\
P6B & 9.65 & 22.8 & 12.2 & 12.6 & 13.6 & 13.9 & 9.85 & 11.8 & \textcolor{gray}{11.7}\textsubscript{i} & 12.6 & 14.0 & 15.7 & 35.0 \\
P6C & 9.17 & 11.8 & 12.0 & 11.4 & 13.1 & 13.3 & \textcolor{gray}{9.05}\textsubscript{i} & \textcolor{gray}{14.7}\textsubscript{i} & \textcolor{gray}{10.0}\textsubscript{i} & \textcolor{gray}{11.5}\textsubscript{i} & 11.6 & \textcolor{gray}{13.5}\textsubscript{i} & 33.6 \\
P6D & \textbf{8.74} & \textbf{8.28} & \textbf{8.16} & \textcolor{gray}{9.76}\textsubscript{i} & 13.6 & \textbf{12.6} & \textcolor{gray}{8.42}\textsubscript{i} & \textbf{10.6} & \textbf{9.48} & \textbf{10.9} & \textbf{10.8} & \textcolor{gray}{13.5}\textsubscript{i} & 37.7 \\
P6F & \textbf{8.74} & \textbf{8.28} & \textbf{8.16} & \textbf{9.76} & \textbf{11.9} & \textbf{12.6} & \textbf{8.42} & \textbf{10.6} & \textbf{9.48} & \textcolor{gray}{10.8}\textsubscript{i} & \textbf{10.8} & 12.7 & \textcolor{gray}{26.6}\textsubscript{i} \\
P6G & 9.71 & 13.9 & 12.0 & 13.1 & 14.8 & \textcolor{gray}{180}\textsubscript{i,ii} & 10.5 & 15.2 & 13.7 & \textcolor{gray}{22.1}\textsubscript{i,ii} & 13.3 & 15.8 & \textbf{8.16} \\
P6I & 10.1 & 14.3 & 12.5 & 15.2 & \textcolor{gray}{172}\textsubscript{i,ii} & 29.2 & 11.8 & 17.4 & \textcolor{gray}{22.5}\textsubscript{i,ii} & 16.1 & 13.4 & \textcolor{gray}{42.7}\textsubscript{i} & 24.6 \\
P6J & 9.83 & 14.0 & 14.1 & 20.0 & 15.3 & 14.8 & 10.9 & 16.8 & \textcolor{gray}{15.6}\textsubscript{ii} & \textcolor{gray}{15.3}\textsubscript{ii} & 13.3 & \textcolor{gray}{17.7}\textsubscript{ii} & 21.1 \\
P6K & 9.38 & 9.62 & 10.5 & 13.9 & 15.0 & 15.2 & 9.89 & 12.6 & 10.5 & 11.4 & 11.2 & \textbf{12.4} & 14.7 \\
P7 & 10.2 & 18.2 & 15.8 & 13.8 & 14.8 & 14.8 & 10.6 & 13.0 & 12.5 & 13.4 & 15.5 & 16.3 & 36.5 \\
\hline
\end{tabular}%
}
\end{minipage}
\end{center}

\phantomsection
\label{sec:joint-matched-moe-lllo}
\emph{Matched MoE LLLO holdout.}
The LLLO columns in Table~\ref{tab:joint-sparse-rmse} rank endpoints within the
pooled PKM--MoE dataset, whereas the MoE LLLO columns in
Table~\ref{tab:moe-rmse} rank endpoints within the MoE dataset alone. To isolate the effect of sharing coefficients, we therefore
construct a matched evaluation from the exact MoE LLLO test sets. For each
fraction, the separately fitted MoE law is trained on the retained MoE
endpoints, while the joint law is refitted on those same retained MoE endpoints
plus all 192 PKM endpoints. Both laws are then evaluated on the identical
held-out MoE endpoints; Table~\ref{tab:joint-matched-moe-lllo} reports the
results. All five refits
converge; the held-out endpoints match the MoE folds exactly, every
held-out endpoint is absent from joint training, and every PKM endpoint remains
in joint training.

\begin{center}
\begin{minipage}{\textwidth}
\centering
\refstepcounter{table}\label{tab:joint-matched-moe-lllo}
\textbf{Table \thetable.} \emph{Matched MoE LLLO evaluation of the separately
fitted MoE law and joint PKM--MoE law. Errors are reported in
$\mathrm{e}{-3}$ cross-entropy units; the ratio is joint RMSE divided by MoE
RMSE.}

\resizebox{0.92\textwidth}{!}{%
\begin{tabular}{lrrrrrrr}
\hline
Split & MoE train & MoE test & MoE MAE & Joint MAE & MoE RMSE & Joint RMSE & RMSE ratio \\
\hline
LLLO$_{10}$ & 111 & 13 & 4.28 & 9.07 & 5.36 & 11.01 & 2.05 \\
LLLO$_{20}$ & 99 & 25 & 4.50 & 7.61 & 6.08 & 10.01 & 1.65 \\
LLLO$_{30}$ & 86 & 38 & 5.99 & 9.29 & 8.49 & 12.27 & 1.44 \\
LLLO$_{40}$ & 74 & 50 & 6.53 & 10.23 & 9.11 & 13.01 & 1.43 \\
LLLO$_{50}$ & 62 & 62 & 6.83 & 10.61 & 9.79 & 13.19 & 1.35 \\
\hline
\end{tabular}%
}
\end{minipage}
\end{center}

The separately fitted MoE law has lower MAE and RMSE in every matched
evaluation. Therefore, for these observed architectures and extrapolation
directions, adding PKM data while forcing one shared coefficient vector
consistently degrades prediction of the held-out MoE endpoints. This matched
comparison directly supports architecture-specific coefficients for the
allocation analysis and provides evidence that PKM and MoE follow different
scaling behavior over the measured regime.

The diagnostics below are grouped by evaluation type. Each group contains the
PKM, MoE, joint PKM--MoE, Dense Transformer, and STEM charts at their original
full-page width.

\subsubsection{Full-data and LLLO RMSE}

\begin{center}
\begin{minipage}{\textwidth}
\centering
\refstepcounter{figure}\label{fig:pkm-full-lllo}
\includegraphics[width=0.98\textwidth,height=0.36\textheight,keepaspectratio,trim=17.2bp 20.7bp 11.3bp 29.8bp,clip]{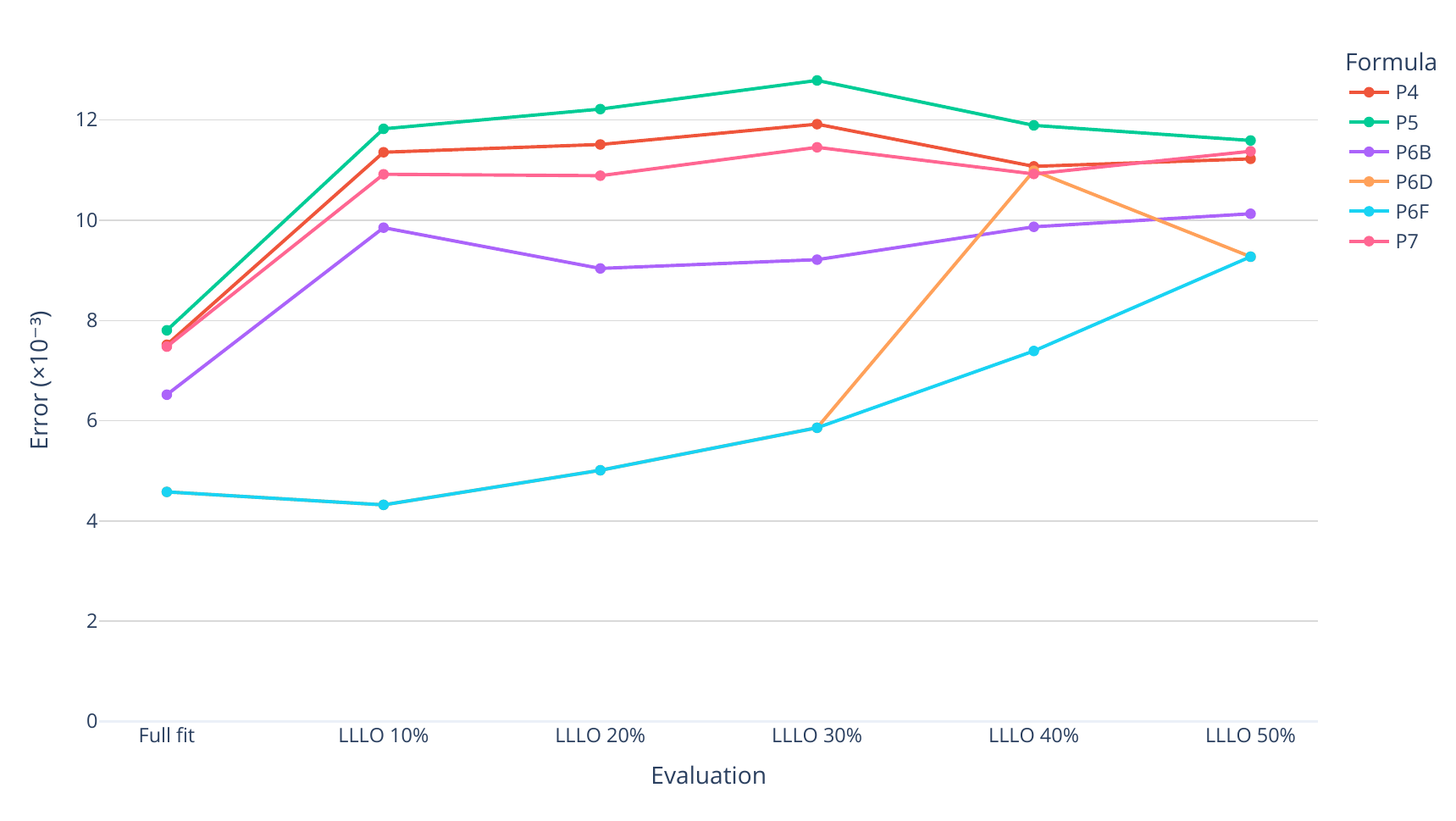}

{\small\textbf{Figure \thefigure.} \emph{PKM full-data and LLLO evaluation.}
RMSE is shown in $10^{-3}$ cross-entropy units; lower values are better. Lines
connect the separately refitted evaluation settings only as visual guides.\par}
\end{minipage}
\end{center}

\begin{center}
\begin{minipage}{\textwidth}
\centering
\refstepcounter{figure}\label{fig:moe-full-lllo}
\includegraphics[width=0.98\textwidth,height=0.36\textheight,keepaspectratio,trim=17.2bp 20.7bp 11.3bp 29.8bp,clip]{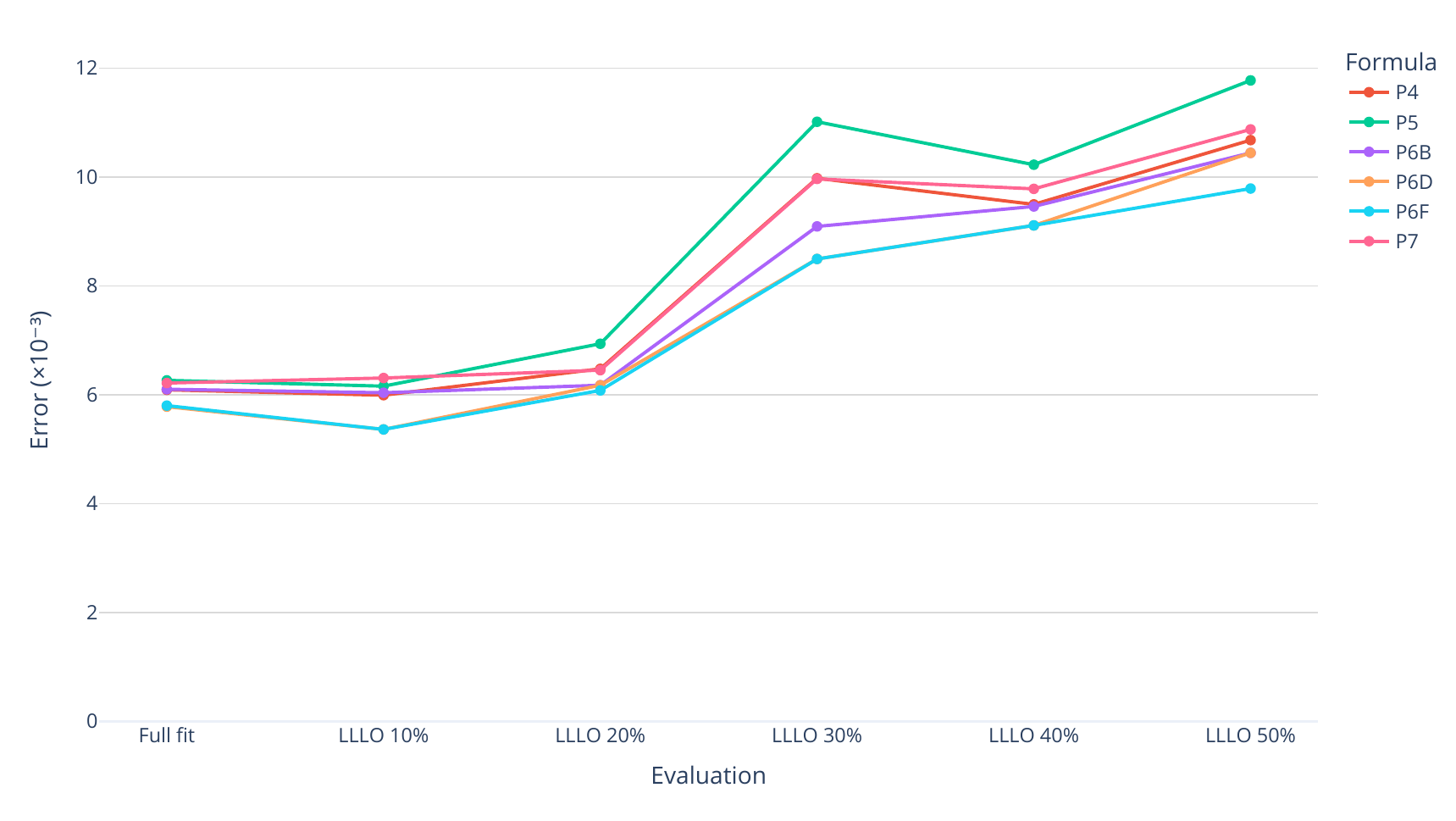}

{\small\textbf{Figure \thefigure.} \emph{MoE full-data and LLLO evaluation.}
RMSE is shown in $10^{-3}$ cross-entropy units; lower values are better. Lines
connect the separately refitted evaluation settings only as visual guides. A
missing point denotes an invalid family metric.\par}
\end{minipage}
\end{center}

\begin{center}
\begin{minipage}{\textwidth}
\centering
\refstepcounter{figure}\label{fig:joint-sparse-full-lllo}
\includegraphics[width=0.98\textwidth,height=0.36\textheight,keepaspectratio,trim=17.2bp 20.7bp 11.3bp 29.8bp,clip]{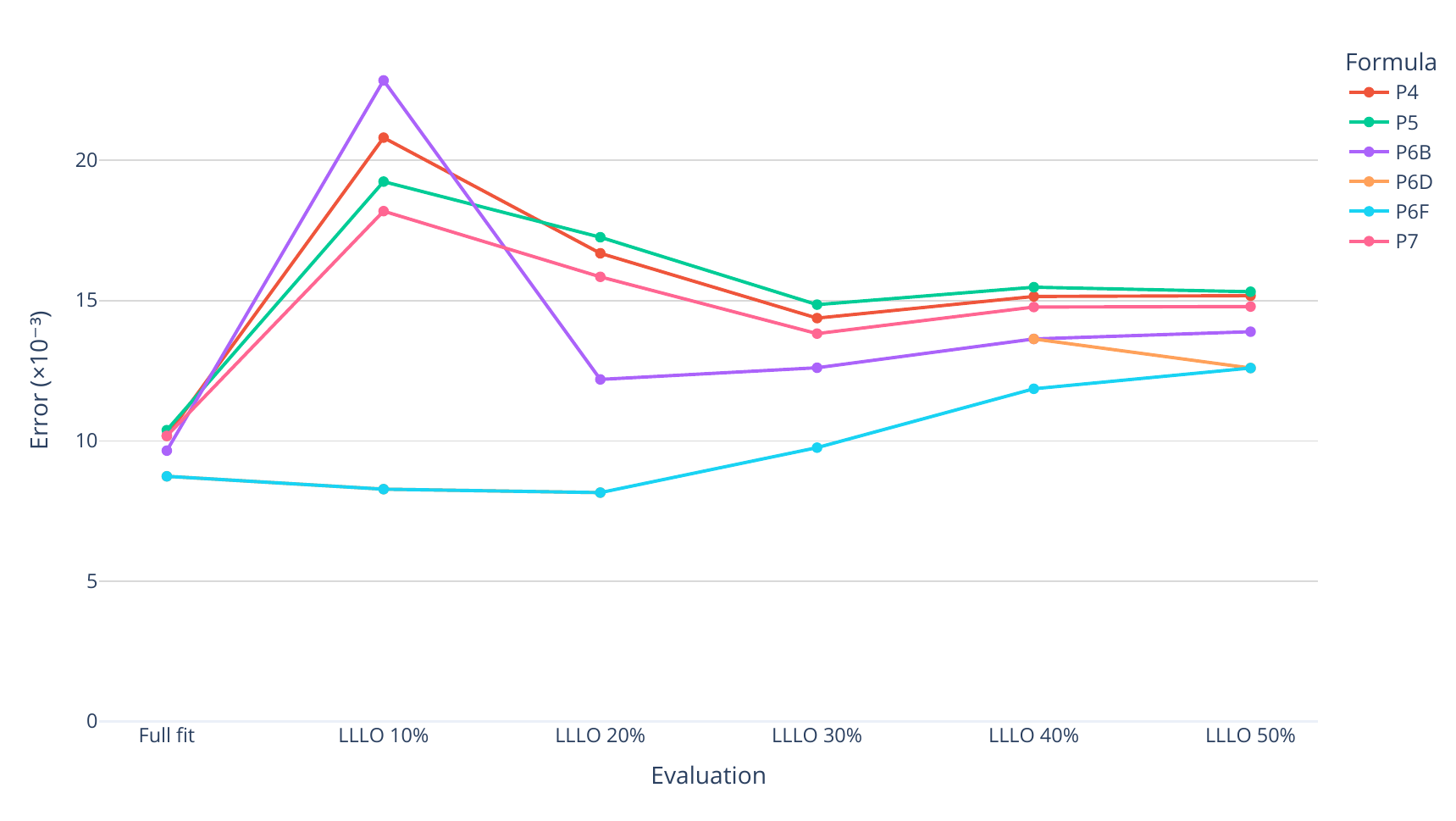}

{\small\textbf{Figure \thefigure.} \emph{Joint PKM--MoE full-data and LLLO
evaluation.} The joint three-variable laws are evaluated on the pooled PKM
and MoE endpoints. RMSE is shown in $10^{-3}$ cross-entropy units; lower values
are better, and missing points denote invalid family metrics.\par}
\end{minipage}
\end{center}

\begin{center}
\begin{minipage}{\textwidth}
\centering
\refstepcounter{figure}\label{fig:vanilla-full-lllo}
\includegraphics[width=0.98\textwidth,height=0.36\textheight,keepaspectratio,trim=17.2bp 20.7bp 14.0bp 29.8bp,clip]{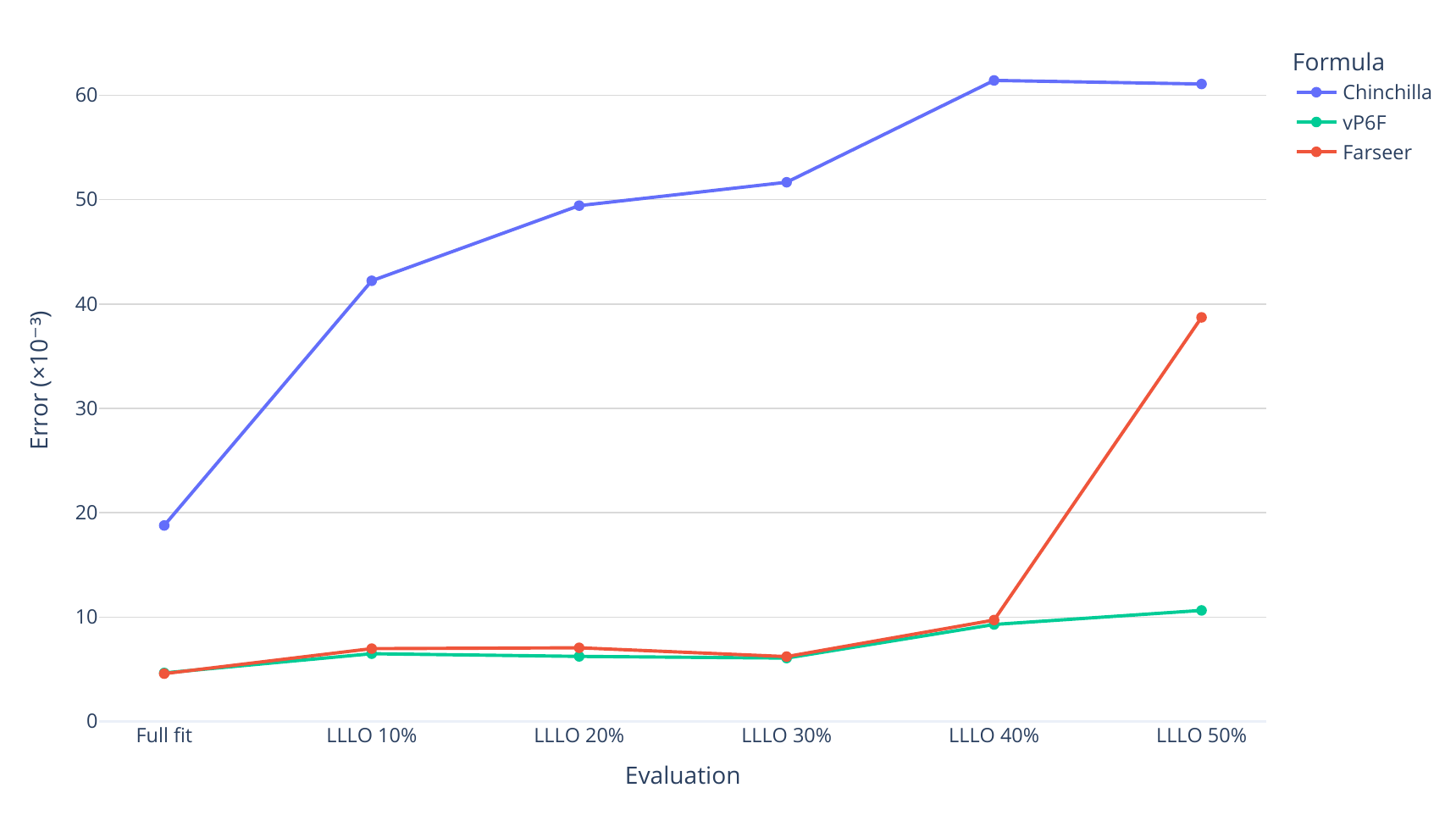}

{\small\textbf{Figure \thefigure.} \emph{Dense Transformer full-data and
LLLO evaluation.} RMSE is shown in $10^{-3}$ cross-entropy units; lower values
are better. Lines connect the separately refitted evaluation settings only as
visual guides.\par}
\end{minipage}
\end{center}

\begin{center}
\begin{minipage}{\textwidth}
\centering
\refstepcounter{figure}\label{fig:stem-full-lllo}
\includegraphics[width=0.98\textwidth,height=0.36\textheight,keepaspectratio,trim=17.2bp 20.7bp 14.0bp 29.8bp,clip]{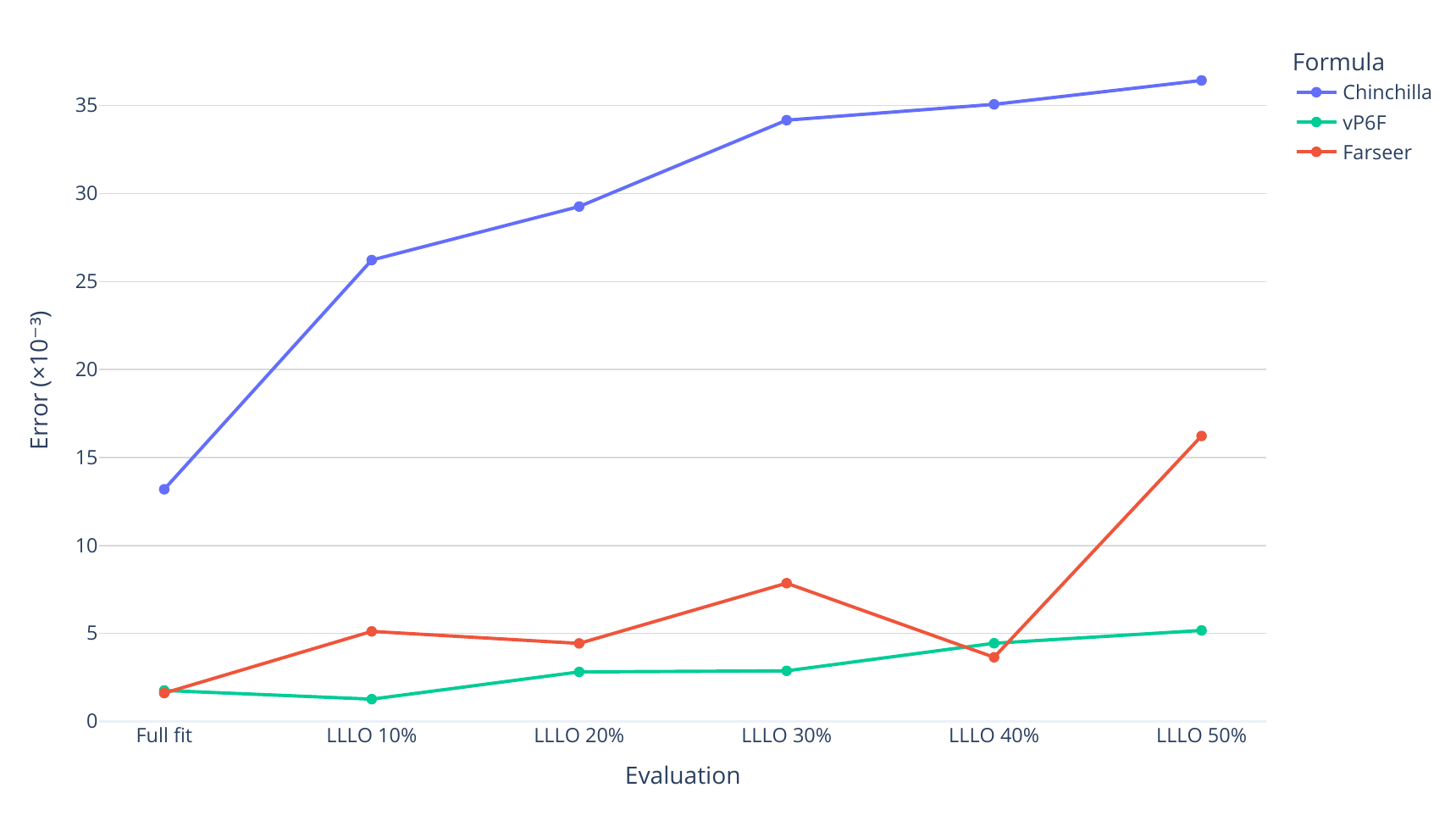}

{\small\textbf{Figure \thefigure.} \emph{STEM full-data and LLLO evaluation.}
RMSE is shown in $10^{-3}$ cross-entropy units; lower values are better. Lines
connect the separately refitted evaluation settings only as visual guides.\par}
\end{minipage}
\end{center}

\subsubsection{LLLO$_{30}$ observed versus predicted cross-entropy}

\begin{center}
\begin{minipage}{\textwidth}
\centering
\refstepcounter{figure}\label{fig:pkm-p6f-lllo30-observed-predicted}
\includegraphics[width=0.98\textwidth,height=0.36\textheight,keepaspectratio,trim=68.3bp 16.8bp 134.7bp 22.5bp,clip]{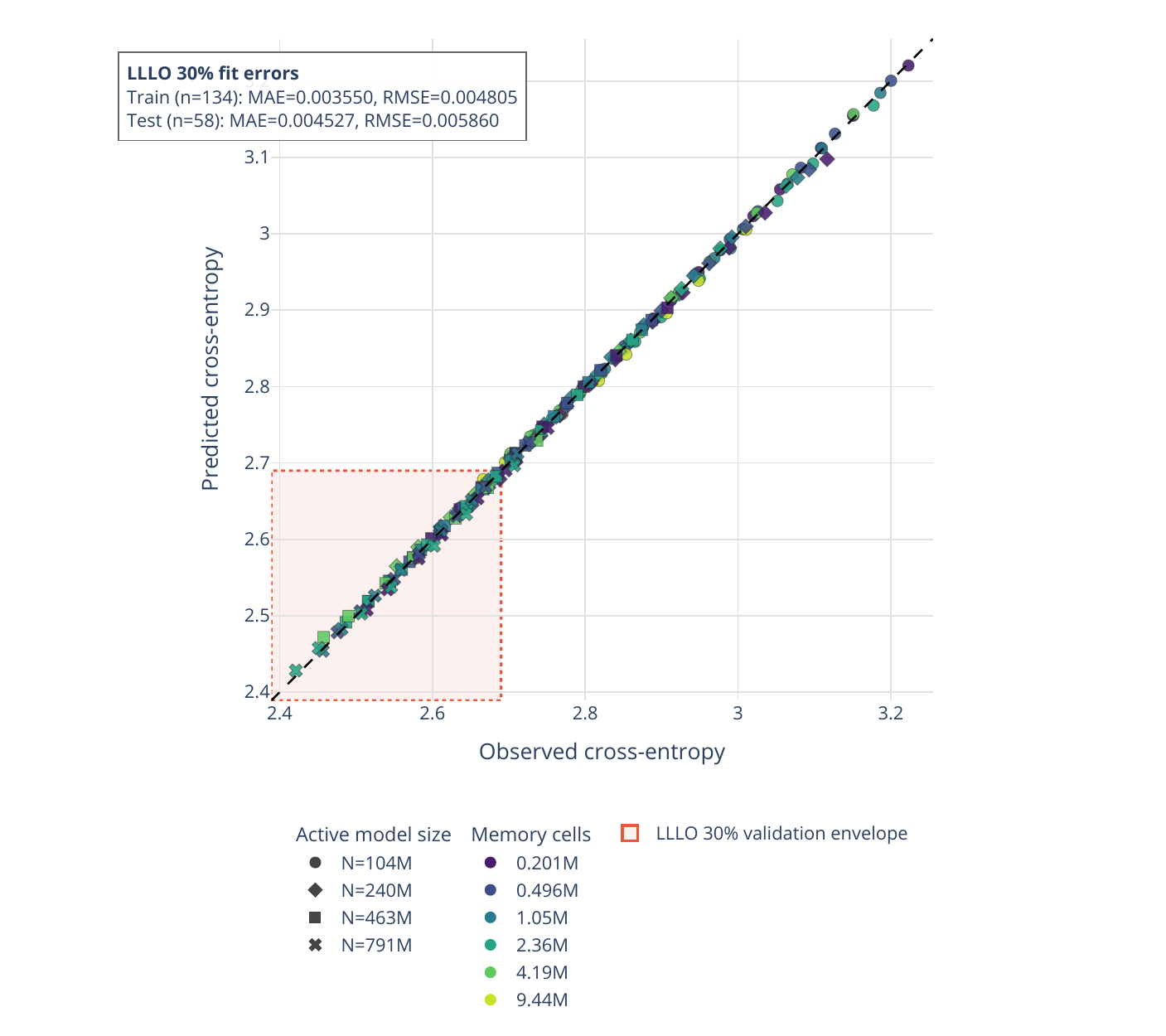}

{\small\textbf{Figure \thefigure.} \emph{PKM P6F LLLO$_{30}$ observed versus
predicted cross-entropy.} P6F is fitted on the highest-loss 70\% of endpoints
and evaluated on the held-out lowest-loss 30\%. The shaded square marks the
validation envelope and the dashed diagonal denotes exact prediction. Marker
shape identifies $N$, and color identifies memory-table size.\par}
\end{minipage}
\end{center}

\begin{center}
\begin{minipage}{\textwidth}
\centering
\refstepcounter{figure}\label{fig:moe-p6f-lllo30-observed-predicted}
\includegraphics[width=0.98\textwidth,height=0.36\textheight,keepaspectratio,trim=68.3bp 17.0bp 121.2bp 22.5bp,clip]{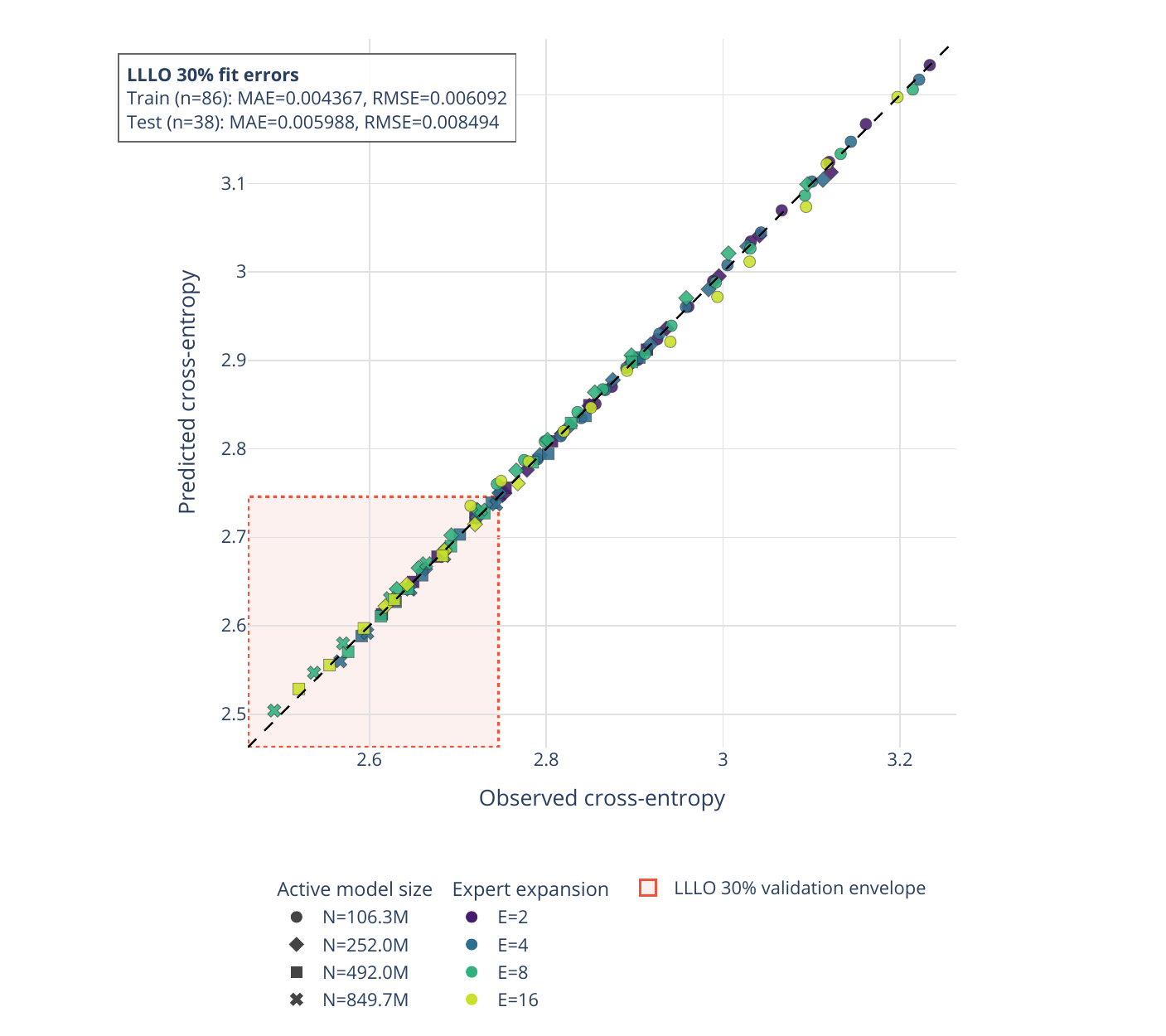}

{\small\textbf{Figure \thefigure.} \emph{MoE P6F LLLO$_{30}$ observed versus
predicted cross-entropy.} P6F is fitted on the highest-loss 70\% of endpoints
and evaluated on the held-out lowest-loss 30\%. The shaded square marks the
validation envelope and the dashed diagonal denotes exact prediction. Marker
shape identifies $N$, and color identifies expert expansion.\par}
\end{minipage}
\end{center}

\begin{center}
\begin{minipage}{\textwidth}
\centering
\refstepcounter{figure}\label{fig:joint-sparse-p6f-lllo30-observed-predicted}
\includegraphics[width=0.98\textwidth,height=0.36\textheight,keepaspectratio,trim=68.3bp 17.0bp 121.2bp 22.5bp,clip]{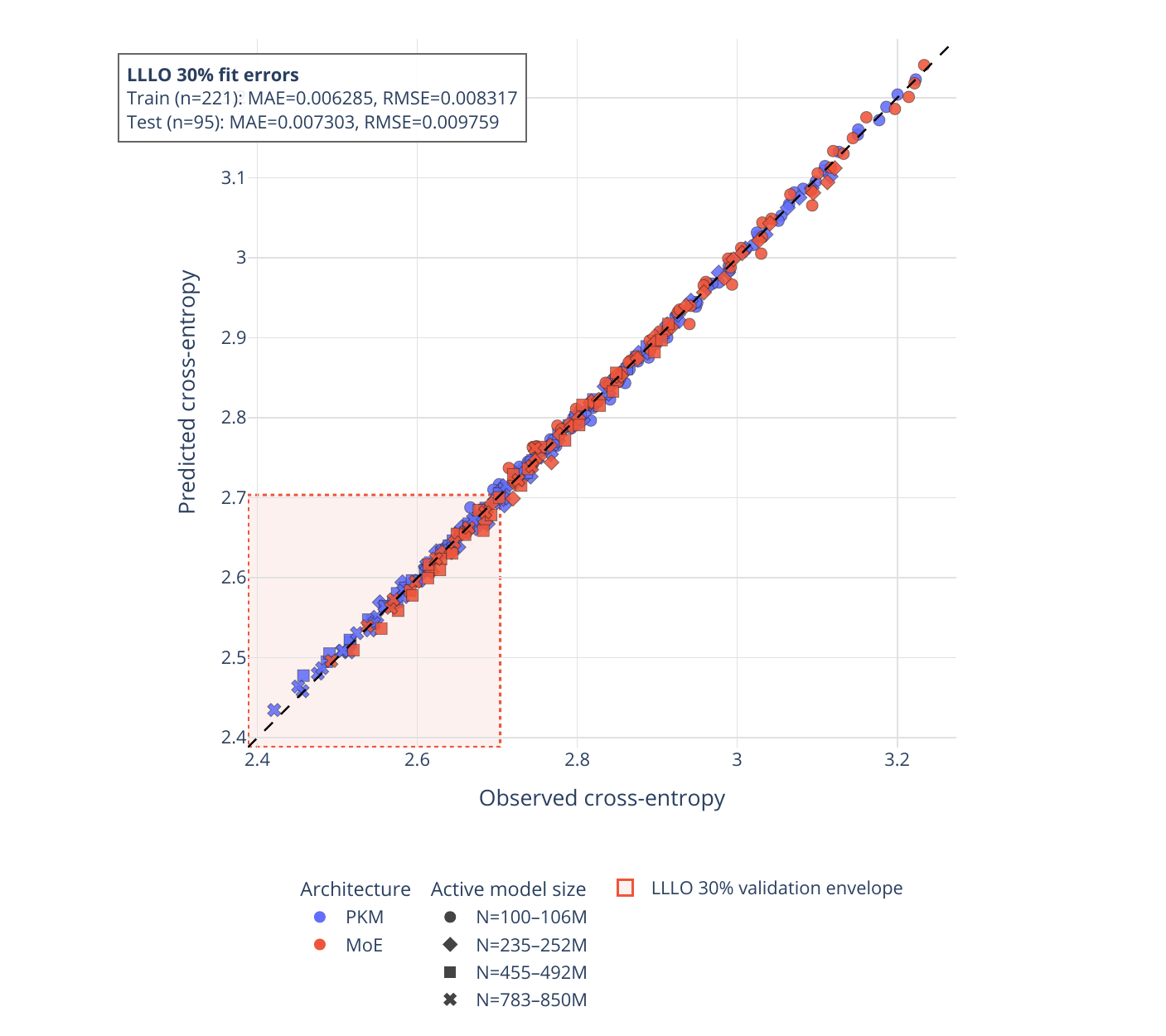}

{\small\textbf{Figure \thefigure.} \emph{Joint PKM--MoE P6F LLLO$_{30}$
observed versus predicted cross-entropy.} The joint PKM--MoE P6F law is fitted on the
highest-loss 70\% of endpoints and evaluated on the held-out lowest-loss 30\%.
The shaded square marks the validation envelope and the dashed diagonal
denotes exact prediction. Marker shape identifies the shared active-model-size
range, and color identifies architecture.\par}
\end{minipage}
\end{center}

\begin{center}
\begin{minipage}{\textwidth}
\centering
\refstepcounter{figure}\label{fig:vanilla-vp6f-lllo30-observed-predicted}
\includegraphics[width=0.98\textwidth,height=0.36\textheight,keepaspectratio,trim=10.5bp 15.8bp 5.0bp 22.5bp,clip]{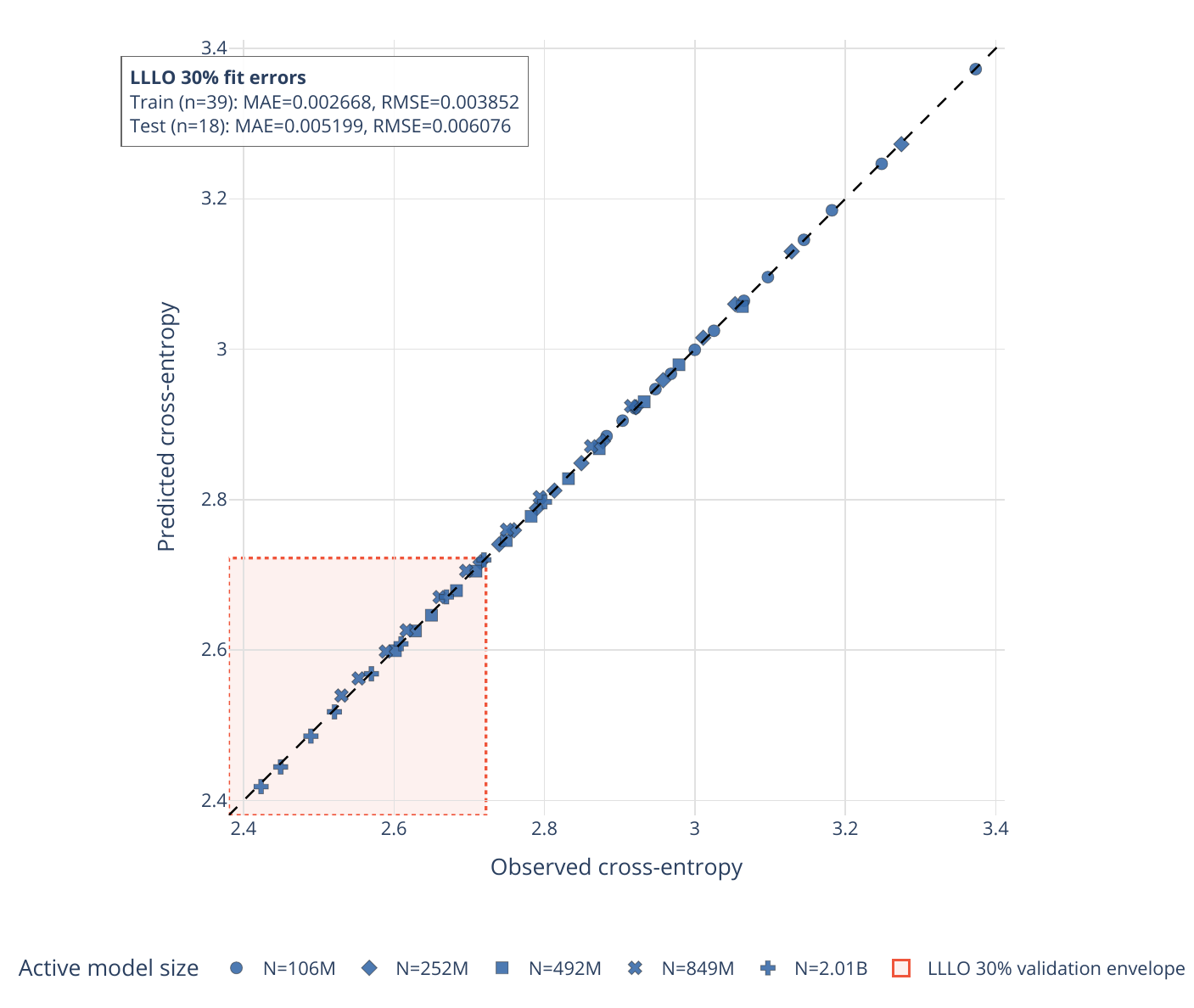}

{\small\textbf{Figure \thefigure.} \emph{Dense Transformer vP6F
LLLO$_{30}$ observed versus predicted cross-entropy.} vP6F is fitted on the
highest-loss 70\% of endpoints and evaluated on the held-out lowest-loss 30\%.
The shaded square marks the validation envelope and the dashed diagonal
denotes exact prediction. Marker shape identifies $N$.\par}
\end{minipage}
\end{center}

\begin{center}
\begin{minipage}{\textwidth}
\centering
\refstepcounter{figure}\label{fig:stem-vp6f-lllo30-observed-predicted}
\includegraphics[width=0.98\textwidth,height=0.36\textheight,keepaspectratio,trim=58.1bp 15.8bp 38.7bp 22.5bp,clip]{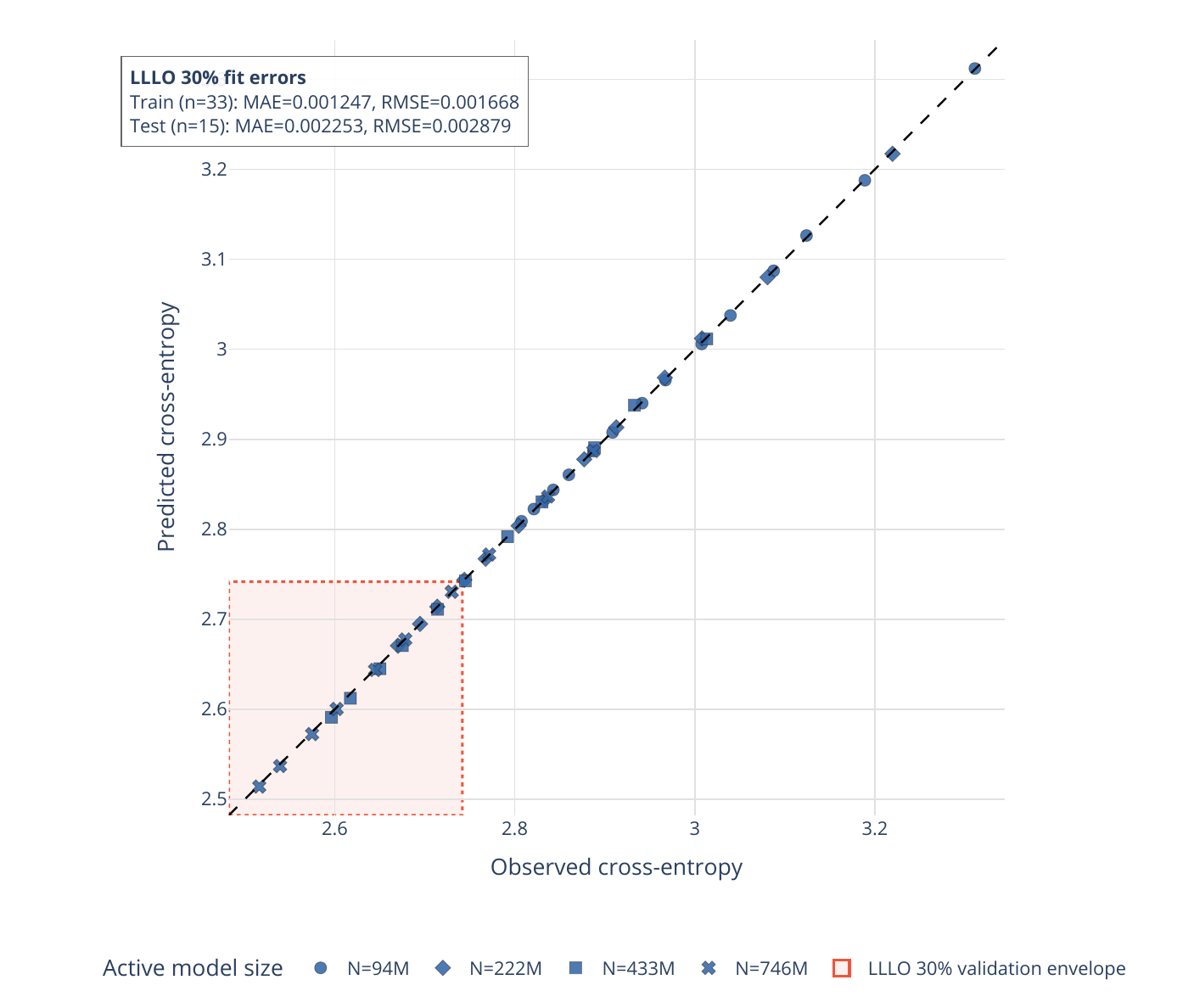}

{\small\textbf{Figure \thefigure.} \emph{STEM vP6F LLLO$_{30}$ observed
versus predicted cross-entropy.} vP6F is fitted on the highest-loss 70\% of
endpoints and evaluated on the held-out lowest-loss 30\%. The shaded square
marks the validation envelope and the dashed diagonal denotes exact
prediction. Marker shape identifies $N$.\par}
\end{minipage}
\end{center}

\subsubsection{Measured endpoints and fitted P6F and vP6F curves over tokens}

\begin{center}
\begin{minipage}{\textwidth}
\centering
\refstepcounter{figure}\label{fig:pkm-p6f-token-curves}
\includegraphics[width=0.98\textwidth,height=0.36\textheight,keepaspectratio,trim=15.8bp 14.4bp 32.1bp 21.8bp,clip]{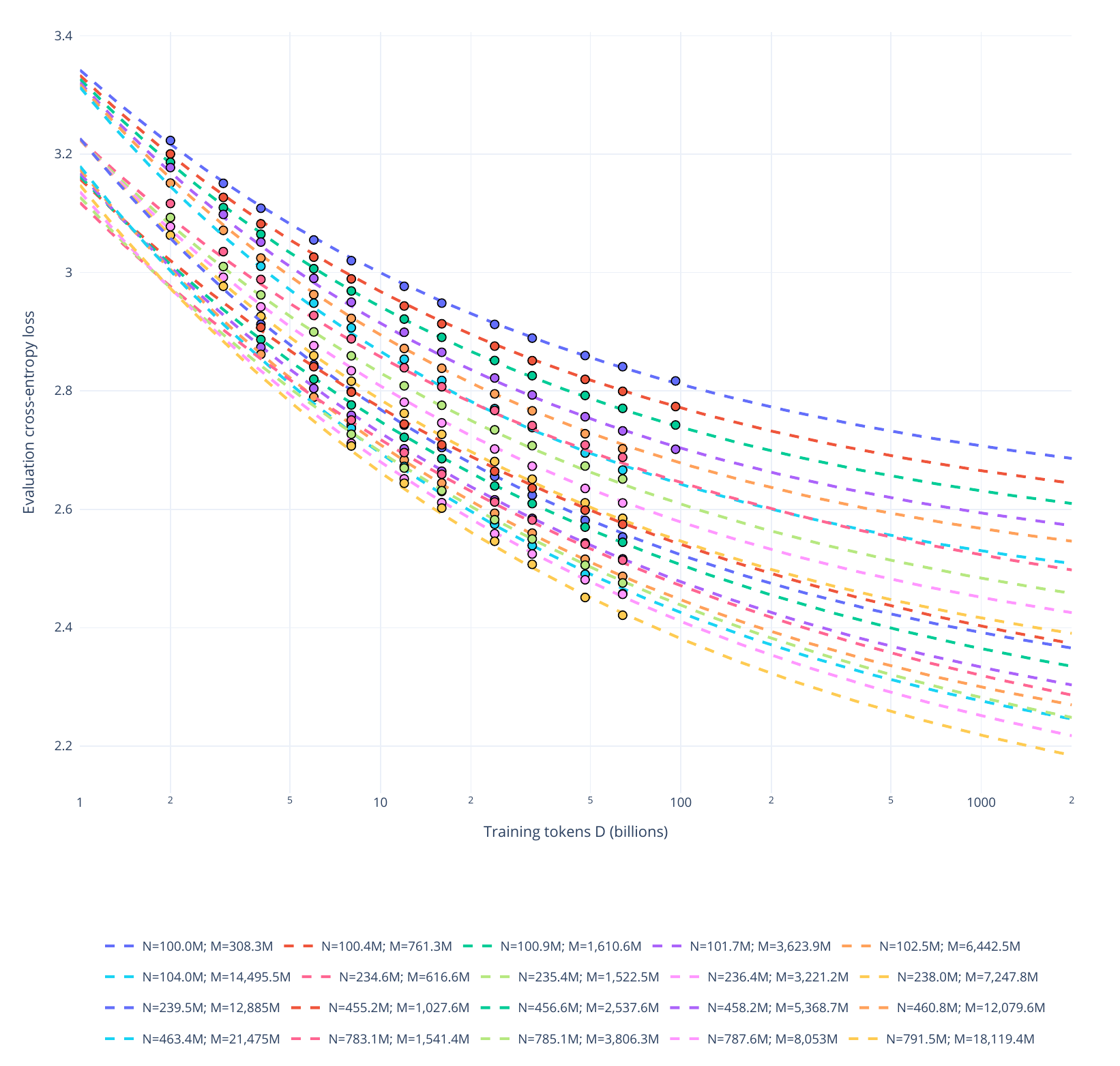}

{\small\textbf{Figure \thefigure.} \emph{PKM measured endpoints and fitted
P6F curves over tokens.} Markers show measured trained-model endpoints; dashed
lines show fitted and extrapolated P6F curves.\par}
\end{minipage}
\end{center}

\begin{center}
\begin{minipage}{\textwidth}
\centering
\refstepcounter{figure}\label{fig:moe-p6f-token-curves}
\includegraphics[width=0.98\textwidth,height=0.36\textheight,keepaspectratio,trim=15.8bp 15.1bp 32.1bp 22.5bp,clip]{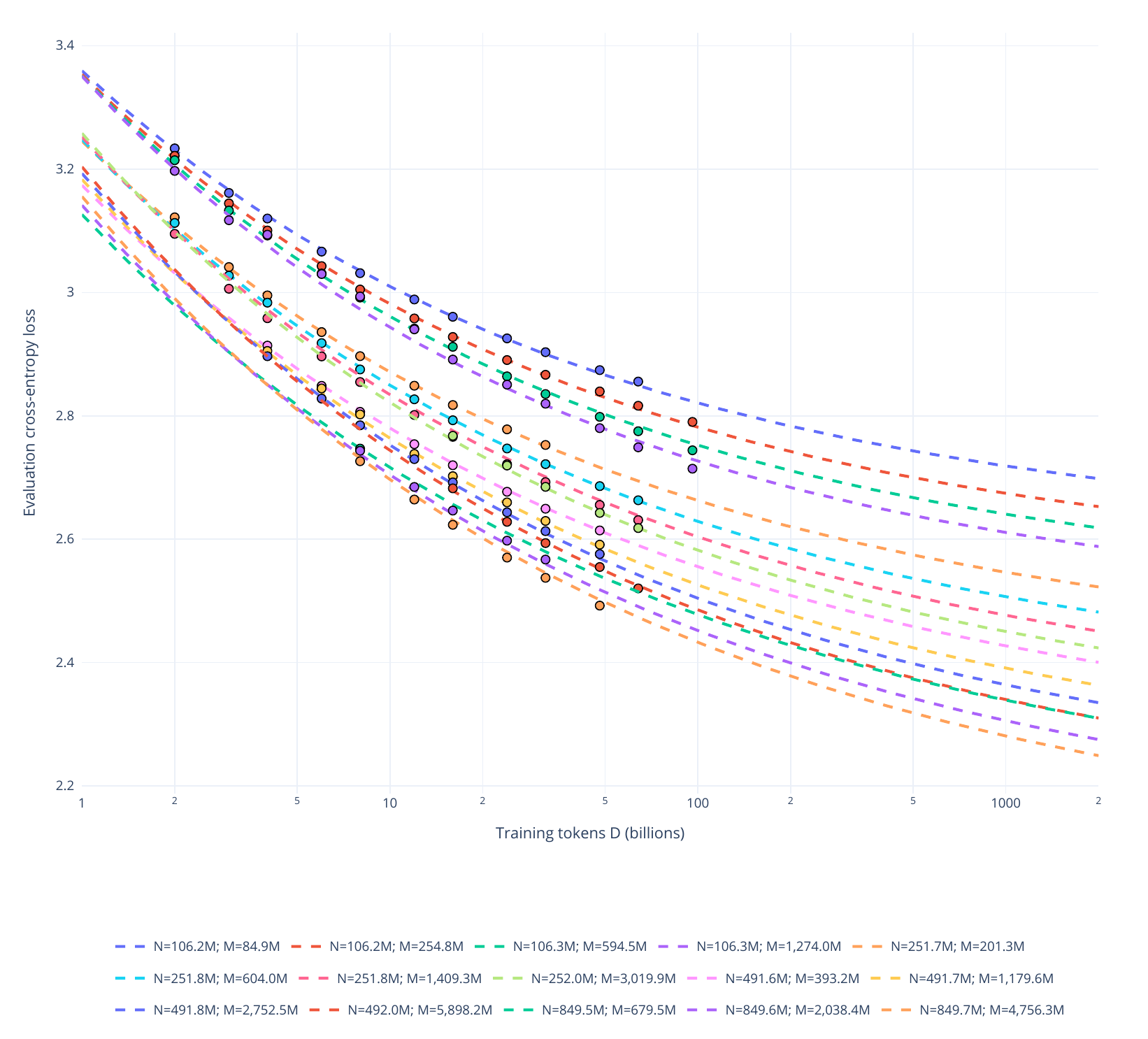}

{\small\textbf{Figure \thefigure.} \emph{MoE measured endpoints and fitted
P6F curves over tokens.} Markers show measured trained-model endpoints; dashed
lines show fitted and extrapolated P6F curves.\par}
\end{minipage}
\end{center}

\begin{center}
\begin{minipage}{\textwidth}
\centering
\refstepcounter{figure}\label{fig:vanilla-p6f-token-curves}
\includegraphics[width=0.98\textwidth,height=0.36\textheight,keepaspectratio,trim=15.8bp 14.5bp 32.1bp 22.5bp,clip]{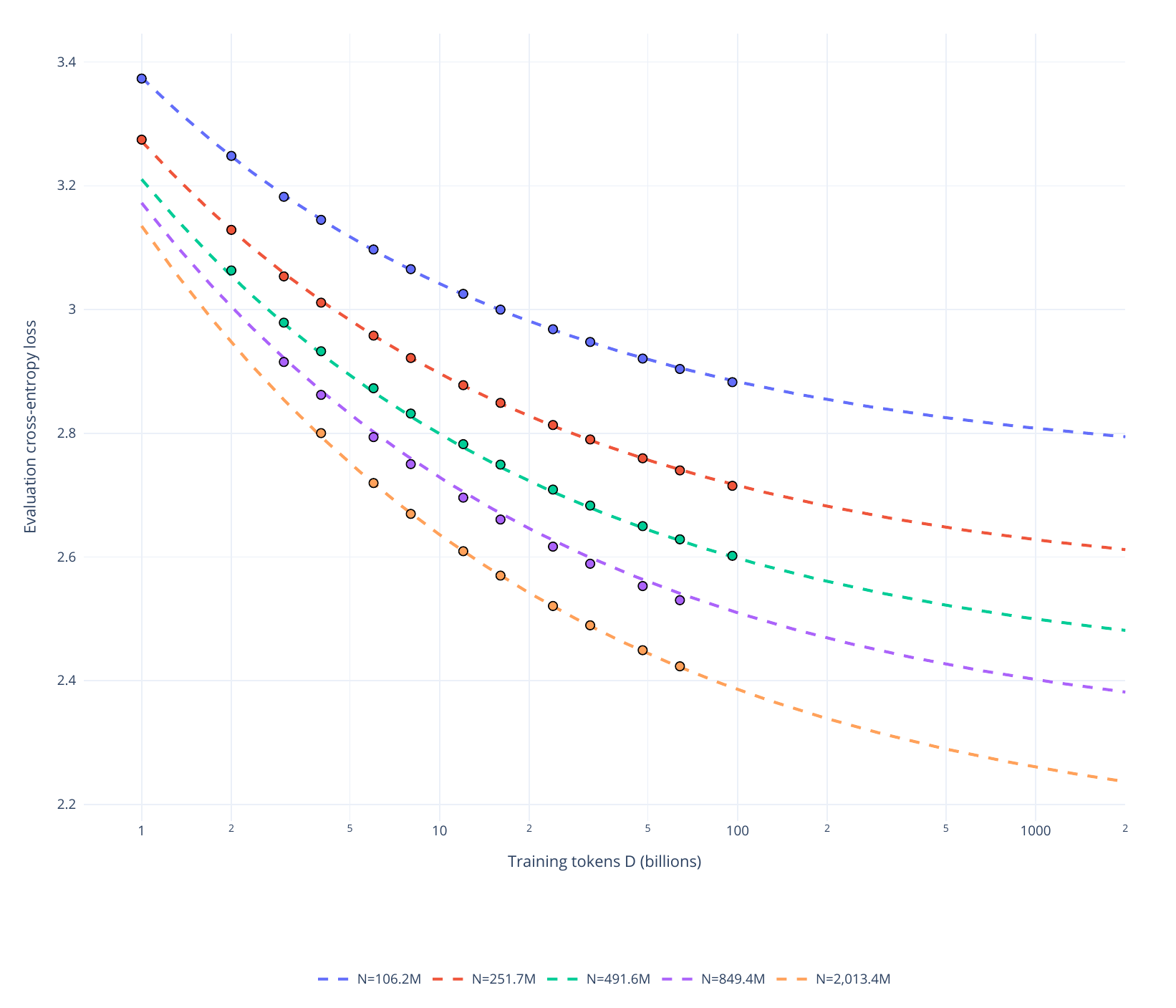}

{\small\textbf{Figure \thefigure.} \emph{Dense Transformer measured endpoints and
fitted vP6F curves over tokens.} Markers show measured trained-model endpoints;
dashed lines show fitted and extrapolated vP6F curves.\par}
\end{minipage}
\end{center}

\begin{center}
\begin{minipage}{\textwidth}
\centering
\refstepcounter{figure}\label{fig:stem-p6f-token-curves}
\includegraphics[width=0.98\textwidth,height=0.36\textheight,keepaspectratio,trim=15.8bp 15.5bp 32.1bp 22.5bp,clip]{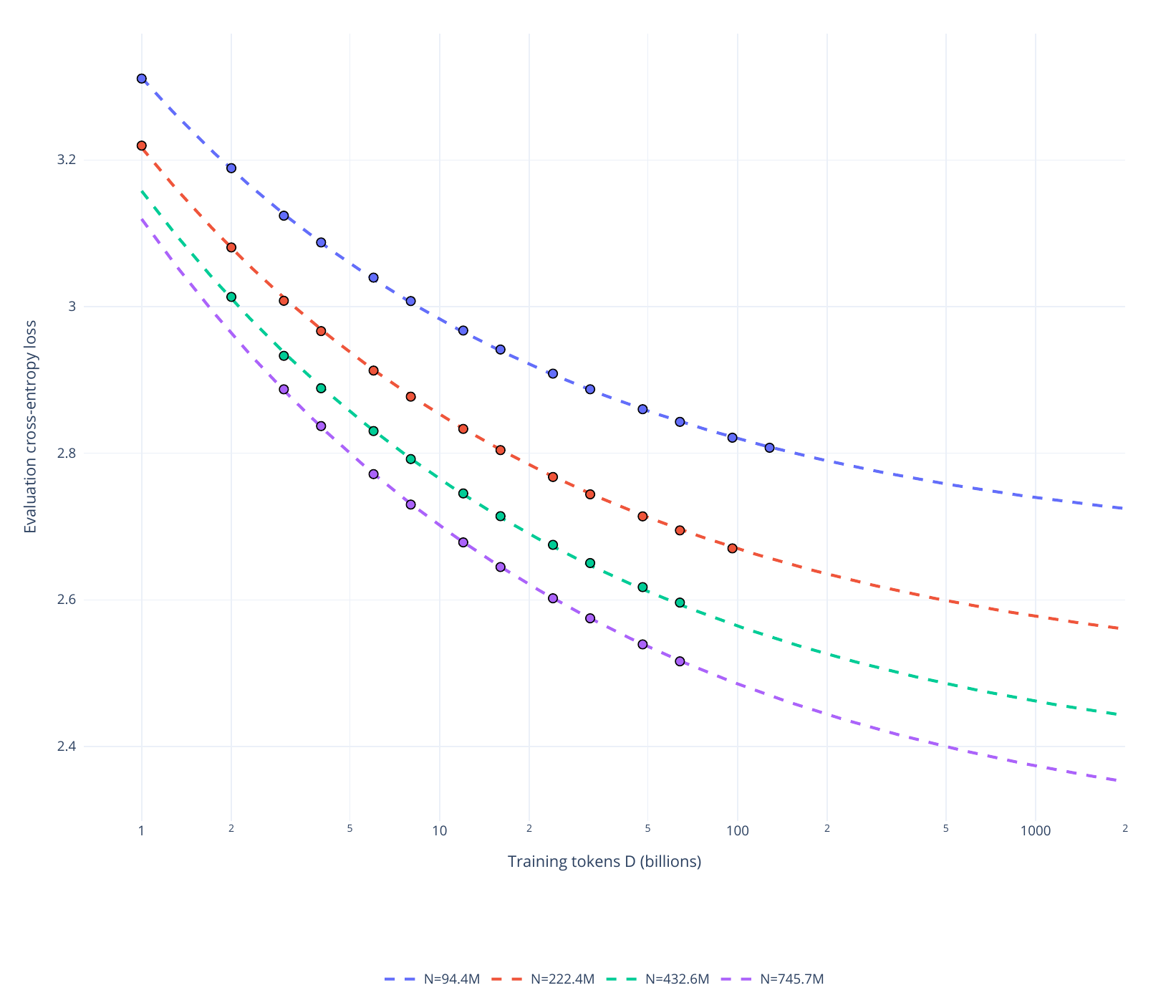}

{\small\textbf{Figure \thefigure.} \emph{STEM measured endpoints and fitted
vP6F curves over tokens.} Markers show measured trained-model endpoints; dashed
lines show fitted and extrapolated vP6F curves.\par}
\end{minipage}
\end{center}

\section{Lowest-Loss Allocation at Fixed Compute and Total Parameters}
\label{sec:lowest-loss-allocation}

Our fitted three-variable scaling law predicts final cross-entropy from the
number of active non-embedding parameters $N$, the training-token budget $D$,
and the number of inactive parameters $M$:
\begin{equation}
q_D(N,M)=\beta+\zeta\log M+\xi\log N,
\label{eq:allocation-token-exponent}
\end{equation}
\begin{equation}
\widehat L(N,D,M)
=A M^{-\kappa}N^{-\alpha}
+B N^{-\lambda}M^{-\mu}D^{-q_D(N,M)}.
\label{eq:allocation-sparse-law}
\end{equation}
Here, $\log$ is the natural logarithm and all parameter and token coordinates
use the same raw-count units as the fitted law. The coefficient vector
\begin{equation}
\boldsymbol{\theta}
=(A,B,\alpha,\beta,\kappa,\lambda,\mu,\zeta,\xi)
\end{equation}
contains all fitted parameters of the law. The following calculations
transform this function and introduce no additional fitted coefficients.

We seek the lowest predicted loss for a fixed total non-embedding parameter
count $T$ and a fixed training-compute budget $C$. The complete parameter
budget is allocated between active and inactive parameters, so
\begin{equation}
T=N+M.
\label{eq:allocation-total-parameters}
\end{equation}
Using the active fraction $r=N/T$ and inactive fraction $S=M/T=1-r$ gives
\begin{equation}
N=rT,
\qquad
M=(1-r)T,
\qquad
0<r<1.
\label{eq:allocation-fractions}
\end{equation}
We approximate total training compute by
\begin{equation}
C=6ND.
\label{eq:allocation-compute-proxy}
\end{equation}
Consequently, fixing $(T,C)$ and choosing $r$ uniquely determines all three
scaling-law coordinates:
\begin{equation}
N=rT,
\qquad
M=(1-r)T,
\qquad
D=\frac{C}{6rT}.
\label{eq:allocation-coordinate-substitution}
\end{equation}
The factor $6ND$ is an active-parameter FLOP approximation and excludes
operation costs not represented by this compute model.

Substitution of \eqref{eq:allocation-coordinate-substitution} into
\eqref{eq:allocation-token-exponent} yields the allocation-dependent
token exponent
\begin{equation}
q_D(r,T)
=\beta+\zeta\log\!\bigl((1-r)T\bigr)
+\xi\log(rT).
\label{eq:allocation-transformed-token-exponent}
\end{equation}
Substitution into the complete loss law gives a one-dimensional objective for
each fixed $(T,C)$ pair:
\begin{align}
\widehat L_{\mathrm{alloc}}(r;T,C)
:={}&\widehat L\!\left(rT,\frac{C}{6rT},(1-r)T\right)
\nonumber\\
={}&A T^{-(\alpha+\kappa)}
r^{-\alpha}(1-r)^{-\kappa}
\nonumber\\
&+B T^{-(\lambda+\mu)}
r^{-\lambda}(1-r)^{-\mu}
\left(\frac{C}{6rT}\right)^{-q_D(r,T)}.
\label{eq:allocation-transformed-loss}
\end{align}
Thus, $T$ and $C$ are held fixed and the optimization changes only how $T$ is
partitioned between $N$ and $M$. Every feasible allocation uses the complete
total parameter count through the equality $N+M=T$.

Let $\mathcal{R}(T,C)$ denote the nonempty compact active-fraction set used
for a reported calculation. It is chosen as a closed subset of the physically
feasible region with positive coordinates and a positive effective token
exponent:
\begin{equation}
\begin{aligned}
\emptyset\ne\mathcal{R}(T,C)
&\subseteq\biggl\{r\in(0,1):
N=rT>0,\ M=(1-r)T>0,\\
&\hspace{5.7em}D=\frac{C}{6rT}>0,\ q_D(r,T)>0\biggr\},\\
\mathcal{R}(T,C)&\text{ is compact}.
\end{aligned}
\label{eq:allocation-feasible-set}
\end{equation}
Any prescribed closed analysis-domain bounds on $N$, $M$, and $D$ are
incorporated into $\mathcal{R}(T,C)$. These restrictions change the feasible
set but not the transformed objective. Because $\widehat L_{\mathrm{alloc}}(r;T,C)$ is continuous on
the nonempty compact set $\mathcal{R}(T,C)$, its minimum is attained.

The globally optimal allocation is
\begin{equation}
\mathcal{R}^{*}(T,C)
=\mathop{\mathrm{argmin}}_{r\in\mathcal{R}(T,C)}\widehat L_{\mathrm{alloc}}(r;T,C).
\label{eq:allocation-argmin-set}
\end{equation}
If the minimizer is unique, we denote it by $r_{\mathrm{opt}}(T,C)$. If
multiple allocations attain the same minimum, any
$r_{\mathrm{opt}}(T,C)\in\mathcal{R}^{*}(T,C)$ gives the same optimal loss.
The lowest predicted loss is therefore
\begin{equation}
\begin{aligned}
\widehat L_{\mathrm{opt}}(T,C)
&=\widehat L_{\mathrm{alloc}}\!\left(r_{\mathrm{opt}}(T,C);T,C\right)\\
&=\min_{r\in\mathcal{R}(T,C)}\widehat L_{\mathrm{alloc}}(r;T,C).
\end{aligned}
\label{eq:allocation-optimal-loss}
\end{equation}
The associated optimal inactive fraction, active parameters, inactive
parameters, and training tokens are
\begin{equation}
S_{\mathrm{opt}}(T,C)
=1-r_{\mathrm{opt}}(T,C),
\label{eq:allocation-optimal-sparsity}
\end{equation}
\begin{equation}
N_{\mathrm{opt}}(T,C)
=r_{\mathrm{opt}}(T,C)T,
\label{eq:allocation-optimal-active}
\end{equation}
\begin{equation}
M_{\mathrm{opt}}(T,C)
=\left[1-r_{\mathrm{opt}}(T,C)\right]T,
\label{eq:allocation-optimal-inactive}
\end{equation}
\begin{equation}
D_{\mathrm{opt}}(T,C)
=\frac{C}{6r_{\mathrm{opt}}(T,C)T}.
\label{eq:allocation-optimal-tokens}
\end{equation}
Here, $S_{\mathrm{opt}}=M_{\mathrm{opt}}/T$ is the inactive-parameter fraction,
reported as model sparsity.

The optimum generally has no closed-form power-law solution. To see why,
write \eqref{eq:allocation-transformed-loss} as
$\widehat L_{\mathrm{alloc}}(r;T,C)=U(T,r)+V(T,C,r)$, where $U$ and $V$ are respectively its first and
second terms, and define
\begin{equation}
D(r)=\frac{C}{6rT},
\qquad
q_D'(r,T)=-\frac{\zeta}{1-r}+\frac{\xi}{r}.
\label{eq:allocation-q-derivative}
\end{equation}
The logarithmic derivatives of $U$ and $V$ are
\begin{equation}
\frac{1}{U}\frac{\partial U}{\partial r}
=-\frac{\alpha}{r}+\frac{\kappa}{1-r},
\label{eq:allocation-model-log-derivative}
\end{equation}
\begin{equation}
\frac{1}{V}\frac{\partial V}{\partial r}
=\frac{q_D(r,T)-\lambda}{r}
+\frac{\mu}{1-r}
-q_D'(r,T)\log D(r).
\label{eq:allocation-data-log-derivative}
\end{equation}
An interior stationary allocation must consequently satisfy
\begin{align}
0={}&U(T,r)
\left(-\frac{\alpha}{r}+\frac{\kappa}{1-r}\right)
\nonumber\\
&+V(T,C,r)
\left[
\frac{q_D(r,T)-\lambda}{r}
+\frac{\mu}{1-r}
-q_D'(r,T)\log D(r)
\right].
\label{eq:allocation-stationarity}
\end{align}
Because $q_D(r,T)$ and $D(r)$ both depend on $r$, this equation contains
coupled logarithmic and power-law terms. It may also admit multiple local
minima, so solving one stationary equation from one initial value is
insufficient to establish the global minimum.

We therefore solve \eqref{eq:allocation-optimal-loss} numerically.
For a feasible interval $[r_{\min},r_{\max}]$, the allocation is represented
in logit coordinates,
\begin{equation}
u=\log\frac{r}{1-r},
\qquad
r=\sigma(u)=\frac{1}{1+e^{-u}}.
\label{eq:allocation-logit}
\end{equation}
A uniform grid in $u$ provides denser coverage near both allocation
boundaries than a uniform grid in $r$:
\begin{equation}
u_j=u_{\min}+\frac{j}{J-1}(u_{\max}-u_{\min}),
\qquad
r_j=\sigma(u_j),
\qquad j=0,\ldots,J-1.
\label{eq:allocation-logit-grid}
\end{equation}
The objective is evaluated at every feasible $r_j$. Every sampled point whose
loss is no larger than its feasible neighbors is retained as a candidate
local minimum; a feasible boundary is retained when it is a one-sided sampled
minimum. Each interior candidate is refined independently by bounded scalar
minimization over its neighboring grid interval. Finally, the candidate with
the smallest refined loss is selected:
\begin{equation}
\widehat r_{\mathrm{opt}}(T,C)
\in\mathop{\mathrm{argmin}}_{r\in\mathcal{Q}(T,C)}\widehat L_{\mathrm{alloc}}(r;T,C),
\label{eq:allocation-numerical-selection}
\end{equation}
where $\mathcal{Q}(T,C)$ contains the refined local minima and feasible
boundary candidates. This multi-candidate procedure does not assume that
$\widehat L_{\mathrm{alloc}}(r;T,C)$ is unimodal. A solution on a feasibility boundary is recorded as
a boundary optimum because its allocation may be controlled by the imposed
analysis domain rather than by an interior stationary point. The resulting
$\widehat r_{\mathrm{opt}}(T,C)$ is the numerical approximation to
$r_{\mathrm{opt}}(T,C)$ used when evaluating the reported quantities in
Equations~\plaineqref{eq:allocation-optimal-loss}--\plaineqref{eq:allocation-optimal-tokens}.

Equations~\plaineqref{eq:allocation-optimal-loss}--\plaineqref{eq:allocation-optimal-tokens}
define the reported lowest-loss surface and its associated optimal allocation
for the fitted scaling law. PKM and MoE use the same transformation and
optimization procedure, but their independently fitted coefficient vectors
produce separate functions $\widehat L_{\mathrm{opt}}(T,C)$.

\section{Additional Results}
\label{sec:additional-results}

This appendix provides the MoE optimal-sparsity view corresponding to the PKM
result in Figure~\ref{fig:pkm-optimal-sparsity}, a fixed-budget comparison of
the sparsity selected between Dense and MoE, the coupled resource-scaling
path of Section~\ref{sec:coupled-resource-scaling} without the operational
cost, and the mean absolute error
(MAE) tables corresponding to the root-mean-square error results in
Appendix~\ref{sec:fitting-scaling-laws}.

\begin{center}
\begin{minipage}{\textwidth}
\centering
\refstepcounter{figure}\label{fig:moe-optimal-sparsity}
\includegraphics[width=0.62\linewidth,trim=13.7bp 51.9bp 13.1bp 31.0bp,clip]{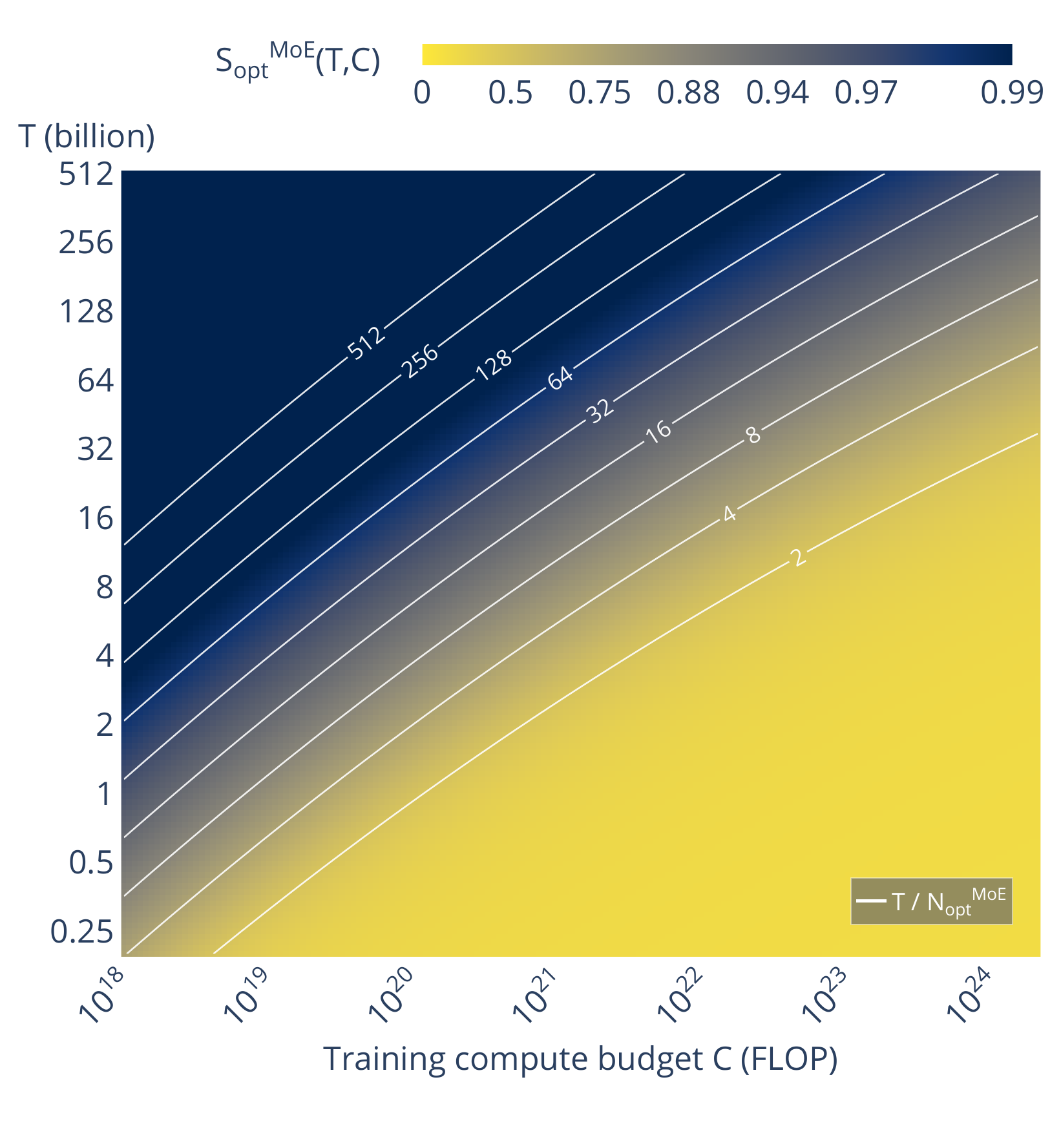}
\par\smallskip
{\small\raggedright\textbf{Figure \thefigure.} \emph{MoE optimal sparsity $S$.}
The optimal inactive fraction
$S_{\mathrm{opt}}^{\mathrm{MoE}}(T,C)=M_{\mathrm{opt}}^{\mathrm{MoE}}(T,C)/T$
is shown over training-compute and total-parameter budgets. The axes, color
mapping, and $T/N_{\mathrm{opt}}^{\mathrm{MoE}}$ contour levels match
Figure~\ref{fig:pkm-optimal-sparsity}.
\par}
\end{minipage}
\end{center}

\begin{center}
\begin{minipage}{\textwidth}
\centering
\refstepcounter{figure}\label{fig:coupled-training-only}
\begin{minipage}[t]{0.49\textwidth}
\vspace{0pt}
\centering
\refstepcounter{figurepanel}\label{fig:coupled-training-only-loss}
{\small\textbf{(\alph{figurepanel})} Lowest predicted loss
$\widehat L_{\mathrm{opt}}$\par}
\smallskip
\includegraphics[width=0.98\linewidth,trim=0bp 4bp 10bp 20bp,clip]{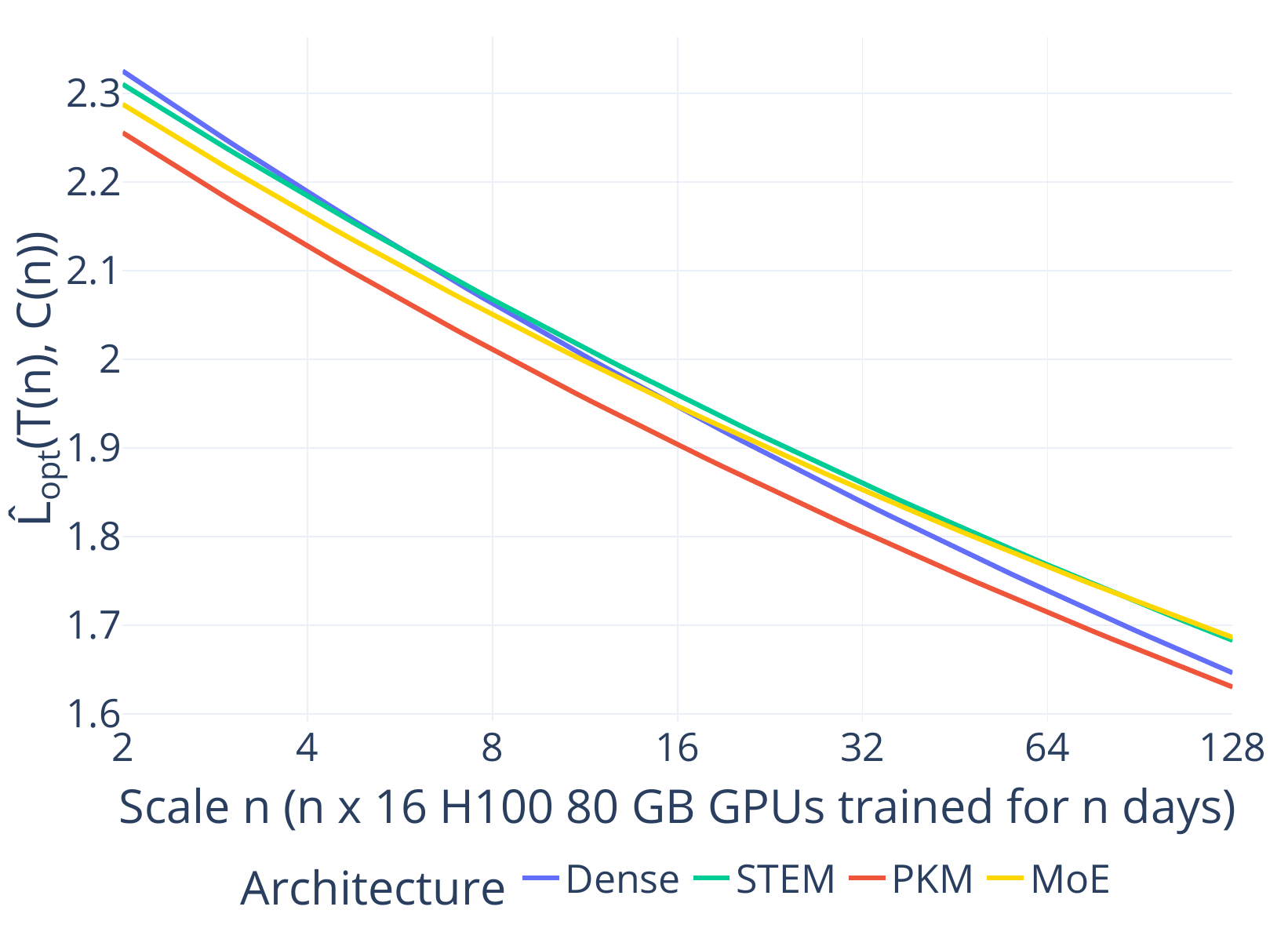}
\end{minipage}\hfill
\begin{minipage}[t]{0.49\textwidth}
\vspace{0pt}
\centering
\refstepcounter{figurepanel}\label{fig:coupled-training-only-ratio}
{\small\textbf{(\alph{figurepanel})} Optimal total-to-active ratio
$T/N_{\mathrm{opt}}$\par}
\smallskip
\includegraphics[width=0.98\linewidth,trim=0bp 4bp 10bp 20bp,clip]{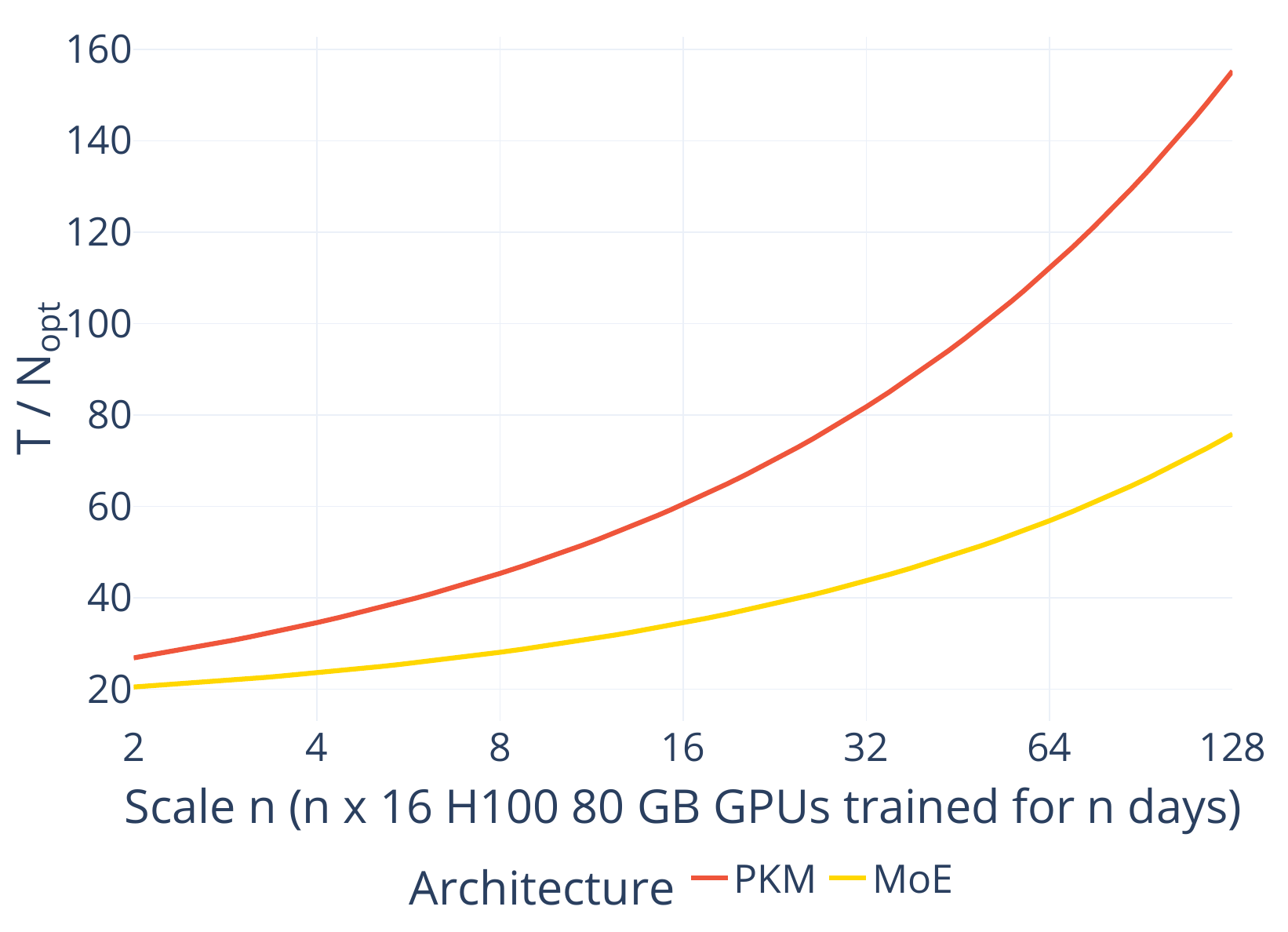}
\end{minipage}
\par\medskip
{\small\raggedright\textbf{Figure \thefigure.} \emph{Coupled GPU-and-time
scaling without the operational cost.} Same budget path as
Figure~\ref{fig:h100-coupled-scaling},
$C(n)=4.424\times10^{20}n^2$ FLOP and $T(n)=14.69n$ billion parameters, but
with $D_{\mathrm{op}}=0$, so the entire compute budget is spent on training,
$C(n)=6ND$. (a) Lowest predicted loss
$\widehat L_{\mathrm{opt}}(T(n),C(n))$ of each architecture. (b)
$T(n)/N_{\mathrm{opt}}(T(n),C(n))$ for the independently optimized PKM and
MoE allocations.\par}
\end{minipage}
\end{center}

\begin{center}
\begin{minipage}{\textwidth}
\centering
\refstepcounter{table}\label{tab:pkm-mae}
\textbf{Table \thetable.} \emph{PKM MAE (all values in $10^{-3}$ cross-entropy units).}
\\{\small Gray values come from fits that failed validity condition~(i) or (ii) (Appendix~\ref{sec:fitted-formula-results}).}%

\resizebox{\textwidth}{!}{%
\begin{tabular}{l*{13}{c}}
\hline
Formula & \multicolumn{6}{c}{Full-data and LLLO} &
\multicolumn{7}{c}{Grouped and directional holdouts} \\
\cline{2-7}\cline{8-14}
& Full & LLLO$_{10}$ & LLLO$_{20}$ & LLLO$_{30}$ & LLLO$_{40}$ & LLLO$_{50}$
& LOMO & LOSO & LOCO & CF & FT & UCF & LC \\
\hline
P0 & 29.6 & 37.4 & 45.7 & 52.2 & 57.5 & 104 & 39.6 & 35.0 & 43.5 & 51.0 & 39.9 & 84.7 & 70.7 \\
P1A & 8.69 & 15.4 & 24.9 & 28.9 & 28.4 & 31.1 & 9.53 & 13.0 & 8.69 & 9.43 & 16.3 & 15.4 & 25.6 \\
P1B & 9.67 & 21.7 & 26.3 & 27.7 & 25.5 & 26.6 & 11.6 & 15.6 & 10.7 & 11.3 & 16.2 & 16.7 & 26.1 \\
P1C & 11.5 & 13.8 & 15.4 & 25.8 & 28.2 & 36.9 & 12.9 & 15.5 & 12.1 & 12.8 & 24.6 & 22.6 & 21.3 \\
P2B & 9.67 & 21.7 & 26.3 & 27.7 & 25.5 & 26.6 & 10.8 & 15.6 & 10.7 & 11.3 & 24.6 & 16.7 & 26.1 \\
P3A & 9.41 & 20.9 & 26.3 & 27.8 & 25.8 & 26.8 & 10.3 & 14.6 & 10.1 & 10.7 & 15.9 & 16.0 & \textcolor{gray}{25.5}\textsubscript{i} \\
P3B & 8.18 & 20.8 & 23.5 & 24.1 & 22.1 & 21.2 & 10.6 & 17.4 & 10.6 & 12.2 & 12.4 & 13.7 & 5.19 \\
P3C & 7.80 & 20.2 & 22.7 & 23.3 & 21.6 & 20.9 & 9.81 & \textcolor{gray}{16.6}\textsubscript{i} & 9.71 & \textcolor{gray}{11.3}\textsubscript{i} & 12.1 & 12.6 & 7.40 \\
P4 & 5.74 & 8.59 & 9.01 & 9.52 & 9.10 & 9.17 & 8.21 & 12.1 & 8.44 & 9.25 & 9.70 & 14.2 & 5.44 \\
P5 & 5.80 & 9.03 & 9.59 & 10.2 & 9.62 & 9.08 & 7.75 & \textcolor{gray}{10.7}\textsubscript{i} & 8.40 & 9.41 & 9.50 & 14.1 & 13.7 \\
P6B & 4.73 & 7.65 & 6.96 & 7.23 & 8.01 & 8.17 & 5.98 & 9.18 & 6.77 & 7.99 & 8.74 & 12.2 & 7.42 \\
P6C & 3.92 & 4.89 & 6.65 & \textcolor{gray}{7.99}\textsubscript{i} & 8.84 & 7.68 & \textcolor{gray}{5.12}\textsubscript{i} & 9.86 & 5.30 & 6.05 & 7.24 & 9.12 & 5.65 \\
P6D & \textbf{3.48} & \textbf{3.27} & \textbf{3.65} & \textbf{4.53} & 8.94 & \textbf{7.53} & \textbf{4.30} & \textbf{6.74} & \textbf{5.08} & \textbf{5.67} & 5.91 & \textcolor{gray}{7.99}\textsubscript{i} & 2.85 \\
P6F & \textbf{3.48} & \textbf{3.27} & \textbf{3.65} & \textbf{4.53} & \textbf{5.76} & \textbf{7.53} & 4.58 & 7.10 & \textbf{5.08} & \textbf{5.67} & 5.91 & \textbf{7.77} & 2.85 \\
P6G & 5.47 & 8.95 & 8.23 & 9.38 & \textcolor{gray}{18.8}\textsubscript{ii} & 9.76 & \textcolor{gray}{23.7}\textsubscript{i,ii} & \textcolor{gray}{12.4}\textsubscript{ii} & \textcolor{gray}{55.4}\textsubscript{i,ii} & 8.60 & 8.37 & \textcolor{gray}{70.0}\textsubscript{i,ii} & \textbf{2.51} \\
P6I & 5.84 & \textcolor{gray}{54.1}\textsubscript{i,ii} & 8.17 & \textcolor{gray}{82.8}\textsubscript{i,ii} & 20.6 & 9.79 & \textcolor{gray}{35.0}\textsubscript{i,ii} & 15.7 & \textcolor{gray}{28.7}\textsubscript{i,ii} & \textcolor{gray}{12.6}\textsubscript{ii} & 8.89 & \textcolor{gray}{13.4}\textsubscript{i,ii} & 3.82 \\
P6J & \textcolor{gray}{4.79}\textsubscript{ii} & \textcolor{gray}{7.40}\textsubscript{ii} & \textcolor{gray}{6.92}\textsubscript{ii} & \textcolor{gray}{7.86}\textsubscript{ii} & \textcolor{gray}{8.84}\textsubscript{ii} & 9.29 & \textcolor{gray}{6.21}\textsubscript{ii} & \textcolor{gray}{11.0}\textsubscript{ii} & \textcolor{gray}{8.67}\textsubscript{ii} & \textcolor{gray}{7.41}\textsubscript{ii} & \textcolor{gray}{7.24}\textsubscript{ii} & \textcolor{gray}{7.94}\textsubscript{ii} & 3.39 \\
P6K & 4.15 & 5.14 & 5.31 & 6.88 & 8.22 & 9.15 & 6.30 & 8.85 & 6.42 & 6.16 & \textbf{5.66} & 8.40 & \textcolor{gray}{97.0}\textsubscript{i,ii} \\
P7 & 5.54 & 8.31 & 8.47 & 9.08 & 8.90 & 8.94 & 7.15 & 9.87 & 7.95 & 9.04 & 9.14 & 13.3 & 13.4 \\
\hline
\end{tabular}%
}
\end{minipage}
\end{center}

\begin{center}
\begin{minipage}{\textwidth}
\centering
\refstepcounter{table}\label{tab:moe-mae}
\textbf{Table \thetable.} \emph{MoE MAE (all values in $10^{-3}$ cross-entropy units).}
\\{\small Gray values come from fits that failed validity condition~(i) or (ii) (Appendix~\ref{sec:fitted-formula-results}).}%

\resizebox{\textwidth}{!}{%
\begin{tabular}{l*{13}{c}}
\hline
Formula & \multicolumn{6}{c}{Full-data and LLLO} &
\multicolumn{7}{c}{Grouped and directional holdouts} \\
\cline{2-7}\cline{8-14}
& Full & LLLO$_{10}$ & LLLO$_{20}$ & LLLO$_{30}$ & LLLO$_{40}$ & LLLO$_{50}$
& LOMO & LOSO & LOCO & CF & FT & UCF & LC \\
\hline
P0 & 16.9 & 18.5 & 19.0 & 24.1 & 40.3 & 43.0 & 22.7 & 18.7 & 25.3 & 31.5 & 34.8 & 38.2 & 44.8 \\
P1A & 7.25 & 10.5 & 13.1 & 19.6 & 28.3 & 31.8 & 9.78 & 9.33 & 9.14 & 9.94 & 11.4 & 13.9 & 26.5 \\
P1B & 7.32 & 10.7 & 12.4 & 23.0 & 25.6 & 30.1 & 11.2 & 9.92 & 8.67 & 9.44 & 11.9 & 13.4 & 20.1 \\
P1C & 8.86 & 9.22 & 13.2 & 17.6 & 35.1 & 36.0 & 10.2 & 11.0 & 11.2 & 12.3 & 15.8 & 19.3 & 21.8 \\
P2B & 7.32 & 9.22 & 12.4 & 23.0 & 25.6 & 30.1 & 8.55 & 9.92 & 9.30 & 9.44 & 15.8 & 13.4 & 21.8 \\
P3A & 7.21 & 11.0 & 12.6 & 22.3 & 25.5 & 30.0 & 8.25 & 9.86 & 8.56 & 9.22 & 11.8 & 13.4 & 25.3 \\
P3B & 5.81 & 5.30 & 7.30 & 15.5 & 14.3 & 19.5 & 8.35 & 10.5 & 8.53 & 11.2 & 6.29 & 8.05 & 14.7 \\
P3C & 5.57 & 5.37 & 6.81 & 13.9 & 13.3 & 18.9 & 7.62 & \textcolor{gray}{10.4}\textsubscript{i} & 8.42 & 11.0 & \textbf{6.10} & \textbf{7.89} & 15.9 \\
P4 & 4.50 & 5.07 & 4.86 & 7.13 & 6.87 & 7.47 & 7.35 & 7.59 & \textbf{6.90} & 9.47 & 6.62 & 10.9 & 13.6 \\
P5 & 4.65 & 5.10 & 5.08 & 8.26 & 7.46 & 8.45 & \textbf{6.17} & \textbf{5.78} & 7.14 & 9.81 & 6.82 & 11.2 & 14.4 \\
P6B & 4.43 & 4.83 & 4.58 & 6.39 & 6.60 & 7.19 & \textcolor{gray}{6.56}\textsubscript{i} & 7.39 & 7.37 & \textbf{9.00} & 6.51 & 10.5 & 13.0 \\
P6C & 4.29 & \textbf{4.01} & 4.58 & 6.89 & 7.31 & 8.10 & 6.57 & 7.03 & 7.81 & 10.0 & 6.26 & 9.11 & 14.4 \\
P6D & 4.28 & 4.28 & 4.58 & \textbf{5.99} & \textbf{6.53} & 7.19 & \textcolor{gray}{6.24}\textsubscript{i} & \textcolor{gray}{7.13}\textsubscript{i} & \textcolor{gray}{8.27}\textsubscript{i} & 10.3 & 6.16 & 9.25 & 14.2 \\
P6F & \textbf{4.27} & 4.28 & \textbf{4.50} & \textbf{5.99} & \textbf{6.53} & \textbf{6.83} & \textcolor{gray}{11.8}\textsubscript{i} & 6.98 & \textcolor{gray}{45.1}\textsubscript{i} & \textcolor{gray}{10.4}\textsubscript{i} & 6.18 & \textcolor{gray}{41.6}\textsubscript{i,ii} & 14.2 \\
P6G & 4.68 & 5.38 & 5.29 & 6.33 & 7.14 & 7.68 & 6.65 & \textcolor{gray}{25.3}\textsubscript{i,ii} & \textcolor{gray}{15.4}\textsubscript{ii} & \textcolor{gray}{41.6}\textsubscript{i,ii} & 6.59 & \textcolor{gray}{10.1}\textsubscript{ii} & \textcolor{gray}{15.8}\textsubscript{i,ii} \\
P6I & 4.67 & 5.34 & 5.30 & 6.94 & 15.7 & \textcolor{gray}{87.4}\textsubscript{i,ii} & \textcolor{gray}{15.1}\textsubscript{i,ii} & \textcolor{gray}{27.2}\textsubscript{i,ii} & \textcolor{gray}{183}\textsubscript{i,ii} & 10.1 & 6.64 & 14.0 & \textcolor{gray}{2.35}\textsubscript{i,ii} \\
P6J & 4.66 & 5.50 & 5.38 & 6.37 & 7.11 & 7.65 & 6.63 & 7.80 & \textcolor{gray}{24.9}\textsubscript{ii} & 14.7 & 6.60 & \textcolor{gray}{11.8}\textsubscript{ii} & \textbf{1.98} \\
P6K & 4.41 & 5.06 & 4.93 & 6.61 & 7.31 & 7.77 & 7.20 & 7.18 & 7.93 & 10.4 & 6.29 & 8.83 & 8.82 \\
P7 & 4.54 & 4.93 & 4.76 & 7.18 & 7.04 & 7.66 & \textcolor{gray}{5.98}\textsubscript{i} & 6.77 & 7.09 & 9.43 & 6.66 & 10.6 & 14.7 \\
\hline
\end{tabular}%
}
\end{minipage}
\end{center}

\begin{center}
\begin{minipage}{\textwidth}
\centering
\refstepcounter{table}\label{tab:vanilla-mae}
\textbf{Table \thetable.} \emph{Dense Transformer MAE (all values in $10^{-3}$ cross-entropy units).}

\resizebox{\textwidth}{!}{%
\begin{tabular}{lccccccccc}
\hline
Formula & Full & LLLO$_{10}$ & LLLO$_{20}$ & LLLO$_{30}$ & LLLO$_{40}$ & LLLO$_{50}$ & LOMO & FT & LC \\
\hline
Chinchilla & 13.8 & 41.5 & 46.0 & 46.2 & 53.5 & 51.2 & 20.9 & 22.0 & 27.2 \\
vP6F & 3.32 & \textbf{5.34} & \textbf{5.24} & \textbf{5.20} & \textbf{7.55} & \textbf{8.38} & \textbf{11.1} & \textbf{4.52} & 2.30 \\
Farseer & \textbf{3.25} & 5.50 & 5.97 & 5.46 & 8.37 & 29.1 & 46.0 & 5.21 & \textbf{1.66} \\
\hline
\end{tabular}%
}
\end{minipage}
\end{center}

\begin{center}
\begin{minipage}{\textwidth}
\centering
\refstepcounter{table}\label{tab:stem-mae}
\textbf{Table \thetable.} \emph{STEM MAE (all values in $10^{-3}$ cross-entropy units).}

\resizebox{\textwidth}{!}{%
\begin{tabular}{lccccccccc}
\hline
Formula & Full & LLLO$_{10}$ & LLLO$_{20}$ & LLLO$_{30}$ & LLLO$_{40}$ & LLLO$_{50}$ & LOMO & FT & LC \\
\hline
Chinchilla & 9.16 & 25.1 & 26.5 & 30.2 & 30.1 & 30.3 & 14.2 & 14.2 & 25.8 \\
vP6F & 1.43 & \textbf{0.85} & \textbf{2.33} & \textbf{2.25} & 3.74 & \textbf{4.43} & \textbf{4.11} & 2.19 & \textbf{1.11} \\
Farseer & \textbf{1.15} & 4.42 & 3.41 & 6.10 & \textbf{2.88} & 11.9 & 12.0 & \textbf{2.16} & 4.18 \\
\hline
\end{tabular}%
}
\end{minipage}
\end{center}

\begin{center}
\begin{minipage}{\textwidth}
\centering
\refstepcounter{table}\label{tab:joint-sparse-mae}
\textbf{Table \thetable.} \emph{Joint PKM--MoE MAE (all values in $10^{-3}$ cross-entropy units).}
\\{\small Gray values come from fits that failed validity condition~(i) or (ii) (Appendix~\ref{sec:fitted-formula-results}).}%

\resizebox{\textwidth}{!}{%
\begin{tabular}{l*{13}{c}}
\hline
Formula & \multicolumn{6}{c}{Full-data and LLLO} &
\multicolumn{7}{c}{Grouped and directional holdouts} \\
\cline{2-7}\cline{8-14}
& Full & LLLO$_{10}$ & LLLO$_{20}$ & LLLO$_{30}$ & LLLO$_{40}$ & LLLO$_{50}$
& LOMO & LOSO & LOCO & CF & FT & UCF & LC \\
\hline
P0 & 29.5 & 39.7 & 36.5 & 40.8 & 52.8 & 65.7 & 32.1 & 30.2 & 37.7 & 44.4 & 38.1 & 46.8 & 55.7 \\
P1A & 9.88 & 20.3 & 19.9 & 22.4 & 35.3 & 27.9 & 10.7 & 12.1 & 11.0 & 11.4 & 15.3 & 17.4 & 42.1 \\
P1B & 10.3 & 40.6 & 22.6 & 22.9 & 26.2 & 26.8 & 11.4 & 15.6 & 13.3 & 11.6 & 16.1 & 16.5 & 49.2 \\
P1C & 12.9 & 21.2 & 20.7 & 26.9 & 30.9 & 35.9 & 14.2 & 14.1 & 15.3 & 15.5 & 25.0 & 25.7 & 47.9 \\
P2B & 10.3 & 21.2 & 20.7 & 22.9 & 26.2 & 26.8 & 11.2 & 15.8 & 12.9 & 11.6 & 25.0 & 16.5 & 49.2 \\
P3A & 10.1 & 23.1 & \textcolor{gray}{22.4}\textsubscript{i} & 22.8 & 26.5 & \textcolor{gray}{27.1}\textsubscript{i} & \textcolor{gray}{11.0}\textsubscript{i} & \textcolor{gray}{14.8}\textsubscript{i} & \textcolor{gray}{11.2}\textsubscript{i} & 11.4 & 16.1 & 16.8 & 49.5 \\
P3B & 9.80 & 20.0 & 22.0 & 21.5 & 24.8 & 25.5 & 11.2 & 17.3 & 10.9 & 12.8 & 14.6 & 14.3 & 37.5 \\
P3C & 9.58 & 21.7 & 21.2 & 21.0 & 24.5 & 25.5 & \textcolor{gray}{10.9}\textsubscript{i} & \textcolor{gray}{18.1}\textsubscript{i} & 10.7 & 12.7 & 14.7 & 14.8 & 38.8 \\
P4 & 8.11 & 16.0 & 12.4 & 11.1 & 12.1 & 11.9 & 9.82 & 14.6 & 10.8 & 11.4 & 12.5 & 15.9 & 36.9 \\
P5 & 8.17 & 15.2 & 12.8 & 11.3 & 12.2 & 11.7 & 9.84 & 11.6 & 10.9 & 11.5 & 12.5 & 15.5 & 37.1 \\
P6B & 7.47 & 17.6 & 9.52 & 10.0 & 10.7 & 10.5 & 8.79 & 9.22 & \textcolor{gray}{9.94}\textsubscript{i} & 10.5 & 11.3 & 14.4 & 34.9 \\
P6C & 6.89 & 8.66 & 9.27 & 8.80 & 10.1 & 9.96 & \textcolor{gray}{8.01}\textsubscript{i} & \textcolor{gray}{12.3}\textsubscript{i} & \textcolor{gray}{8.53}\textsubscript{i} & \textcolor{gray}{9.66}\textsubscript{i} & 9.38 & \textcolor{gray}{12.6}\textsubscript{i} & 33.5 \\
P6D & \textbf{6.47} & \textbf{6.46} & \textbf{6.51} & \textcolor{gray}{7.30}\textsubscript{i} & 10.7 & \textbf{9.70} & \textcolor{gray}{7.51}\textsubscript{i} & \textbf{7.98} & \textbf{8.01} & \textbf{9.20} & \textbf{8.61} & \textcolor{gray}{12.0}\textsubscript{i} & 37.5 \\
P6F & \textbf{6.47} & \textbf{6.46} & \textbf{6.51} & \textbf{7.30} & \textbf{8.91} & \textbf{9.70} & \textbf{7.51} & \textbf{7.98} & \textbf{8.01} & \textcolor{gray}{9.17}\textsubscript{i} & \textbf{8.61} & 11.9 & \textcolor{gray}{26.5}\textsubscript{i} \\
P6G & 7.62 & 11.1 & 10.1 & 11.1 & 12.5 & \textcolor{gray}{170}\textsubscript{i,ii} & 9.38 & 12.4 & 12.0 & \textcolor{gray}{19.3}\textsubscript{i,ii} & 10.7 & 14.9 & \textbf{7.96} \\
P6I & 8.33 & 11.5 & 10.6 & 12.8 & \textcolor{gray}{122}\textsubscript{i,ii} & 20.8 & 10.8 & 15.3 & \textcolor{gray}{19.1}\textsubscript{i,ii} & 13.8 & 11.3 & \textcolor{gray}{39.4}\textsubscript{i} & 24.3 \\
P6J & 7.77 & 11.2 & 11.4 & 15.4 & 13.0 & 12.1 & 9.45 & 14.2 & \textcolor{gray}{14.0}\textsubscript{ii} & \textcolor{gray}{12.6}\textsubscript{ii} & 10.8 & \textcolor{gray}{16.6}\textsubscript{ii} & 21.0 \\
P6K & 7.23 & 7.77 & 7.87 & 11.6 & 12.6 & 12.5 & 8.71 & 10.9 & 9.09 & 9.71 & 8.96 & \textbf{11.5} & 14.5 \\
P7 & 8.01 & 14.3 & 11.8 & 10.8 & 11.9 & 11.4 & 9.54 & 10.8 & 10.7 & 11.2 & 12.0 & 15.2 & 36.4 \\
\hline
\end{tabular}%
}
\end{minipage}
\end{center}

\section{Roofline Analysis Formulas: Derivations and Parameters}
\label{sec:appendix-hardware}

This appendix derives the quantities $F$, $b_\text{HBM}$, $b_\text{DRAM}$,
and $b_\text{flash}$ for each architecture and regime, and documents the
hardware parameters used in Section~\ref{sec:hardware-implications}.
Table~\ref{tab:notation} defines all symbols used.

\begin{table}[ht]
\centering
\caption{Notation used in this appendix.}
\label{tab:notation}
\small
\begin{tabular}{ll}
\toprule
Symbol & Definition \\
\midrule
\multicolumn{2}{l}{\textit{Architecture}} \\
$d_\text{model}$ & Model width \\
$d_\text{ff}$ & FFN intermediate size \\
$d_\text{ff}^\text{exp}$ & Expert intermediate size $= d_\text{ff}/k$ \\
$d_v$ & PKM value-vector dimension \\
$L$ & Number of layers \\
$k$ & Top-$k$ activated (MoE) \\
$E$ & Number of experts (MoE) \\
$P_e$ & Params per expert $= 3d_\text{model}\,d_\text{ff}^\text{exp}$ \\
$n_\text{heads}$ & PKM router heads \\
$d_q$ & Query dim per router head $= \text{rank}\cdot d_\text{head}$ \\
$d_\text{head}$ & Per-axis key/query dim (PKM router) \\
$\text{rank}$ & Product-key factorisation rank \\
$n_\text{per\_axis}$ & Keys per axis (PKM router) \\
$k_\text{PKM}$ & Values retrieved per PKM head \\
$k_\text{total}$ & Total PKM retrievals $=n_\text{heads}k_\text{PKM}$ \\
$\Delta$ & PKM layer interval: 1 every $\Delta$ layers \\
\midrule
\multicolumn{2}{l}{\textit{Sequence lengths and counts}} \\
$W$ & Processing window (tokens per forward pass) \\
$C$ & Context length (KV cache size) \\
$\tau$ & Tokens per GPU per training step (batch $\times$ seq.\ length) \\
$D$ & Total training token count (dataset) \\
\midrule
\multicolumn{2}{l}{\textit{Quantization}} \\
$q_\text{act}$ & Bytes per activation element (training, bf16) \\
$q_w$ & Bytes per stored weight; precision depends on regime (e.g.\ int4 on mobile) \\
$q_\text{kv}$ & Bytes per KV-cache element (mobile, int8) \\
\midrule
\multicolumn{2}{l}{\textit{Hardware}} \\
$F_\text{peak}$ & Peak compute throughput \\
$\alpha_F$ & Compute-throughput efficiency factor \\
$G_\text{HBM}$ & HBM capacity per GPU \\
$\mathcal{B}_\text{HBM}$ & Peak HBM bandwidth \\
$\alpha_\text{HBM}$ & HBM bandwidth efficiency factor \\
$G_\text{DRAM}$ & Mobile DRAM capacity \\
$\alpha_G^\text{DRAM}$ & Usable mobile-DRAM capacity fraction \\
$\mathcal{B}_\text{DRAM}$ & Peak DRAM bandwidth \\
$\alpha_\text{DRAM}$ & DRAM bandwidth efficiency factor \\
$\mathcal{B}_\text{NVLink}$ & Peak NVLink bandwidth \\
$\alpha_\text{NVLink}$ & NVLink bandwidth efficiency factor \\
$\Theta_\text{GPU}$ & Training throughput in tokens/s/GPU $=\tau/t_\text{step}$ \\
$\mathcal{B}_\text{flash}$ & Peak flash bandwidth \\
$\alpha_\text{flash}$ & Flash bandwidth efficiency factor \\
$t_\text{random}$ & Cold random-read latency per flash request \\
$Q_\text{flash}$ & Maximum outstanding flash requests \\
\bottomrule
\end{tabular}
\end{table}

\subsection{Step-Time Roofline Model}

\paragraph{Training at Scale.}

The available training resource is one GB200 NVL72 node, so EP, TP, and PP
communication all use $\mathcal{B}_\text{NVLink}$ and placements requiring more
than 72 GPUs are invalid.
\begin{equation}
  t_\text{step} = \max\!\bigl(t_\text{compute},\; t_\text{HBM}\bigr) + t_\text{A2A},
  \qquad
  t_\text{compute} = \frac{F}{\alpha_F F_\text{peak}}, \quad
  t_\text{HBM} = \frac{b_\text{HBM}}{\alpha_\text{HBM}\mathcal{B}_\text{HBM}}, \quad
  t_\text{A2A} = \frac{b_\text{A2A}}{\alpha_\text{NVLink}\mathcal{B}_\text{NVLink}},
\end{equation}
where $t_\text{compute}$ is time to perform all FLOPs at sustained throughput
$\alpha_F F_\text{peak}$,
$t_\text{HBM}$ is time to stream weights from HBM, and $t_\text{A2A}$
is GPU-to-GPU communication time, including EP all-to-all,
TP All-Reduce, and PP boundary transfers, which does not overlap with compute.
For a per-GPU training step containing $\tau$ tokens, the token throughput is
\begin{equation}
  \Theta_\text{GPU} = \frac{\tau}{t_\text{step}}
  \qquad \text{tokens/s/GPU}.
  \label{eq:training-throughput-gpu}
\end{equation}

\paragraph{Inference at the Edge.}
\begin{equation}
  t_\text{step} = \max\!\bigl(t_\text{compute},\; t_\text{DRAM},\; t_\text{flash}\bigr),
  \qquad
  t_\text{DRAM} = \frac{b_\text{DRAM}}{\alpha_\text{DRAM}\mathcal{B}_\text{DRAM}}, \quad
  t_\text{flash}
    = t_\text{access}
      + \frac{b_\text{flash}}{\alpha_\text{flash}\mathcal{B}_\text{flash}},
\end{equation}
where $t_\text{DRAM}$ streams static parameters and KV cache, while
$t_\text{flash}$ pages dynamically selected parameters.
$t_\text{access}$ captures queued cold random-read latency and is derived
for MoE and PKM below; the bandwidth term streams the aggregate contiguous
payload $b_\text{flash}$. The efficiency factors convert peak hardware rates
into sustained effective rates. There is no communication term because
inference runs on a single device.

\subsection{FLOPs per Step (\texorpdfstring{$F$}{F})}

\subsubsection{FLOPs per Layer}
\label{sec:flops-per-layer}

\paragraph{Attention.}
With grouped-query attention, let $n_\text{kv}$ be the number of key/value
heads and $d_\text{head}$ the per-head dimension. FLOPs per token, per layer:
\begin{equation}
  F_\text{attn} = 2\bigl(\underbrace{2d_\text{model}^2
    + 2d_\text{model}n_\text{kv}d_\text{head}}_{\text{Q, K, V, O projections}}
    + \underbrace{2d_\text{model}\cdot C}_{\text{QK\textsuperscript{T} + AV}}\bigr).
\end{equation}
This reduces to the usual $4d_\text{model}^2$ projection term when
$n_\text{kv}d_\text{head}=d_\text{model}$.

\paragraph{FFN Dense.}
Three projections (gate, up, down) of size $d_\text{model} \times d_\text{ff}$:
\begin{equation}
  F_\text{FFN}^\text{Dense} = 2 \times 3\,d_\text{model}\,d_\text{ff} = 6\,d_\text{model}\,d_\text{ff}.
\end{equation}

\paragraph{FFN MoE.}
Top-$k$ experts, each with intermediate size $d_\text{ff}^\text{exp} = d_\text{ff}/k$
(ensuring matched active compute):
\begin{equation}
  F_\text{FFN}^\text{MoE}
    = 2 \cdot k \cdot 3\,d_\text{model}\,d_\text{ff}^\text{exp}
    = 2 \cdot k \cdot 3\,d_\text{model}\,\frac{d_\text{ff}}{k}
    = 6\,d_\text{model}\,d_\text{ff}
    = F_\text{FFN}^\text{Dense}.
\end{equation}

\paragraph{FFN PKM.}
PKM has \emph{no up projection}: a \textbf{projection} (gate + down, size $d_\text{model} \times d_v$)
and a \textbf{router} (query projection + product-key lookup over $n_\text{heads}$ heads):
\begin{equation}
  F_\text{FFN}^\text{PKM}
    = \underbrace{4d_\text{model}d_v}_{\text{gate+down}}
    + \underbrace{2\,n_\text{heads}(d_\text{model}\,d_q + \text{rank}\cdot n_\text{per\_axis}\,d_\text{head})}_{\text{router}}.
\end{equation}
The retrieved value vectors
($k_\text{total}=n_\text{heads}k_\text{PKM}$ values of
dimension $d_v$) are loaded from flash but contribute negligible multiply-adds.

\subsubsection{FLOPs per GPU}
\label{sec:per-gpu-compute}

We derive the per-GPU FLOPs from the per-layer FLOPs above.

\paragraph{FLOPs per GPU per layer.}
Column-wise Tensor Parallelism (TP)
shards PKM value storage. Expert Parallelism (EP) shards expert ownership across GPUs. Given $\tau$ defined as the per-GPU token count, balanced routing leaves each GPU processing
approximately $\tau$ expert-token assignments.
With a batch of $\tau$ tokens, the average FLOPs per GPU per layer are:
\begin{align}
  f_\text{GPU}^\text{MoE}
    &= \tau\!\left(F_\text{attn} + F_\text{FFN}^\text{MoE}\right),
  \label{eq:fgpu-moe} \\
  f_\text{GPU}^\text{PKM}
    &= \tau\!\left(F_\text{attn} + \frac{F_\text{FFN}^\text{PKM}}{\Delta}
       + \left(1-\frac{1}{\Delta}\right)F_\text{FFN}^\text{Dense}\right).
  \label{eq:fgpu-pkm}
\end{align}

\paragraph{FLOPs per GPU per step.}
Pipeline Parallelism (PP) assigns $L/\mathrm{PP}$ consecutive layers to each
GPU group. Training requires a forward pass plus backward activation and weight
gradients, which we approximate as $m_\text{train}=3$ times the forward FLOPs.
This factor is the usual matmul accounting: each weight-using linear layer does
one GEMM in the forward pass and two GEMMs in backward. The total per-step
FLOPs per GPU are therefore:
\begin{align}
  F_\text{GPU}^\text{MoE}
    &= m_\text{train}\frac{L}{\mathrm{PP}}\cdot\tau\!\left(F_\text{attn} + F_\text{FFN}^\text{MoE}\right),
  \label{eq:Fgpu-moe} \\
  F_\text{GPU}^\text{PKM}
    &= m_\text{train}\frac{L}{\mathrm{PP}}\cdot\tau\!\left(F_\text{attn} + \frac{F_\text{FFN}^\text{PKM}}{\Delta}
       + \left(1-\frac{1}{\Delta}\right)F_\text{FFN}^\text{Dense}\right).
  \label{eq:Fgpu-pkm}
\end{align}

\subsection{EP/TP/PP Computation and Placement}
\label{sec:parallelism-computation-placement}

All parallelism choices are constrained by the single NVL72 resource envelope.
For MoE, each pipeline stage uses an EP group, so the simulated placement must
satisfy
\begin{equation}
  \mathrm{PP}\cdot\mathrm{EP} \leq 72.
\end{equation}
To fit all $E$ experts within GPU memory, EP must also satisfy
\begin{equation}
  \text{EP} \;\geq\; \frac{E \cdot P_e \cdot q_w \cdot L_\text{PP}}{G_\text{HBM}},
  \label{eq:ep-lower-bound}
\end{equation}
where $L_\text{PP}=L/p$ and the numerator is the total expert weight bytes per
pipeline stage. Since
$P_e = 3d_\text{model}d_\text{ff}/k \propto d_\text{model}^2$ and
$L_\text{PP} \propto d_\text{model}$ in a balanced-depth configuration,
\begin{equation}
  \text{EP} \;\propto\; \frac{E \cdot d_\text{model}^3}{G_\text{HBM}}.
  \label{eq:ep-scaling}
\end{equation}
For PKM, each pipeline stage uses a TP group, giving the analogous constraint
\begin{equation}
  \mathrm{PP}\cdot\mathrm{TP} \leq 72.
\end{equation}
Operating points that violate these bounds fall outside the single-NVL72
resource envelope and are excluded from the roofline curves.

Rather than jointly sweeping EP/TP and PP, the simulator uses the monotonic
structure of the roofline. Candidate pipeline degrees must divide the layer
count and leave at least one PKM interval per pipeline stage,
\begin{equation}
  \mathcal{P}(L)=\{p:p\mid L,\;p\leq72,\;L/p\geq\Delta\}.
\end{equation}
This avoids imbalanced stages in which some stages contain a PKM layer and
others do not. For each $p\in\mathcal{P}(L)$, the simulator first sets the
sharding degree to the smallest value that satisfies the memory constraint,
then selects the pipeline degree with maximum tokens/s/GPU. For MoE,
\begin{align}
  \mathrm{EP}_{\min}(p)
    &= \min_c\left\{c:
      \left\lceil\frac{E}{c}\right\rceil P_e q_w\frac{L}{p}
      \leq B_\text{avail}^\text{MoE}(p)\right\}, \\
  p_\text{MoE}^\star
    &= \operatorname*{arg\,max}_{p\in\mathcal{P}(L):\;p\,\mathrm{EP}_{\min}(p)\leq72}
      \Theta_\text{GPU}^\text{MoE}\bigl(p,\mathrm{EP}_{\min}(p)\bigr).
\end{align}
For PKM,
\begin{align}
  L_\text{PKM}^{\max}(p)
    &= \max_{s\in\{1,\ldots,p\}}\#\{\text{PKM layers assigned to stage }s\}, \\
  \mathrm{TP}_{\min}(p)
    &= \left\lceil
      \frac{B_\text{bank}\,L_\text{PKM}^{\max}(p)}{B_\text{avail}^\text{PKM}(p)}
      \right\rceil, \\
  p_\text{PKM}^\star
    &= \operatorname*{arg\,max}_{p\in\mathcal{P}(L):\;p\,\mathrm{TP}_{\min}(p)\leq72}
      \Theta_\text{GPU}^\text{PKM}\bigl(p,\mathrm{TP}_{\min}(p)\bigr).
\end{align}
Here $B_\text{avail}(p)$ is the per-GPU HBM left after static parameters,
activations, and a reserve margin, and $B_\text{bank}$ is the per-layer PKM
value-bank footprint before TP sharding. We use the maximum number of PKM banks
resident on any pipeline stage because HBM capacity is a per-GPU constraint.
Thus EP/TP are never increased
beyond the minimum needed to fit memory; PP is the only enumerated degree of
freedom.

We include pipeline bubbles using $M=16$ pipeline microbatches. The pipeline
efficiency for candidate pipeline degree $p$ is
\begin{equation}
  \eta_\text{PP}(p) = \frac{M}{M+p-1}, \qquad M=16,
\end{equation}
and the reported throughput applies this multiplicative factor to the ideal
steady-state tokens/s/GPU.

EP reduces resident expert weights per GPU, while attention and the active
expert computation for the local per-GPU token batch remain unsharded in the
per-GPU step model. In PKM, TP is used to fit the value bank in memory and to
synchronise retrieved representations, but it does not divide the PKM layer
FLOPs. Attention, PKM compute, and the dense FFN layers between PKM layers are
therefore replicated within each pipeline stage.
PP reduces the number of resident layers from $L$ to $L/\mathrm{PP}$ and adds
the pipeline activation/gradient traffic included in $b_\text{PP}$.

\subsection{Memory and Communication Bytes per Step}

\subsubsection{\texorpdfstring{$b_\text{HBM}$}{b\_HBM} - Training at Scale}

In a single forward pass every active weight is streamed from HBM once.
Training additionally requires reading weights in the backward pass and
reading/writing gradients and optimizer states. We approximate these streams
with the same effective bytes-per-element $q_w$:
\begin{equation}
  b_\text{HBM} \approx 4\,\frac{N}{\mathrm{PP}}\,q_w,
\end{equation}
where $N$ is the full-model active-parameter count used throughout the paper
and $q_w$ is the effective byte width for training weight/gradient-state
traffic. Because PP assigns each GPU group roughly $1/\mathrm{PP}$ of the
layers, the per-GPU HBM traffic entering the step-time model scales as
$N/\mathrm{PP}$. Activation tensors are included in the HBM capacity budget but
not in this bandwidth term; this keeps $b_\text{HBM}$ focused on repeated
weight/gradient-state streams. For matched active $N$, this baseline HBM term
is comparable across Dense, MoE, and PKM.

\subsubsection{\texorpdfstring{$b_\text{A2A}$}{b\_A2A} - Training at Scale}
\label{sec:bnvlink}

MoE performs all-to-all collectives to dispatch tokens across EP GPUs and
combine expert outputs, in both forward and backward passes. The per-layer
expert-parallel traffic is
\begin{equation}
  b_\text{EP}^\text{MoE}
    = 4\,k\,\frac{\mathrm{EP}-1}{\mathrm{EP}}\,\tau\,d_\text{model}\,q_\text{act}.
\end{equation}

PKM synchronises the retrieved value representation via an All-Reduce over the
TP group, in both forward and backward passes. The per-PKM-layer
tensor-parallel traffic is
\begin{equation}
  b_\text{TP}^\text{PKM}
    = 2\,\frac{\mathrm{TP}-1}{\mathrm{TP}}\,\tau\,d_\text{model}\,q_\text{act}.
\end{equation}
PP adds forward activation transfers and backward gradient transfers across
pipeline boundaries. We use the maximum per-stage traffic across pipeline
stages, counting one forward/backward pair for each adjacent pipeline neighbor,
\begin{equation}
  b_\text{PP} = 2\,\min(\mathrm{PP}-1,2)\,\tau\,d_\text{model}\,q_\text{act}.
\end{equation}
The per-GPU step-time model therefore uses
\begin{align}
  b_\text{A2A}^\text{MoE}
    &= \frac{L}{\mathrm{PP}}\,b_\text{EP}^\text{MoE} + b_\text{PP}, \\
  b_\text{A2A}^\text{PKM}
    &= \frac{L}{\mathrm{PP}\Delta}\,b_\text{TP}^\text{PKM} + b_\text{PP}.
\end{align}
Ignoring the common PP boundary term, their ratio is
\begin{equation}
  \frac{b_\text{A2A}^\text{PKM}}{b_\text{A2A}^\text{MoE}}
  \approx
  \frac{1}{2k\Delta}\,
  \frac{(\mathrm{TP}-1)/\mathrm{TP}}{(\mathrm{EP}-1)/\mathrm{EP}},
\end{equation}
so PKM has substantially lower communication whenever TP and EP are comparable;
for $k{=}1$ and $\Delta{=}4$, the non-PP term is about $8\times$ smaller at
large EP/TP.

\paragraph{Throughput degradation with EP and TP.}
Since $t_\text{A2A}$ adds directly to step time, tokens/s/GPU degrades for all
$\mathrm{EP}>1$: as EP grows, per-GPU expert storage falls while $t_\text{A2A}$
grows toward
$4k\tau d_\text{model}\,q_\text{act}/(\alpha_\text{NVLink}\mathcal{B}_\text{NVLink})$
per layer.
Since EP is forced upward by the memory constraint
(Eq.~\ref{eq:ep-lower-bound}), large models incur compounding communication
overhead and training throughput collapses.

For PKM, growing the value bank increases capacity without increasing
$b_\text{TP}^\text{PKM}$ at a fixed TP degree: the All-Reduce payload depends on
the active representation size, not on the number of stored values. If the bank
no longer fits in HBM, the memory constraint can still force a larger TP degree,
which increases the factor $(\mathrm{TP}-1)/\mathrm{TP}$. At
$\mathrm{TP}{=}1$, this factor is zero and no TP communication occurs.

\subsubsection{\texorpdfstring{$b_\text{DRAM}$}{b\_DRAM} - Inference at the Edge}

Attention parameters and KV cache are held in DRAM. Both MoE and PKM also
keep their router weights in DRAM. In a PKM model, only $L/\Delta$ layers
are PKM; the remaining $L(1-1/\Delta)$ regular layers keep their dense FFN
in DRAM as well:
\begin{align}
  b_\text{DRAM}^\text{Dense}
    &= L\bigl(\underbrace{4d_\text{model}^2\,q_w}_{\text{attn weights}}
      + \underbrace{2d_\text{model}\cdot C\,q_\text{kv}}_{\text{KV cache}}
      + \underbrace{3\,d_\text{model}\,d_\text{ff}\,q_w}_{\text{FFN weights}}\bigr), \\
  b_\text{DRAM}^\text{MoE}
    &= L\bigl(\underbrace{4d_\text{model}^2\,q_w}_{\text{attn weights}}
      + \underbrace{2d_\text{model}\cdot C\,q_\text{kv}}_{\text{KV cache}}
      + \underbrace{d_\text{model}\cdot E\,q_w}_{\text{router}}\bigr), \\
  b_\text{DRAM}^\text{PKM}
    &= L\bigl(\underbrace{4d_\text{model}^2\,q_w}_{\text{attn weights}}
      + \underbrace{2d_\text{model}\cdot C\,q_\text{kv}}_{\text{KV cache}}\bigr)
      + \tfrac{L}{\Delta}\bigl(\underbrace{2d_\text{model}d_v\,q_w}_{\text{gate+down}}
        + \underbrace{N_\text{router}\,q_w}_{\text{router}}\bigr)
      \notag\\
    &\quad + L\!\left(1-\tfrac{1}{\Delta}\right)
      \underbrace{3\,d_\text{model}\,d_\text{ff}\,q_w}_{\text{regular FFN}},
\end{align}
where $C$ is context length.
Because the $2d_\text{model}C\,q_\text{kv}$ KV-cache term grows with context length while
all weight terms are fixed, $b_\text{DRAM}$ is dominated by the KV cache
at long context for all three architectures.

\subsubsection{\texorpdfstring{$b_\text{flash}$}{b\_flash} - Inference at the Edge}

\emph{Static} parameters (same weights used regardless of the input token)
are held in DRAM. \emph{Dynamic} parameters (selected per token by the
routing or retrieval mechanism) are paged from flash.

For \textbf{Dense}, all FFN parameters are static:
$b_\text{flash}^\text{Dense} = 0$.

For \textbf{MoE}, the activated expert matrices are dynamic; $\min(E,W)$
distinct expert matrices are loaded per window of $W$ tokens. Each distinct
expert incurs one cold random-read request followed by a contiguous transfer of
its weights:
\begin{align}
  b_\text{flash}^\text{MoE}
    &= L\,q_w\,\min(E,\,W)\cdot P_e,
    \qquad P_e = 3\,d_\text{model}\,d_\text{ff}, \\
  t_\text{access}^\text{MoE}
    &= L\left\lceil\frac{\min(E,\,W)}{Q_\text{flash}}\right\rceil
      t_\text{random}, \\
  t_\text{flash}^\text{MoE}
    &= t_\text{access}^\text{MoE}
      + \frac{b_\text{flash}^\text{MoE}}
        {\alpha_\text{flash}\mathcal{B}_\text{flash}}.
\end{align}

For \textbf{PKM}, the retrieved value vectors are dynamic; only the
$L/\Delta$ PKM layers contribute flash traffic. Each retrieved value is stored
as one contiguous row, but its address is assumed cold and random. The model
charges one request for every retrieved row and queues the $Wk_\text{total}$
requests in each PKM layer:
\begin{align}
  b_\text{flash}^\text{PKM}
    &= \frac{L}{\Delta}\,q_w\,W\,k_\text{total}\,d_v,
    \qquad k_\text{total}=n_\text{heads}k_\text{PKM}, \\
  t_\text{access}^\text{PKM}
    &= \frac{L}{\Delta}
      \left\lceil\frac{Wk_\text{total}}{Q_\text{flash}}\right\rceil
      t_\text{random}, \\
  t_\text{flash}^\text{PKM}
    &= t_\text{access}^\text{PKM}
      + \frac{b_\text{flash}^\text{PKM}}
        {\alpha_\text{flash}\mathcal{B}_\text{flash}}.
\end{align}
With top-$1$ MoE, $P_e \approx 78.6\text{M}$ parameters per expert, while
$k_\text{total}\,d_v \approx 0.3\text{M}$ parameters are retrieved per PKM
layer. The per-token flash-traffic gap is therefore about
$78.6/0.3 \approx 260\times$, independent of $T/N$.

\paragraph{Bandwidth component.}
Ignoring access latency, MoE and PKM have the same arithmetic intensity per
loaded byte:
$\text{AI}_\text{flash} = 2/q_w\;\text{FLOPs/byte}$.
PKM's bandwidth advantage comes from loading $260\times$ fewer bytes, while
its realized throughput also depends on $t_\text{access}^\text{PKM}$.

\subsection{Hardware Parameters}

\begin{table}[ht]
\centering
\caption{Cluster hardware parameters used in Section~\ref{sec:hardware-implications}.}
\label{tab:hardware-params-cluster}
\small
\begin{tabular}{lll}
\toprule
Parameter & Value & Efficiency factor \\
\midrule
Peak compute ($F_\text{peak}$) & 4500\,TFLOP/s & $\alpha_F=0.35$ \\
HBM capacity ($G_\text{HBM}$) & 192\,GB/GPU & --- \\
HBM bandwidth ($\mathcal{B}_\text{HBM}$) & 8\,TB/s & $\alpha_\text{HBM}=0.80$ \\
NVLink bandwidth ($\mathcal{B}_\text{NVLink}$) & 900\,GB/s & $\alpha_\text{NVLink}=0.70$ \\
GPUs per NVLink domain & 72 & --- \\
\bottomrule
\end{tabular}
\end{table}

\begin{table}[ht]
\centering
\caption{Mobile hardware parameters used in Section~\ref{sec:hardware-implications}.}
\label{tab:hardware-params-mobile}
\small
\begin{tabular}{lll}
\toprule
Parameter & Value & Efficiency factor \\
\midrule
Peak compute ($F_\text{peak}$) & 45\,TOPS & $\alpha_F=0.35$ \\
DRAM capacity ($G_\text{DRAM}$) & 12\,GB & $\alpha_G^\text{DRAM}=0.35$ \\
DRAM bandwidth ($\mathcal{B}_\text{DRAM}$) & 96\,GB/s & $\alpha_\text{DRAM}=0.70$ \\
Flash bandwidth ($\mathcal{B}_\text{flash}$) & 4\,GB/s & $\alpha_\text{flash}=0.70$ \\
Cold random-read latency ($t_\text{random}$) & $10\,\mu\text{s}$ & --- \\
Effective flash queue depth ($Q_\text{flash}$) & 32 & --- \\
\bottomrule
\end{tabular}
\end{table}

\subsection{Model Parameters for the Mobile Inference Simulation}

The mobile inference figure uses shared backbone hyperparameters:
$d_\text{model}{=}2560$, $d_\text{ff}{=}10240$, and $L{=}40$ layers. PKM uses
layer interval $\Delta{=}4$ (10 PKM layers and 30 regular dense layers).
Replacing every fourth FFN with the smaller PKM projection reduces its measured
active count to 3.3B, compared with 3.93B for MoE. These are the closest active
parameter counts obtainable while holding width, depth, and the other backbone
hyperparameters fixed; the comparison is therefore approximately, rather than
exactly, matched in active parameters.
Attention weights are in DRAM; FFN parameters in flash for PKM and MoE.
PKM uses $n_\text{heads}{=}4$ memory heads and retrieves
$k_\text{PKM}{=}32$ values per head, so $k_\text{total}{=}128$.
Three expansion ratios $T/N \in \{8, 16, 32\}$ are evaluated.

\paragraph{Expert count from $T/N$.}
Let $\alpha \approx 0.2$ be the attention fraction of active parameters.
With top-$k{=}1$:
\begin{equation}
  E = \frac{T/N - \alpha}{1-\alpha},
  \label{eq:expert-count}
\end{equation}
giving $E \in \{10, 20, 40\}$ after rounding for $T/N \in \{8, 16, 32\}$.

\paragraph{KV-cache crossover.}
KV-cache traffic equals FFN flash traffic when
$2d_\text{model}\,C\,L\,q_\text{kv} = 3\,d_\text{model}\,d_\text{ff}\,L\,q_w$, giving
$C = \tfrac{3\,d_\text{ff}\,q_w}{2\,q_\text{kv}}$.
With $q_w{=}0.5$ (int4) and $q_\text{kv}{=}1$ (int8):
$C = \tfrac{3 \times 10240 \times 0.5}{2 \times 1} \approx 7.7\text{k}$ tokens.

\subsection{Simulation Assumptions}

The mobile inference simulation assumes:
\begin{itemize}
  \item \textbf{MoE}: perfect per-window expert cache, each expert loaded
    at most once per window $W$.
  \item \textbf{PKM}: no value caching or request coalescing.
\end{itemize}
Both architectures use the storage parameters in
Table~\ref{tab:hardware-params-mobile}; the $10\,\mu\text{s}$ random-read latency
and queue depth 32 define one effective storage scenario, not universal device
constants. For simplicity, we do not model request reordering, PKM row caching
or prefetching, or overlap of flash I/O with compute. These caching assumptions
are conservative toward PKM; they do not establish a lower bound on end-to-end
device performance.

At $W{=}512$, the modeled PKM speedups over MoE are $7.61\times$,
$15.23\times$, and $30.45\times$ for $T/N{=}8$, 16, and 32, respectively. We
report the lowest-expansion result as the conservative representative value;
PKM throughput remains constant as total capacity grows in this model.

\paragraph{Communication overlap assumption.}
The training roofline models GPU-to-GPU communication as
sequential with compute. Under chunk-pipelined collectives
(e.g.\ Megatron-LM interleaved dispatch),
communication and compute can overlap and the formula reduces to
$t_\text{step} = \max(t_\text{compute},\, t_\text{HBM},\, t_\text{A2A})$;
the sequential assumption is therefore conservative toward MoE.

\subsection{Hardware Assumptions for Coupled Resource Scaling}
\label{sec:coupled-scaling-hardware-assumptions}

We calibrate the coupled scaling path in
Section~\ref{sec:coupled-resource-scaling} to a configuration benchmarked in
the Megatron Core MoE technical report \citep{nvidia2026megatronmoe}:
training Qwen3-235B, an MoE model with 235B total and 22B active parameters,
on 256 H100 GPUs in BF16 precision, the precision used to train our models.
This configuration achieved 320 TFLOP/s per GPU. The report does not state its
MFU; relative to the H100 dense BF16 peak of 989.5 TFLOP/s, the measured
throughput corresponds to an MFU of $320/989.5\approx32.3\%$.

Dividing the total parameter count by the number of GPUs gives
$235\mathrm{B}/256\approx0.918$ billion parameters per GPU. This is an
aggregate ratio, not a single-device storage limit.

The measured throughput corresponds to
$320\ \mathrm{TFLOP/s}\times86{,}400\ \mathrm{s}=27.648$ EFLOP of compute per
GPU-day. With $16n$ GPUs trained for $n$ days,
$C(n)=16n\cdot n\cdot27.648\times10^{18}=4.424\times10^{20}n^2$ FLOP and
$T(n)=16n\cdot0.918\times10^{9}=14.69n\times10^{9}$ parameters.

\section{Stability of the Architecture Comparison under Resampling}
\label{sec:stability}
\raggedbottom

This appendix tests how much the winner map of
Figure~\ref{fig:architecture-memory-compute-optimal-loss} and the PKM--MoE
ranking of Figure~\ref{fig:h100-coupled-scaling} depend on the measurements
the laws were fitted to.

\paragraph{Procedure.}
We resample the measured token-budget endpoints of each architecture with
replacement and in their original number (192 PKM, 124 MoE, 57 dense
Transformer, 48 STEM), and refit its selected law (P6F for PKM and MoE, vP6F
for the dense Transformer and STEM) with the robust objective of the
validation fits. This gives 200 refits each for PKM and MoE and 100 each for
the dense Transformer and STEM. Each of 100 joint draws takes one refit per architecture and
recomputes the lowest-loss allocation (Appendix~\ref{sec:lowest-loss-allocation})
on a $41\times41$ grid of the
Figure~\ref{fig:architecture-memory-compute-optimal-loss} window and along the
path of Figure~\ref{fig:h100-coupled-scaling}. As a contrast, we repeat the
procedure with lineages, i.e.\ all endpoints of one $(N,M)$ configuration
(Appendix~\ref{sec:fitting-scaling-laws}), as the resampling unit (20 PKM, 15
MoE, 5 dense Transformer, 4 STEM). With so few lineages, a lineage resample keeps on average about
two-thirds of the configurations, whereas an endpoint resample keeps nearly
all of them; we therefore report endpoint resampling first and lineage
resampling as a more conservative contrast.

\paragraph{Winner map.}
Figure~\ref{fig:stability-map} colors every budget by the architecture that
wins most often across the draws. Above the dash-dotted line, the most frequent
winner equals the fitted one in 99\% of cells, and agreement with the fitted
winner rises from 66\% of endpoint draws for cells next to a region boundary to
over 99\% for cells at least a decade in compute or capacity away from one.
Agreement is also higher where the laws extrapolate less: 92\% in the
lower-left quarter of the window, nearest the measured models, against 87\% in
the upper-right quarter. For a given compute budget, the MoE--PKM
boundary stays within a quarter decade of capacity of its fitted position in
94\% of draws, and the upper edge of the dense Transformer region in 96\%.

\paragraph{Extreme overtraining.}
The dash-dotted line marks the compute at which a dense Transformer trains on
1000 tokens per parameter. Below the line the fitted laws prefer mostly MoE
or PKM, but only by a median of 0.016 in predicted
loss, and this preference is unstable (agreement 0.60 across endpoint draws).
There the sparse laws place most of the parameters in the active part,
close to the dense limit, whereas every measured sparse model is far sparser
(at most 34\% active parameters for PKM and 56\% for MoE). Laws fitted to
sparse models therefore do not constrain this nearly dense regime, and their
comparison with the dense law is unreliable there. Joint scaling laws for
dense and MoE models find the dense model optimal when memory is tight relative
to compute \citep[Table~2]{joint}, so we expect the dense Transformer to win
this region. Figure~\ref{fig:architecture-memory-compute-optimal-loss} shows it
accordingly: it displays the dense Transformer wherever the tokens-to-parameters
ratio is over 1000.

\paragraph{Coupled path.}
Figure~\ref{fig:stability-path} adds the spread of the bootstrap draws to
Figure~\ref{fig:h100-coupled-optimal-loss}; with endpoints resampled, the PKM
and MoE bands overlap only from $n\approx61$. Comparing the losses within each
draw, PKM has a lower predicted loss than MoE in
97\% of endpoint draws at $n=16$ and 83\% at $n=128$, and a
lower loss than the dense Transformer in at least 99\% up to $n=64$ and 97\%
at $n=128$ (Figure~\ref{fig:stability-path-dense}).

\paragraph{Resampling unit.}
Resampling whole lineages instead widens every band
(Figures~\ref{fig:stability-map-lineage}, \ref{fig:stability-path-lineage}
and~\ref{fig:stability-path-dense-lineage}):
above the 1000-token line, the average share of draws in which a cell keeps
its fitted winner falls from 92\% to 82\%, and the MoE--PKM boundary stays within
a quarter decade of its fitted position in 80\% of draws and the upper edge of
the dense Transformer region in 64\%. PKM has a lower predicted loss than MoE
in 88\% of lineage draws at $n=16$ and 69\% at $n=128$, and a lower loss than the dense
Transformer in at least 99\% up to $n=32$, 88\% at $n=64$ and 69\% at
$n=128$.

\begin{center}
\begin{minipage}{\textwidth}
\centering
\refstepcounter{figure}\label{fig:stability-map}
\begin{minipage}[t]{0.49\textwidth}
\vspace{0pt}
\centering
\refstepcounter{figurepanel}\label{fig:stability-map-endpoint}
{\small\textbf{(\alph{figurepanel})} Endpoints resampled\par}
\smallskip
\includegraphics[width=0.98\linewidth]{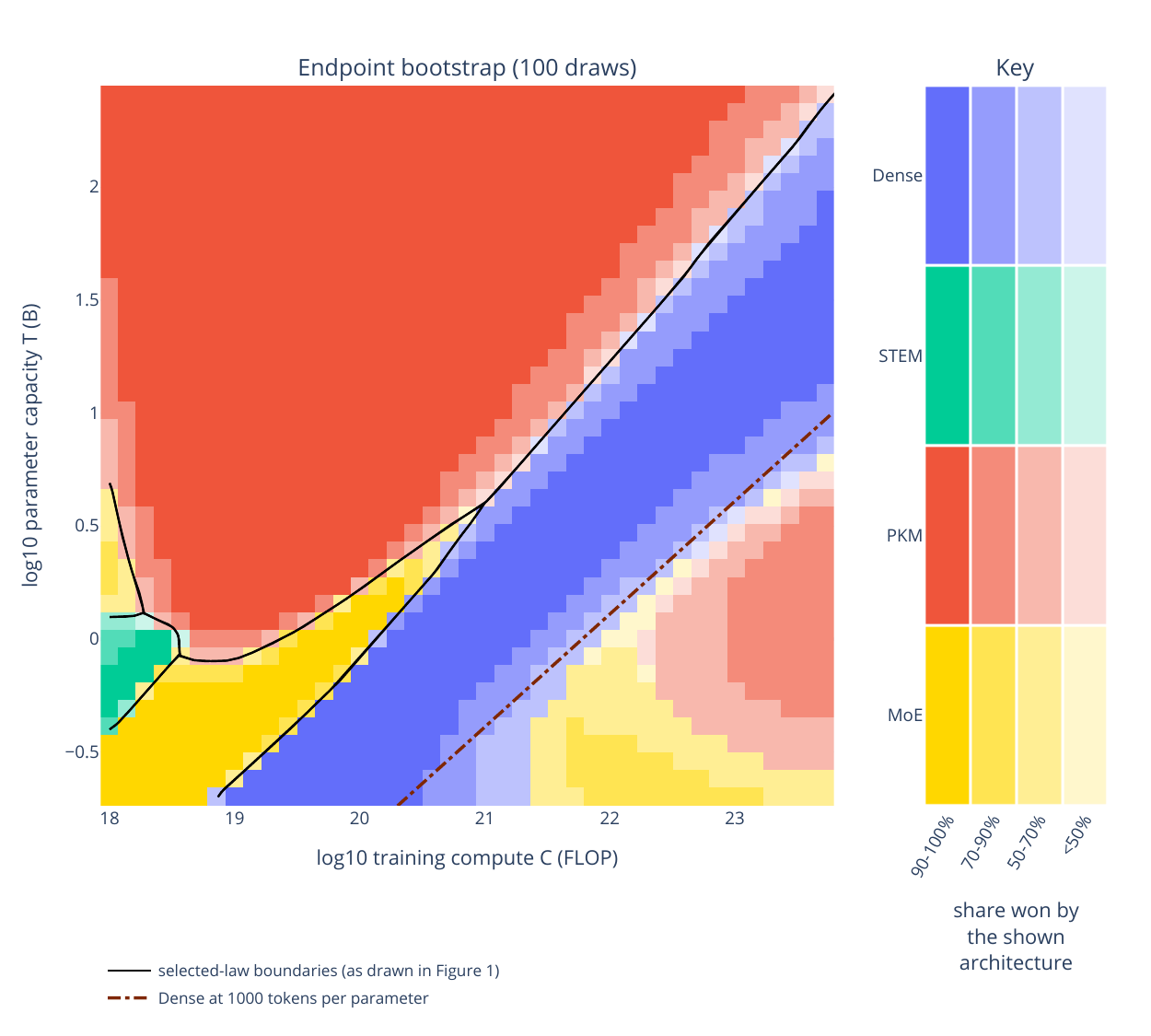}
\end{minipage}\hfill
\begin{minipage}[t]{0.49\textwidth}
\vspace{0pt}
\centering
\refstepcounter{figurepanel}\label{fig:stability-map-lineage}
{\small\textbf{(\alph{figurepanel})} Lineages resampled\par}
\smallskip
\includegraphics[width=0.98\linewidth]{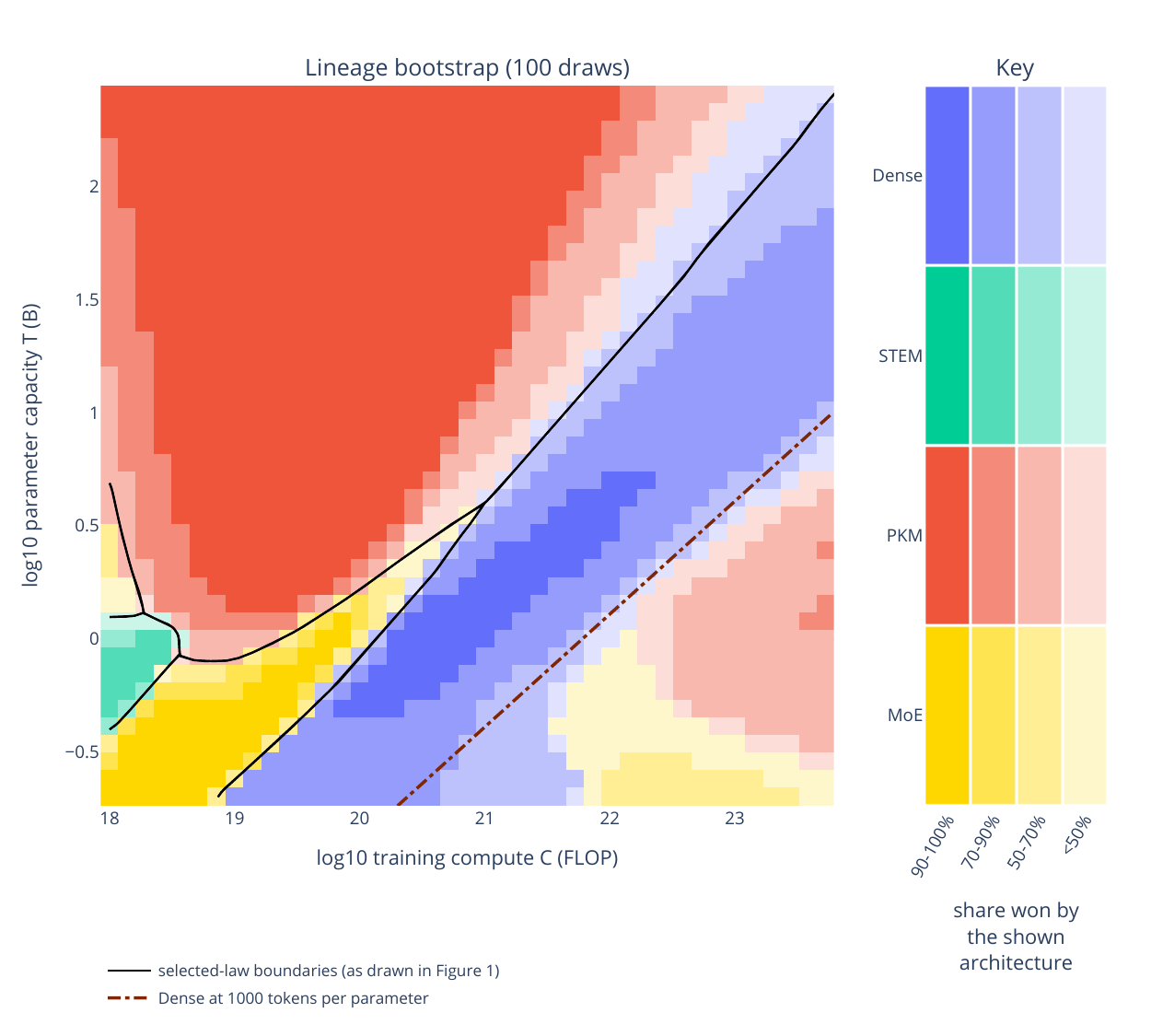}
\end{minipage}
\par\medskip
{\small\raggedright\textbf{Figure \thefigure.} \emph{Stability of the winner
map of Figure~\ref{fig:architecture-memory-compute-optimal-loss}.} Hue: the
architecture that wins most often across 100 joint bootstrap draws; shade: the
share of draws it wins. Black lines: region boundaries as drawn in
Figure~\ref{fig:architecture-memory-compute-optimal-loss}. Dash-dotted line: a
dense Transformer trained on 1000 tokens per parameter.\par}
\end{minipage}
\end{center}

\begin{center}
\begin{minipage}{\textwidth}
\centering
\refstepcounter{figure}\label{fig:stability-path}
\begin{minipage}[t]{0.49\textwidth}
\vspace{0pt}
\centering
\refstepcounter{figurepanel}\label{fig:stability-path-endpoint}
{\small\textbf{(\alph{figurepanel})} Endpoints resampled\par}
\smallskip
\includegraphics[width=0.98\linewidth]{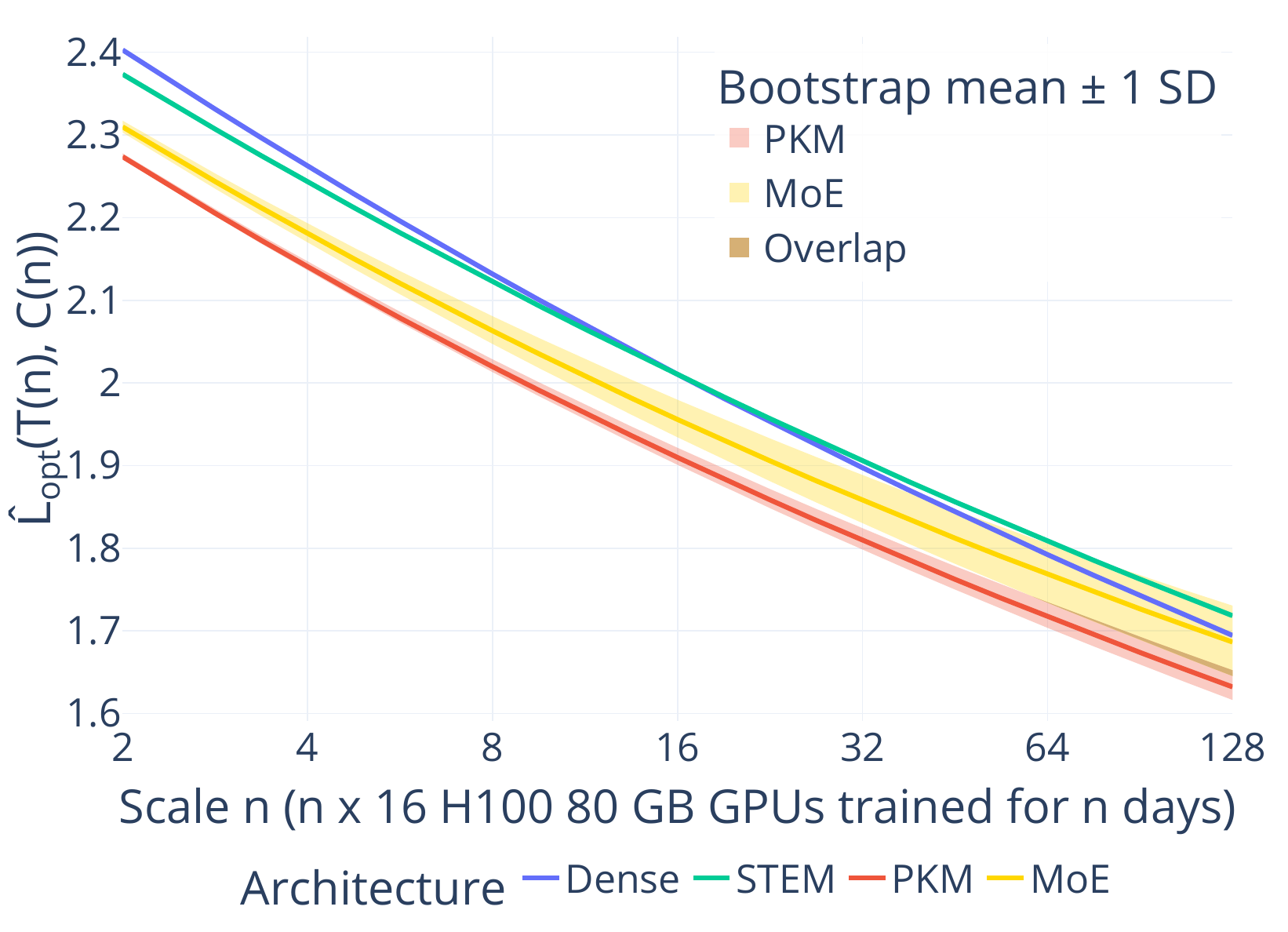}
\end{minipage}\hfill
\begin{minipage}[t]{0.49\textwidth}
\vspace{0pt}
\centering
\refstepcounter{figurepanel}\label{fig:stability-path-lineage}
{\small\textbf{(\alph{figurepanel})} Lineages resampled\par}
\smallskip
\includegraphics[width=0.98\linewidth]{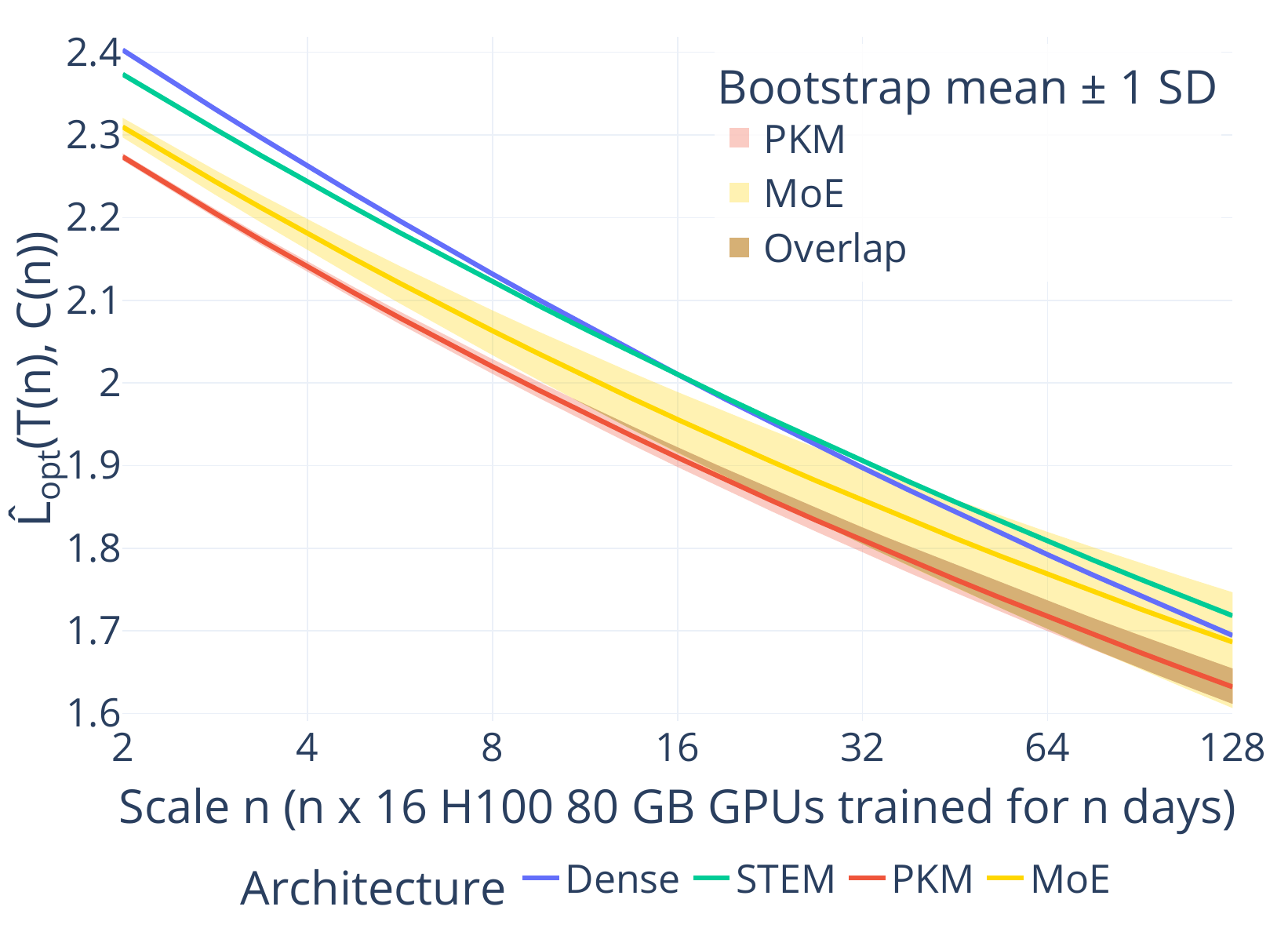}
\end{minipage}
\par\medskip
{\small\raggedright\textbf{Figure \thefigure.} \emph{Stability of the ranking
of Figure~\ref{fig:h100-coupled-scaling}.} Lowest predicted loss along the
coupled path, as in Figure~\ref{fig:h100-coupled-optimal-loss}. Lines: fitted
laws; bands: mean $\pm$ 1 standard deviation of the PKM and MoE losses over
100 joint bootstrap draws; darker: the two bands overlap.\par}
\end{minipage}
\end{center}

\begin{center}
\begin{minipage}{\textwidth}
\centering
\refstepcounter{figure}\label{fig:stability-path-dense}
\begin{minipage}[t]{0.49\textwidth}
\vspace{0pt}
\centering
\refstepcounter{figurepanel}\label{fig:stability-path-dense-endpoint}
{\small\textbf{(\alph{figurepanel})} Endpoints resampled\par}
\smallskip
\includegraphics[width=0.98\linewidth]{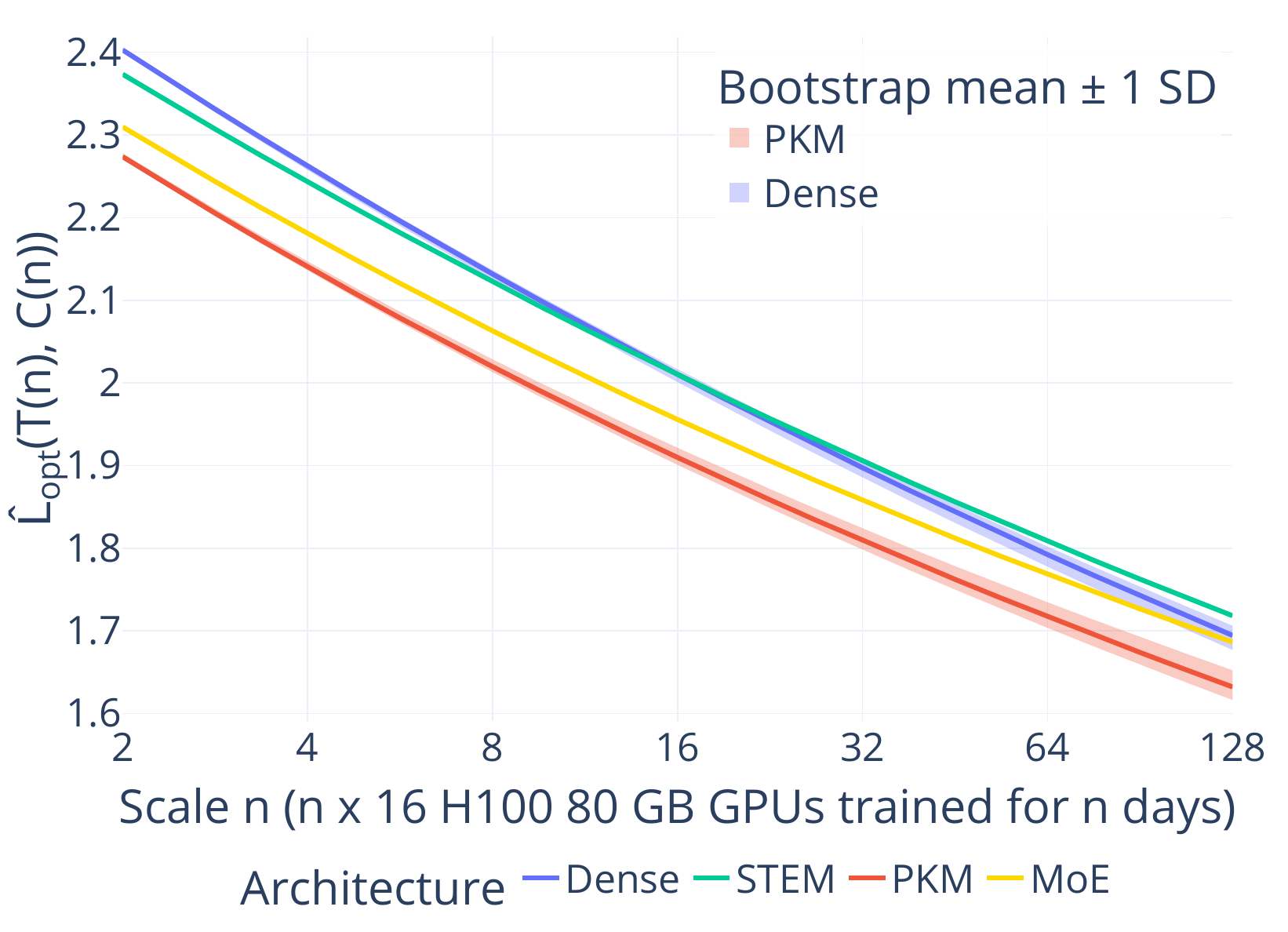}
\end{minipage}\hfill
\begin{minipage}[t]{0.49\textwidth}
\vspace{0pt}
\centering
\refstepcounter{figurepanel}\label{fig:stability-path-dense-lineage}
{\small\textbf{(\alph{figurepanel})} Lineages resampled\par}
\smallskip
\includegraphics[width=0.98\linewidth]{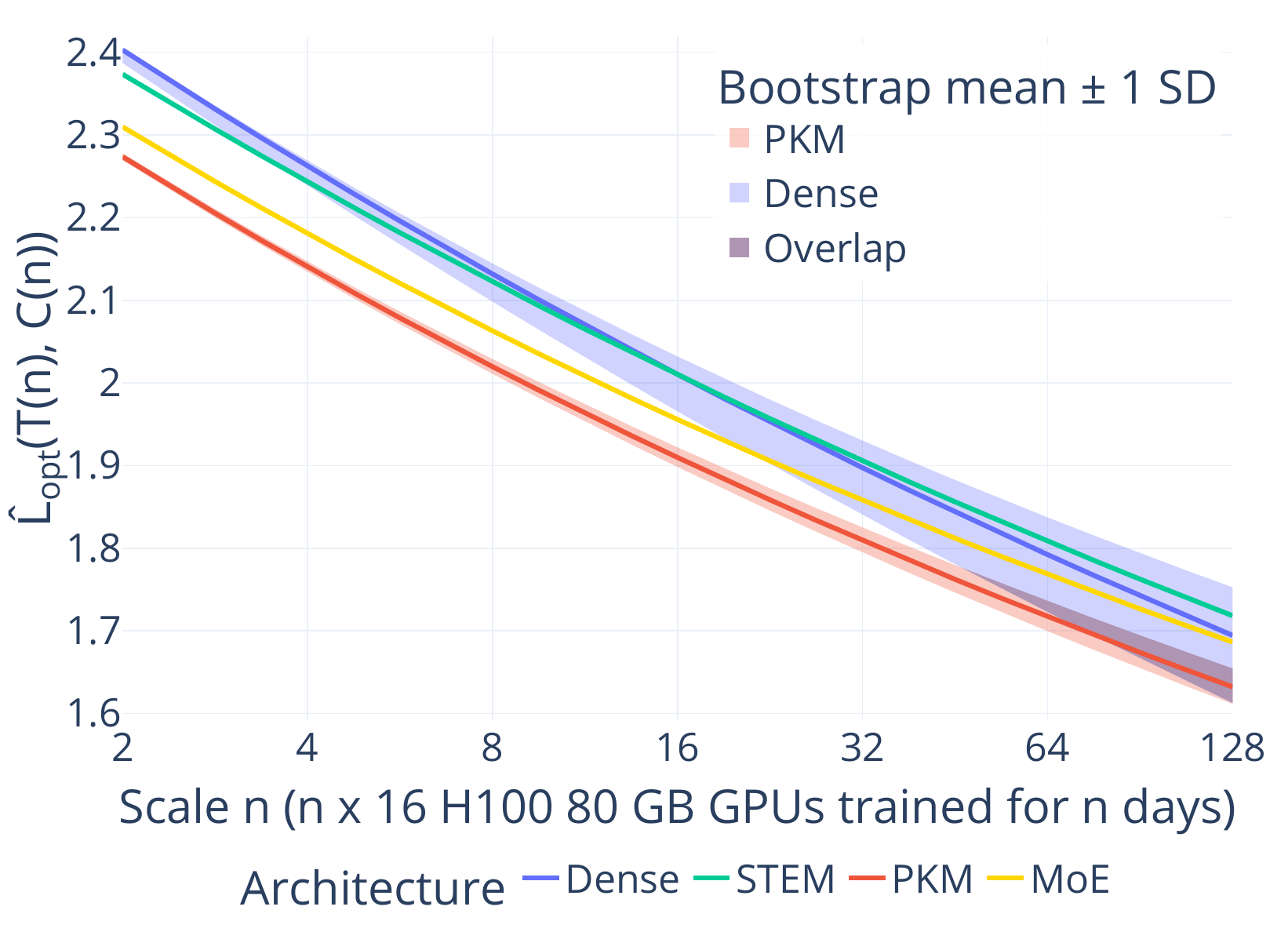}
\end{minipage}
\par\medskip
{\small\raggedright\textbf{Figure \thefigure.} \emph{Stability of the PKM--dense
Transformer ranking of Figure~\ref{fig:h100-coupled-scaling}.} As
Figure~\ref{fig:stability-path}, with bands for PKM and the dense Transformer
instead of PKM and MoE.\par}
\end{minipage}
\end{center}

\section{Optimal Training-Token Ratio along Coupled Scaling}
\label{sec:coupled-token-ratio}

Figure~\ref{fig:coupled-token-ratio} reports the ratio of training tokens to
total parameters, $D_{\mathrm{opt}}/T(n)$, chosen by the loss-minimizing PKM and
MoE allocations along the coupled scaling path of
Section~\ref{sec:coupled-resource-scaling}. The calibration of that path is per
GPU, so with $g$ H100 GPUs per unit of $n$ both budgets scale linearly with
$g$: $C(n)=4.424\times10^{20}(g/16)\,n^2$ FLOP and $T(n)=14.69\,(g/16)\,n$
billion parameters, while the operational budget remains
$D_{\mathrm{op}}(n)=n\times100$B tokens. The solid curves, $g=16$, correspond to
Figure~\ref{fig:h100-coupled-scaling}; the dashed and dotted curves use
$g=32$ and $g=64$.

With $g=16$, the optimal ratio rises from 9.7 to 52.3 for PKM and from
8.5 to 24.5 for MoE across the displayed range. More GPUs per unit of $n$
lower the ratio: with $g=64$ it reaches at most 34.7 for PKM and 17.4 for
MoE.

\begin{center}
\begin{minipage}{\textwidth}
\centering
\refstepcounter{figure}\label{fig:coupled-token-ratio}
\includegraphics[width=0.7\linewidth,trim=0bp 4bp 10bp 20bp,clip]{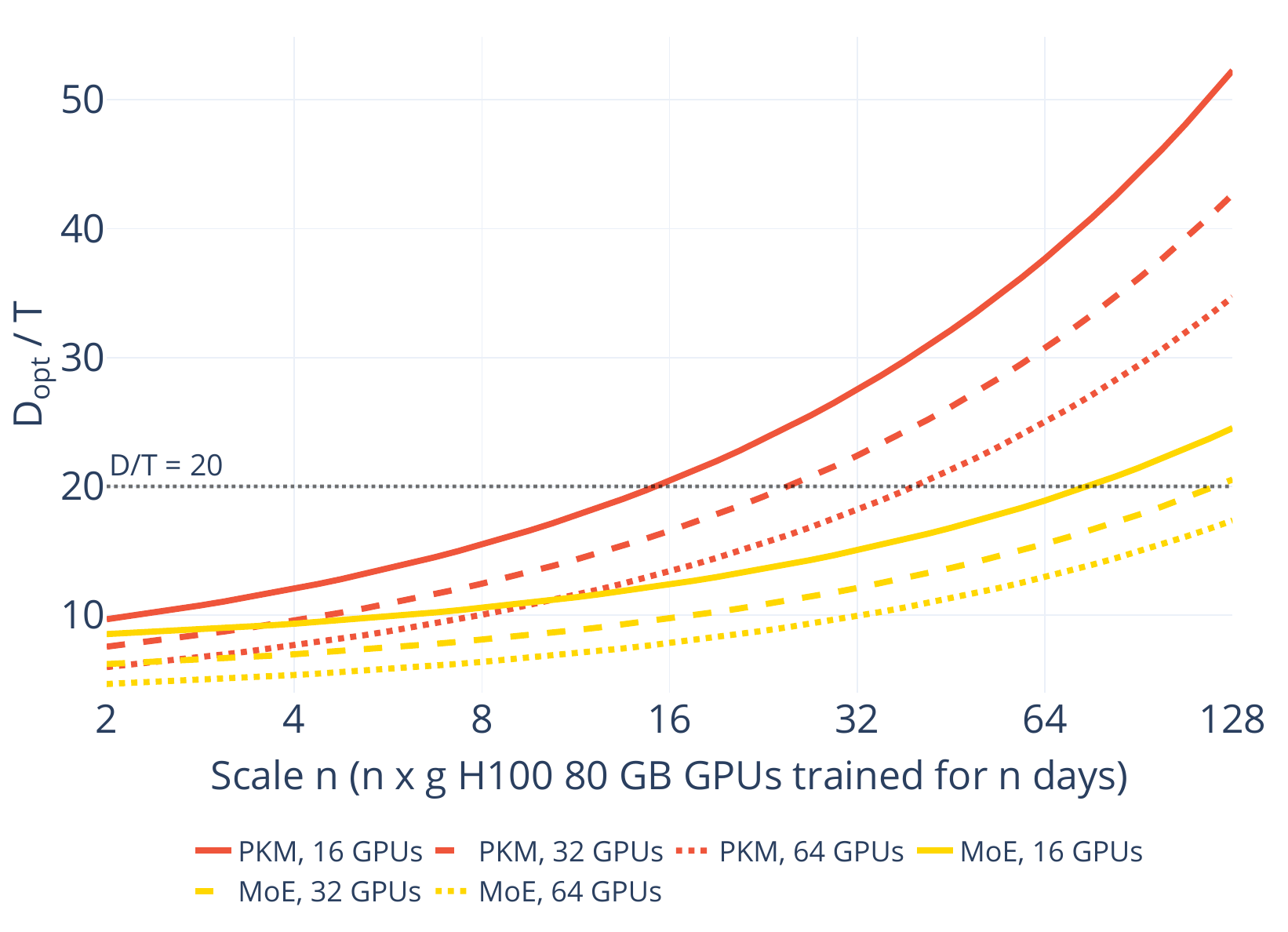}
\par\smallskip
{\small\raggedright\textbf{Figure \thefigure.} \emph{Optimal training tokens
per total parameter along coupled GPU-and-time scaling.} Budget path and
operational cost of Figure~\ref{fig:h100-coupled-scaling} with $g=16$ (solid),
32 (dashed), and 64 (dotted) H100 80 GB GPUs per unit of $n$. Each curve
reports $D_{\mathrm{opt}}(T(n),C(n))/T(n)$ of the independently optimized PKM
or MoE allocation; the gray dotted line marks $D/T=20$.\par}
\end{minipage}
\end{center}

\end{document}